%% file: main.tex
\documentclass[preprint]{elsarticle}
\input{macro}

\journal{Robotics and Autonomous Systems}


\begin{document}
\sloppypar
\begin{frontmatter}

% \title{LLM-Assisted Planning for Multi-Agent Systems}
\title{\mytitle}

\author[1]{Enrico Saccon\corref{cor1}}
\ead{enrico.saccon@unitn.it}
\author[2]{Matteo Saveriano}
\author[1]{Edoardo Lamon}
\author[1]{Luigi Palopoli}
\author[1]{Marco Roveri}

\cortext[cor1]{Corresponding author}

\affiliation[1]{
    organization={Department of Engineering and Computer Science, University of Trento}, 
    city={Trento},
    country={Italy}
}
\affiliation[2]{
    organization={Department of Industrial Engineering, University of Trento}, 
    city={Trento},
    country={Italy}
}

\sloppypar
\begin{abstract}
\input{sections/0-abstract}
\end{abstract}

% \begin{graphicalabstract}
% \includegraphics[]{}  
% \end{graphicalabstract}

%% Required for RAS (https://www.sciencedirect.com/journal/robotics-and-autonomous-systems/publish/guide-for-authors)
%% Examples can be found here: https://www.elsevier.com/researcher/author/tools-and-resources/highlights
% \begin{highlights}
% \item Introduces \textbf{\frameworkname}(PLanning with Natural language for Task-Oriented Robots), a novel framework that integrates Large Language Models (LLMs) with Prolog-based knowledge management and planning for multi-robot systems.
% \item Employs a \emph{two-phase \kbase generation} process using LLMs to create a structured Prolog \kbase, ensuring \textit{reusability} and \emph{compositional reasoning}.
% \item Implements a \emph{three-step planning procedure} that handles temporal dependencies, resource constraints, and parallel task execution via constraint-based scheduling, generating executable \emph{behavior trees}.
% \item Demonstrates the effectiveness of the framework in \emph{multi-robot assembly tasks}, showing that LLM-generated \kbases, with modest human feedback, can support scalable planning.
% \end{highlights}

%% From 1 to 7 max

\end{frontmatter}

% To be commented before sumbission
% \clearpage
% \tableofcontents
% \clearpage

\input{sections/include}

\section*{Acknowledgements}
Co-funded by the European Union under the INVERSE project (Grant Agreement No. 101136067) and the MAGICIAN project (Grant Agreement No. 101120731).

\bibliographystyle{my-elsarticle-num}
\bibliography{biblio.bib}

%%%%%%%%%%%%%%%%%%%%%%%%%%%%%%%%%%%%%%%%%%%%%%%%%%%%%%%%%%%%%%%%%%%%%%%%
\input{sections/8-appendix}

% \section{Reviews}
% \includepdf[pages=-]{RASreviews.pdf}

\end{document}

%% file: macro.tex
\usepackage[frozencache, cachedir=minted-cache]{minted2}
\setminted{escapeinside=\#\%}

\usepackage{graphicx}%
\usepackage{multirow}%
\usepackage{amsmath,amssymb,amsfonts}%
\usepackage{amsthm}%
\usepackage{mathrsfs}%
\usepackage[title]{appendix}%
\usepackage{xcolor}%
\usepackage{textcomp}%
\usepackage{manyfoot}%
\usepackage{booktabs}%
\usepackage{listings}%

\makeatletter
\@ifpackageloaded{hyperref}
  {} % Already loaded: do nothing
  {\usepackage[hidelinks]{hyperref}}
\makeatother

\usepackage{lipsum}
\usepackage[inline]{enumitem}
\usepackage[listings, minted, most]{tcolorbox}
\usepackage{etoolbox}
\usepackage{subcaption}
\usepackage{pdfpages}
\usepackage{pgfplots}
\pgfplotsset{compat=1.18}

\usepackage{microtype}

\usepackage{mathabx}
\usepackage{times}
\usepackage{xspace}

\tikzset{
    bar/.style={
        fill=blue!70,
        draw=blue!80!black,
        line width=0.4pt
    }
}
\usetikzlibrary{arrows.meta,calc}

\usepackage{todonotes}
\usepackage{cleveref}
\crefname{equation}{Equation}{Equations}

\usepackage[ruled,vlined]{algorithm2e}

\usepackage{makecell}

\setlist{noitemsep, nolistsep}
\usetikzlibrary{shapes.geometric}

\newtheoremstyle{thmstyleone}%
  {3pt}% Space above
  {3pt}% Space below
  {\itshape}% Body font
  {}% Indent amount
  {\bfseries}% Theorem head font
  {.}% Punctuation after theorem head
  { }% Space after theorem head
  {}% Theorem head spec

\newtheoremstyle{thmstyletwo}%
  {3pt}% Space above
  {3pt}% Space below
  {\itshape}% Body font
  {}% Indent amount
  {\bfseries}% Theorem head font
  {.}% Punctuation after theorem head
  { }% Space after theorem head
  {}% Theorem head spec

\newtheoremstyle{thmstylethree}%
  {3pt}% Space above
  {3pt}% Space below
  {\itshape}% Body font
  {}% Indent amount
  {\bfseries}% Theorem head font
  {.}% Punctuation after theorem head
  { }% Space after theorem head
  {}% Theorem head spec

\theoremstyle{thmstyleone}

\theoremstyle{thmstyletwo}%
\theoremstyle{thmstylethree}%

\newcommand{\aStart}[1]{
\ifmmode
\texttt{$#1$}_\vdash % These cannot be changed to \verb
\else 
$\texttt{$#1$}_\vdash$
\fi
}
\newcommand{\aEnd}[1]{
\ifmmode
\texttt{$#1$}_\dashv % These cannot be changed to \verb
\else 
$\texttt{$#1$}_\dashv$
\fi
}
\newcommand{\pc}[1]{pre(#1)}
\newcommand{\eff}[1]{eff(#1)}
\newcommand{\inv}[1]{inv(#1)}
\newcommand{\fl}[1]{fl(#1)}
\newcommand{\ach}[1]{\text{Enbl}(#1)}
\newcommand{\tn}[1]{\textnormal{#1}}

\newtcolorbox{textbox}[2][]{
    colback=gray!5!white,colframe=gray!75!black,
    left=1.5mm, lefttitle=4mm, right=1.5mm, 
    title=#2,
    #1
}

\newtcolorbox{textboxerror}[1][]{
    colback=red!5!white,colframe=red!75!black,
    #1
}

\newtcblisting[list inside=loe]{codebox}[2]{
    listing engine=minted, minted language={#1}, listing only,
    minted options={breaklines, autogobble, linenos, fontsize=\scriptsize, numbersep=3mm},
    colback=gray!5!white,colframe=gray!75!black,
    left=5.5mm, lefttitle=5mm, right=1.5mm, 
    bottom=0mm, top=0mm,
    title=#2, fonttitle=\small
}

\makeatletter
\newcommand{\writings}[1]{#1\xspace}
\newcommand{\oldwritings}[1]{%
    #1%
    \@ifnextchar.{}{%
        \@ifnextchar,{}{%
            \@ifnextchar:{}{%
                \@ifnextchar'{}{%
                    \@ifnextchar;{}{%
                        \@ifnextchar-{}{%
                            \@ifnextchar){}{\ }%
                        }%
                    }%
                }%
            }%
        }%
    }%
}

\newcommand{\LL}[0]{\protect\writings{low-level}}
\newcommand{\HL}[0]{\protect\writings{high-level}}

\newcommand{\llms}[0]{\protect\writings{LLMs}}

\newcommand{\kb}[0]{\protect\writings{KB}}
\newcommand{\kbs}[0]{\protect\writings{KBs}}
\newcommand{\kbase}[0]{\protect\writings{knowledge-base}}

\newcommand{\kbases}[0]{\protect\writings{knowledge-bases}}

\newcommand{\bt}[0]{\protect\writings{BT}}
\newcommand{\bts}[0]{\protect\writings{BTs}}
\newcommand{\btree}[0]{\protect\writings{behavior tree}}

\newcommand{\Btrees}[0]{\protect\writings{Behavior Trees}}

\newcommand{\add}[1]{\protect\writings{\texttt{add(#1)}}} % This cannot be changed to \verb
\newcommand{\del}[1]{\protect\writings{\texttt{del(#1)}}} % This cannot be changed to \verb

\newcommand{\frameworkname}[0]{\protect\writings{PLANTOR}}

\newcommand{\cm}[0]{\textbf{\checkmark}}

\def\mytitle{Combining Large Language Models and Symbolic Reasoning for Multi-Robot Temporal Planning through Explainable Knowledge Bases}
\newcommand{\mykeywords}[0]{Large Language Models\sep Symbolic Reasoning\sep Temporal Task Planning\sep Multi-Robot Systems\sep Knowledge Representation}
\makeatother

%% file: sections/0-abstract.tex
% This paper presents a framework called \frameworkname (PLanning with Natural language for Task-Orientated Robots), which integrates Large Language
% Models (LLMs) with Prolog-based knowledge management and planning
% for multi-robot tasks. The system employs a two-phase generation of
% a robotic \kbase to support reuse across implementations,
% as well as a three-step planning procedure
% that handles temporal dependencies and
% parallel task execution via constraint-based scheduling. The
% final plan is converted into a Behavior Tree for direct use on the robots. 
% We tested the framework in multi-robot assembly tasks within a
% block world and an arch-building scenario. Results indicate that
% LLMs can produce structured \kbases with modest human
% feedback, while Prolog provides a transparent and inspectable
% representation. This approach underscores the potential of LLM
% integration for robotics tasks requiring flexible and
% human-understandable planning.
We present \frameworkname, a framework for generating and executing multi-robot task plans from natural-language task descriptions through LLM-assisted knowledge-base construction. The approach uses large language models to synthesize a structured Prolog \kbase, applies consistency checks to detect and repair modeling errors, generates a high-level symbolic plan, refines it into low-level robot actions, and computes a temporally optimized schedule that is converted into an executable behavior tree. The framework is designed to preserve inspectability by exposing the generated \kbase, intermediate plans, and scheduling constraints. We evaluate the approach on scenarios inspired by the Blocks World and Grippers benchmark across multiple language models, and we report both the quality of generated \kbases and the runtime of the planning pipeline. We further demonstrate end-to-end execution in a real multi-arm assembly scenario. The results show that LLM-generated \kbases can substantially reduce manual modeling effort, but may still require human correction. Overall, the paper argues for a hybrid workflow in which language models are used to produce structured symbolic artifacts, while correctness-critical planning and scheduling remain symbolic and inspectable.

%% file: sections/include.tex
%%%%%%%%%%%%%%%%%%%%%%%%%%%%%%%%%%%%%%%%%%%%%%%%%%%%%%%%%%%%%%%%%%%%%%%%
\section{Introduction}\label{sec:intro}
\input{sections/1.IntroRevised}
% \input{sections/1-introduction}

%%%%%%%%%%%%%%%%%%%%%%%%%%%%%%%%%%%%%%%%%%%%%%%%%%%%%%%%%%%%%%%%%%%%%%%%
\section{Problem Description and Solution Overview}\label{sec:probdesc}
\input{sections/2-probdesc}

%%%%%%%%%%%%%%%%%%%%%%%%%%%%%%%%%%%%%%%%%%%%%%%%%%%%%%%%%%%%%%%%%%%%%%%%
\section{Related Work}\label{sec:relatedwork}
\input{sections/6-relatedworks}

%%%%%%%%%%%%%%%%%%%%%%%%%%%%%%%%%%%%%%%%%%%%%%%%%%%%%%%%%%%%%%%%%%%%%%%%

\section{Background and Definitions}\label{sec:background}
\input{sections/3-background}
\input{sections/4-methodology}
\section{Experimental Evaluation}\label{sec:experiments}
\input{sections/5-experiments}

%%%%%%%%%%%%%%%%%%%%%%%%%%%%%%%%%%%%%%%%%%%%%%%%%%%%%%%%%%%%%%%%%%%%%%%%
\section{Final Remarks and Conclusion}\label{sec:conclusions}
\input{sections/7-conclusion}

%% file: sections/1.IntroRevised.tex
The emergence of Large Language Models (LLMs)~\cite{brown_language_2020,zhao_LLM_survey_2024} has created unprecedented opportunities in the field of robotics. For the first time, robotic tasks can be specified directly through natural-language procedural descriptions~\cite{pmlr-huang_language_2022-huang22a,singh_progprompt_2022}, or through existing technical documentation~\cite{tie2025manualskill}. But these are only some of the opportunities offered by this technology. Many LLMs are able to exploit multi-modal features, enabling robots to build a coherent representation of virtually all the information that can be processed by a human being. 
One of the most important examples is offered by Vision Language Models (VLMs)~\cite{radford_learning_2021,chen_open-vocabulary_2022}, which can interpret and describe a scene captured by a camera or a video clip. The collected information introduces a semantic layer in the representation of the environment, which can be directly matched with the robot task description, creating continuity between perception and acting. 
Vision-Language-Action models (VLAs) allow for the most ambitious perspective of pushing this frontier to its limits, enabling the robot not only to understand its reality but also to operate in ways learned from the observation of humans~\cite{zitkovich_rt2_2023}. Ideally, in the same way that robots learn to decode language constructs and interpret reality, they can understand and imitate how humans operate to accomplish simple operations and combine them to fulfill more complex missions.

This is certainly a fascinating perspective; however, current end-to-end approaches are not yet sufficient for many robotic applications and have some important limitations that must be addressed. First of all, physical interaction data are much more expensive and difficult to collect and store than text and image data readily available on the internet. Moreover, simple visual observations can offer only partial clues, and the result is that, despite the important investments and improvements in VLAs, their performance falls short of expectations, especially when faced with language-conditioned manipulation tasks that require common-sense transfer, semantic interpretation, and long-horizon reasoning~\cite{Zhang_2025_ICCV,Fei_2026_CVPR}. Second, many robotic applications are safety critical and mission critical. Thus, problems such as hallucinations~\cite{ji_survey_2023,ren_robots_2023}, sycophancy~\cite{sharma_skill_2021,malmqvist2025sycophancy}, i.e., the tendency to align with the user's beliefs or preferences even when they are incorrect, and the lack of explainability of the decisions taken~\cite{pmlr-v305-haon25a} could lead to catastrophic failures. In such contexts, the ability to inspect, validate, and correct the decision-making process is not a secondary requirement but a central design constraint.

Until these challenges are adequately addressed, a more pragmatic alternative is to use LLMs not as direct robot policies, but as tools for producing structured artifacts, which can subsequently be checked and executed by symbolic or algorithmic components. This holds the promise of enabling the application of LLMs to industrial scale robotic problems even within the current technical framework. A straightforward example is the code-as-policies paradigm, in which an LLM can generate executable code by composing calls from a finite set of robot APIs \cite{Lia23}, a much narrower scope than VLAs. Moreover, it comes with important advantages, namely reducing the burden on code developers, exploiting existing robot skills, and producing artifacts that can be read, edited, and debugged by human operators.

% Until these challenges are adequately addressed, we have a more pragmatic alternative, which holds the promise to enable the application of LLMs to industrial scale robotic problems even within the current technical framework: code-as-a-policy. The idea pioneered in robotics by Liang et al.~\cite{Lia23} is that LLMs should not be trained to learn policies but to synthesise code relying on APIs that encode a finite repertoire of possible robot actions. This alternative is obviously more narrow in scope than VLAs, since it constrains the AI to choose among a finite set of possible actions, rather than looking for smooth combinations of learned behaviors. However, it comes with significant advantages. First, from the perspective of the human developer, LLMs can radically simplify robot software development, a notoriously complex activity. Second, guiding the code generation in a principled way leads to reusable results. Third, the outcome of the LLM can be read, interpreted and fine-tuned by a human, leading to a natural interpretability of the robot operations.

Procedural code, however, describes how to accomplish a task rather than what is known about the domain; for instance, the objects involved, their relations, and the constraints imposed on the resources. This distinction has practical consequences: the generated code can be read by a human, but it cannot be formally queried to detect and resolve logical inconsistencies through automated reasoning. Moreover, since the code is tightly coupled to implementation details (classes, function calls, APIs) it is difficult to reuse across different domains or to exploit it directly for planning. A natural candidate to address these limitations is the Planning Domain Definition Language (PDDL)~\cite{mcdermott_pddl_1998,fox_pddl21_2003}, which provides a structured, domain-independent representation of planning problems. This is indeed an interesting option (see Section~\ref{sec:relatedwork} for a brief survey of the different authors that advocated PDDL generation). However, PDDL is designed for plan generation, not for knowledge representation: it cannot be directly queried to perform arbitrary logical inference. Our objective in this work is slightly different: we seek an intermediate representation that is not only suitable for planning but also acts as an explicit, queryable, incrementally revisable \kbase, with clear, explainable inference rules, and such that possible inconsistencies can be logically identified. For this reason, we explore a Prolog-centered pipeline rather than a purely PDDL-centered one. Our goal is not to replace PDDL-based planners, but to introduce a Prolog-based knowledge-management layer that can support validation, querying and explainability and that could in principle be coupled with PDDL planners when planner scalability is the main concern. 

Prolog~\cite{bratko_prolog_2001} offers a useful middle ground between the flexibility of natural-language and the reliability of symbolic reasoning. Knowledge is represented through facts and rules, and it can be queried, validated, and used for plan generation \cite{clocksin_programming_2003}. These properties have already made logic-based knowledge representation relevant in robotics, for example, in systems for robotic knowledge management, reasoning, and planning \cite{tenorth_knowrob_2013,dix_planning_2010}. In a Prolog \kbase, domain facts, action preconditions, action effects, object relations, and resource constraints can be represented in an explicit and inspectable form. This makes it possible to trace why an action is applicable, why a goal can or cannot be reached, and where a modeling inconsistency may arise.

\paragraph{Paper Contributions} This paper proposes PLANTOR\footnote{The code-base is available online at \url{https://github.com/icosac/plantor_tp/tree/RAS-2026}. The artifacts for experimental evaluation are archived on Zenodo~\cite{saccon_2026_21630757}.}, a framework that uses LLMs to synthesize a \kbase (\kb) in Prolog, hence encoding in logical form the domain in which the robots operate, the objects they manipulate, and the actions they can execute. This choice naturally supports a predictable, explainable, and formally verifiable decision process, qualities that procedural code and PDDL cannot fully provide. Specifically, we make the following contributions within PLANTOR:

\begin{enumerate}
\item A semi-automatic human-in-the-loop procedure to generate a Prolog \kb from natural language descriptions, with automated consistency checks that detect syntactic and semantic errors before planning commences. The \kb is organized into a high-level component, capturing abstract tasks, and a low-level component, encoding the specific capabilities of the robotic platform. This two-layer structure supports maintainability %, reuse across different platforms,
and efficiency in plan generation.

\item A logic-based planning algorithm that performs a forward state-space search over the \kb, generating a total-order plan, i.e., an ordered sequence of actions that satisfies the goal with respect to the validated KB.

\item A constraint-preserving optimization framework that extracts parallelism from the agent-grounded plan and minimizes its makespan.

\item A translator that converts the optimized plan into an executable behavior tree, enabling direct deployment to robotic platforms.
\end{enumerate}

The experimental results, conducted in multi-robot assembly scenarios of increasing complexity, indicate that LLMs can generate consistent \kbases with modest human supervision, and that the planning pipeline produces feasible, parallelized plans within practical time bounds. A real-life experiment on two collaborative robotic arms confirms that the framework bridges the gap between natural language task descriptions and concrete robot operations, demonstrating its potential for deployment in practical robotic applications.

% \paragraph{Paper Organization}
% %The paper is organized as follows. 
% In Section~\ref{sec:probdesc}, we describe our problem in detail and provide an overview of the framework. 
% In Section~\ref{sec:relatedwork}, we discuss the related work, while in Section~\ref{sec:background}, we describe the main background technologies used in this paper and formalize the planning problem. 
% In Section~\ref{sec:kb}, we describe the approach for the creation and the population of the \kbase. 
% In Section~\ref{sec:plangen} and Section~\ref{ssec:bt}, we detail our algorithmic solution to generate an executable plan from the \kb. 
% In Section~\ref{sec:experiments}, we give a full account of the experimental validation of the framework.  
% Finally, in Section~\ref{sec:conclusions}, we offer our conclusions and discuss future work directions.

%% file: sections/2-probdesc.tex
We describe the problem and outline our solution strategy; the individual components are detailed in the following sections.

\subsection{Problem Description}

Our goal is to streamline the execution of a robotic task starting from an input composed of the task problem, the natural language description of the environment, and the action repertoire of  the available robots.
The goal is achieved through an explainable process, which generates an executable specification of the actions that will be dispatched to the run-time environment of the robots. 

The key requirements for the framework are the following:
\begin{itemize}
    \item \textbf{R1:} The execution of a task must be \textit{formally correct} with respect to the validated symbolic \kb, i.e., it must accomplish the goals and adhere to the constraints derived from the natural language text.
    \item \textbf{R2:} The process must be \emph{explainable}, meaning that the system exposes inspectable artifacts (e.g., the Prolog \kb, intermediate plans, and scheduling constraints) and explicit checks that allow a human to understand why a plan is feasible or not. 
    \item \textbf{R3:} The knowledge obtained from understanding the task and the problem to be solved must be \emph{reusable} across different implementation scenarios (e.g., using one robot or multiple robots).
    \item \textbf{R4:} The generation of the plan must support and optimize the \textit{parallel execution} of actions across the available robotic resources.
    \item \textbf{R5:} The final specification of the plan must be executable on a wide range of robotic devices.
\end{itemize}

We assume deterministic action models; uncertainty handling and probabilistic effects are beyond the scope of this work. The interested reader is referred to our recent paper on the automated generation of Markov Decision Processes for human-robot interaction~\cite{saccon_automated_2026}.

\subsection{Solution Overview}
\label{ssec:contributions}

\begin{figure*}[t!]
    \centering
    \includegraphics[width=\textwidth]{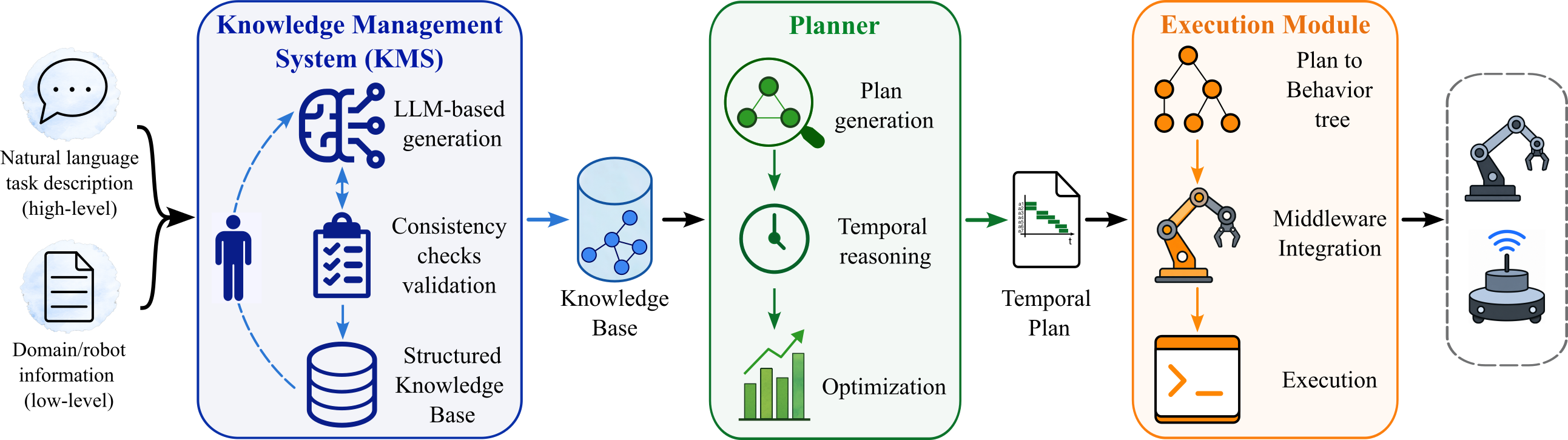}
    \caption{Overview of \frameworkname. Natural-language task descriptions and robot/domain information are processed by the KMS to generate and validate a Prolog \kb. The planner uses it to synthesize and temporally optimize an executable plan. The execution module converts the optimized plan into a behavior tree and dispatches it through the robotic middleware to the different agents in the scene.}
    \label{fig:arch_LLM_pKB}
\end{figure*}

The problem outlined above is addressed through the software framework depicted in Figure~\ref{fig:arch_LLM_pKB}. The framework is composed of the following modules:
\begin{itemize}
    \item \textbf{Knowledge Management System} (KMS): It takes the initial natural language inputs and extracts a \kb in Prolog.
    \item \textbf{Planner}: It determines an executable plan from the \kb.
    \item \textbf{Execution Module}: It executes the plan by leveraging integration with a middleware.
\end{itemize}

The KMS utilizes \llms to generate the \kb from natural language texts. In the first step, the description of the task and of the environment is used to generate a high-level \kbase, i.e., a set of logical predicates that encode the execution of a task with \HL predicates and actions. In the second step, the knowledge is augmented with low-level robot-specific information, specifying how each \HL action is implemented using the elementary actions provided in the natural language description and that the robot is capable of. The technical details of these steps are provided in Section~\ref{sec:kb}. 

This process is not fully automated, but relies on human-supervised steps, as detailed in Section~\ref{sec:kb}. Moreover, the generative models are used only for the generation of the \kb, while the planning task is purely symbolic. Using a Prolog \kb offers several advantages over directly generating an executable plan:
\begin{enumerate}[nosep]
  \item The \kb contains a formal statement of goal and constraints, which facilitates the generation of formally correct plans (\textbf{R1}).
  \item A Prolog \kb is compact, human-readable, and queryable, enhancing the generation process \emph{explainability} (\textbf{R2}). As shown in Section~\ref{sec:kb}, we have developed a series of predicates to ensure that some rules are respected in the generated \kbs.
  \item The ability to encode and query rules supports modularity and reuse across tasks and implementations (\textbf{R3}).
\end{enumerate}

\noindent The two-phase construction of the \kb offers two principal advantages. First, it enables the reuse of an identical \HL conceptual structure across heterogeneous robotic platforms by requiring modifications only to the \LL actions and the associated mappings. Second, it mitigates the combinatorial complexity of planning by first synthesizing a \HL plan over a compact abstract representation and subsequently refining this plan into a sequence of \LL actions.

Plan generation follows three steps. In the first step, detailed in Section~\ref{ssec:toplangen}, a forward state-space search is carried out starting from the initial state encoded in the \kb. Actions are tried in the order they appear in the \kb (as per Prolog semantics), until the final goal state is reached. This yields a totally ordered set of actions, but does not include timing information.

The second step, discussed in Section~\ref{ssec:poplangen}, utilizes Prolog's capabilities to analyze causal dependencies between actions based on their preconditions and effects. The outcome is a partially ordered plan, where sequencing constraints exist only between causally dependent actions. 

In the third step, detailed in Section~\ref{ssec:poplanopt}, a scheduling optimization model is formulated and solved with OR-Tools\footnote{\label{fn:or-tools}\url{https://developers.google.com/optimization}}, to optimize the timing of the actions. The model encodes causal relationships between actions (\emph{enablers}) and bounds on action durations. If successful, the resulting plan supports parallel execution (\textbf{R4}) and is translated into a Behavior Tree (\bt), a standard formalism for the execution of robotic plans in ROS2 (\textbf{R5}). Further details on this phase are provided in Section~\ref{ssec:bt}.

% If this operation fails, Prolog's backtracking capability can be employed to generate an alternative total-order and repeat the process. Persistent failures indicate a possible error in the \kb or domain description,  a clear sign that a substantial revision of both could be in order.

\paragraph{Extension beyond the preliminary version}

This article substantially extends our preliminary work~\cite{saccon_when_2024}. The present manuscript introduces the following new elements:

\begin{itemize}
    \item a complete knowledge-management pipeline in which the LLM generates the full Prolog knowledge-base, including static predicates, initial and goal states, high-level actions, low-level actions, and mappings, whereas the conference version used the LLM only to generate the initial and goal states;
    \item a two-layer representation separating high-level task knowledge from low-level robot-specific actions, together with explicit mappings between the two layers;
    \item prompt-validation and iterative Prolog-based consistency checks to detect and repair syntactic and semantic errors in the generated \kb before planning;
    \item an extended planning pipeline that replaces the previous uninformed depth-first search with a breadth-first search strategy, extracts causal dependencies, and optimizes the temporal schedule of the partial-order plan;
    \item explainability-oriented interfaces that expose why actions are applicable or not and how alternative branches are explored during symbolic planning;
    \item a substantially broader experimental evaluation, including benchmark-inspired Blocks World and Grippers scenarios, multiple language models, knowledge-base quality metrics, planner-runtime analysis, and a real multi-arm robotic experiment.
\end{itemize}

Overall, while the conference paper introduced the core idea of combining LLM-assisted Prolog knowledge acquisition with symbolic planning, the present article develops it into a full framework for validated knowledge-base generation, temporal multi-agent planning, and robotic execution.

%% file: sections/6-relatedworks.tex
We discuss here the most relevant related work related to our framework, and we highlight the gaps our approach addresses.

\paragraph{Knowledge-Base}
Knowledge representation provides the symbolic substrate for autonomous decision making and approximate common-sense reasoning in robotics \cite{bayat_requirements_2016,manzoor_ontology-based_2021}. Many systems use ontologies to encode entities and relations, often as semantic triples \cite{melnik_framework_2000}. The OpenRobots Ontology (ORO) \cite{lemaignan_oro_2010} exemplifies this approach with a server architecture that supports dynamic insertion and retrieval of symbolic facts. Ontology-based representations are frequently specialized to domains, e.g., surgical robotics \cite{goncalves_knowledge_2015} or industrial human--robot collaboration \cite{sadik_ontology-based_2017}.

Beyond purely ontological storage, knowledge processing frameworks such as KnowRob \cite{tenorth_knowrob_2013} and KnowRob2 \cite{beetz_know_2018} integrate perception and experience with logic-based inference (notably via Prolog), enabling more context-aware reasoning and acquisition of action-centric knowledge. However, these systems typically require carefully engineered and syntactically intricate \kbases, which can complicate maintenance and the integration of new information. We address this bottleneck by using LLMs to generate a Prolog \kb directly from natural language.

Closest to our contribution, \cite{saccon_automated_2026} also uses LLMs to produce a Prolog \kb, but targets sequential decision making by compiling a probabilistic model (MDP) in which action durations do not influence policy search. In contrast, our focus is temporal task planning with durative actions and explicit scheduling to maximize parallelism.

%\subparagraph{LLMs for Knowledge-Base / Domain Generation}
Recent work uses LLMs to translate natural language task descriptions into planning representations, often in PDDL, either by refining an existing domain (e.g., instruction-to-PDDL pipelines and generalized planners with CoT-style reasoning \cite{zhou_isr-llm_2024,wei_chain_2022,silver_generalized_2024}) or by combining LLM outputs with classical planners to improve reliability \cite{valmeekam_large_2022,liu_llmp_2023,oswald_large_2024}. Other methods aim to synthesize domains more directly, with iterative repair loops when generated models are inconsistent \cite{gestrin_nl2plan_2024,wang_llm3large_2024}. Multi-agent extensions such as TwoStep \cite{singh_twostep_2024} decompose goals for coordination, but remain limited in the number of agents and do not explicitly target temporal parallelism.

Our method differs in representation and execution model: we generate a two-layer Prolog \kbase (high-level abstractions mapped to low-level actions) and plan with durative actions followed by a scheduling step that exploits parallelism over multiple agents.

\paragraph{Task Planning}
Classical planning formalizes problem solving as searching over sequences of instantaneous actions with logical preconditions and effects, as introduced in STRIPS and standardized in PDDL \cite{fikes_strips_1971,ghallab_automated_2004,mcdermott_pddl_1998}. Plans may be fully ordered or partially ordered, with the latter exposing concurrency opportunities \cite{penberthy_ucpop_1992,coles_forward-chaining_2010,barrett1994partial}. Temporal planning generalizes this model with durative actions and temporally-scoped conditions/effects (PDDL~2.1) \cite{fox_pddl21_2003}. Temporal constraints are commonly represented with Simple Temporal Networks, enabling efficient consistency checking and dispatchable execution, and clarifying the tight coupling between temporal planning and scheduling \cite{dechter_temporal_1991,tsamardinos_fast_1998,smith_bridging_2000}. We remark that, although the PDDL language is highly-expressive, the support of several features by state-of-the-art planners is rather limited (e.g., disjunctive or existential preconditions are not supported in many classical planners, and not supported in temporal planners).

Hierarchical approaches refine abstract tasks into lower-level actions (HTN planning and variants), sometimes allowing partial ordering within decompositions \cite{erol_htn_1996,nau_shop2_2003,patra_deliberative_2021}. Our mapped temporal task planning formulation (Section~\ref{ssec:bg_task-planning}) is closest in spirit to hierarchical refinement, but we do not propose a new planning formalism; instead, we integrate LLM-based \kb generation with symbolic reasoning, temporal planning, and scheduling within an existing framework.

%\subsubsection{Action Model Learning}
\paragraph{Action model learning} 
The goal is to reduce manual engineering by inferring preconditions and effects from data, historically from execution traces under varying observability assumptions \cite{yang_learning_2007,amir_learning_2008}. Later work increases robustness and expressiveness (e.g., learning from action sequences without predefined predicates, richer relational models, or compiling learning as a planning problem) \cite{cresswell_acquiring_2013,zhuo_learning_2010,aineto_learning_2019}. Recent directions address noisy, partial, and online settings, including conservative/safe learning and interactive data collection \cite{juba_safe_2021,lamanna_online_2021,le_learning_2024,lamanna_lifted_2025}.

These methods are powerful when informative traces are available, but they often assume a predefined symbolic vocabulary or sufficient observational coverage. Complementary work shows that symbolic domains can be recovered from state-space structure \cite{bonet_learning_2020}, and that LLMs can rapidly propose candidate domain models from language \cite{gestrin_nl2plan_2024}. Our approach leverages the logical structure of a Prolog \kbase to support internal consistency checks: rather than treating LLM output purely as a hypothesis requiring external validation, we exploit Prolog inference to verify \kb correctness.

\paragraph{LLMs for Planning and Execution}
LLMs have been explored as high-level planners, but their unreliability in multi-step reasoning and grounding motivates hybrid designs \cite{pmlr-huang_language_2022-huang22a,valmeekam_large_2022}. Representative hybrids constrain language-generated plans through affordances or programmatic interfaces (e.g., SayCan, ProgPrompt, Code as Policies) \cite{ichter_as_2022,singh_progprompt_2022,Lia23}, and incorporate feedback from execution or perception to refine plans and knowledge \cite{ding_integrating_2023,liu_reflect_2023,ha_scaling_2023}. Other frameworks explicitly map high-level language plans onto robot skill libraries to ensure executability, sometimes integrating task-and-motion planning \cite{ding_task_2023,lin_text2motion_2023,li_interactive_2025}. Overall, this literature underscores the value of combining language understanding with structured representations, verification, and planning/scheduling back-ends \cite{kambhampati_position_2024}.

\paragraph{End-to-End Vision--Language--Action Models}
End-to-end vision--language--action models directly map multimodal observations and instructions to actions (e.g., RT-2 and OpenVLA) \cite{zitkovich_rt2_2023,kim_openvla_2024}. While promising, recent evaluations report sensitivity to distribution shift (e.g., visual perturbations and language rephrasing), suggesting that structured reasoning and verification remain important for robustness \cite{Zhang_2025_ICCV,Fei_2026_CVPR,morgan_colosseumv2_2026}.

%% file: sections/3-background.tex
In this section, we provide the background concepts needed as support for our contributions.

\paragraph{Simple Temporal Network}
% In our approach, we use Simple Temporal Networks (STNs)~\cite{dechter_temporal_1991,ghallab_automated_2004} to represent Simple Temporal Problems (STPs). 
A Simple Temporal Problem (STP)\cite{dechter_temporal_1991} is defined by a set of real-valued time point variables $X=\{t_1,t_2,\ldots,t_n\}$, with $t_i\in\mathbb{R}$, and a set of temporal constraints of the following forms:
a) \emph{unary constraints} $\ell_i \leq t_i \leq u_i$ to bound the absolute value of a time point, where $\ell_i,u_i\in\mathbb{R}$ are lower and upper bounds on the timestamp. Binary constraints bound the temporal distance between two time points,
and, b) \emph{binary constraints} $\ell_{ij} \leq t_j - t_i \leq u_{ij}$ such that $\ell_{ij},u_{ij}\in\mathbb{R}$ and $\ell_{ij} \le u_{ij}$, suggesting lower and upper bounds for the pair of time points $(t_i,t_j)$. 
An STP is \emph{consistent} if the set of unary and binary constraints is satisfiable.

The consistency of an STP can be checked efficiently by converting the STP into a Simple Temporal Network (STN)~\cite{dechter_temporal_1991,ghallab_automated_2004} for which efficient algorithms exist based on checking whether the induced directed graph contains no negative-weight cycle.

%\subsection{Task Planning}
\label{ssec:bg_task-planning}

\paragraph{Temporal task planning}
\noindent Following \cite{cushing_when_2007,coles_managing_2009,coles_colin_2012}, a \emph{temporal planning problem} is defined as a tuple $TP = (F, DA, I, G)$, where $F$ is the set of (ground) fluents, $I \subseteq F$ is the initial state, $G \subseteq F$ is the goal condition, and $DA$ is a set of \emph{durative actions}. Intuitively, a fluent is a ground predicate expressing a condition (e.g., on the system's state) that can evolve over time as a consequence of actions (e.g., {\tt at(robot, RoomA)}), and a \emph{state} is a set of ground fluents. A \emph{literal} is a fluent or its negation. Predicates can be written in lifted form, e.g., $l(x)$, where variables $x$ range over objects; a \emph{ground} fluent is obtained by substituting variables with objects (e.g., \verb|available(A)| vs. \verb|available(a1)|). We distinguish between action schemata and their ground instances; an action schema is instantiated by substituting its variables with objects, yielding a ground action.

A durative action $\alpha \in DA$ is defined by: (i) \emph{start and end snap actions}, $\aStart{\alpha}$ and $\aEnd{\alpha}$; (ii) conditions that must hold at the start, throughout the entire duration ($\inv{\alpha}$), and at the end, as well as effects at the start and at the end; and (iii) a duration $\delta(\alpha)\in\mathbb{R}^+$ within the interval $[\delta_{min}(\alpha), \delta_{max}(\alpha)]$.

A start (resp. end) snap action $\aStart{\alpha}$ (resp. $\aEnd{\alpha}$) can be seen as a classical planning action~\cite{fikes_strips_1971} (with no duration) capturing what should hold before (condition at start, resp. end) and happen immediately after the durative action starts (resp. ends).
Each snap action schema has a set of preconditions $\pc{a}$ (we allow both positive and negative preconditions) and effects $\eff{a}$. Effects are specified by an \emph{add set} $\add{a}$ and a \emph{delete set} $\del{a}$ of atoms.
A ground snap action $a$ is \emph{executable} in a state $S$ if all positive preconditions are in $S$ and all negative preconditions are not in $S$. Once executed, the state is updated as $S' = (S \setminus \del{a}) \cup \add{a}$.
An $\aEnd{\alpha}$ has an implicit precondition that the respective $\aStart{\alpha}$ should have been executed before and an implicit effect that enables $\aStart{\alpha}$ to be applicable afterwards if needed.

We will consistently use Greek letters (e.g., $\alpha$) with reference to durative actions and Latin letters (e.g., $a$) with reference to snap actions.

A \emph{temporal plan}, represented here in time-triggered form, is a set $\pi = \{tta_1,\ldots,tta_n\}$ of time-triggered ground actions $tta_i = \langle t_i,\alpha_i,\delta_i\rangle$, where $t_i$ is the start time, $\alpha_i$ is a durative action, and $\delta_i$ is its duration. $\pi$ is \emph{valid} if there exists an assignment of start times and durations such that all start/overall/end conditions are satisfied, effects are applied at the correct time points, and the resulting state satisfies the goal $G$~\cite{benton_temporal_2012,cushing_when_2007,coles_managing_2009,coles_colin_2012}.

An effective approach to find a temporal plan, is state-space temporal planning search. The intuition behind this approach is to combine
\begin{enumerate*}[label=(\roman*)]
\item a classical state-space search to generate a candidate classical plan in terms of the snap action; and 
\item a temporal reasoner to check its temporal feasibility~\cite{cushing_when_2007,coles_managing_2009,coles_colin_2012}.
\end{enumerate*}
By considering the durative actions $\alpha\in DA$ as the start and end snap actions, one can generate an abstract classical planning problem \cite{fikes_strips_1971}\footnote{Classical (STRIPS) task planning can be seen as the special case where all actions are \emph{snap actions} (null duration). Formally, a \emph{(STRIPS) classical planning problem} is a tuple $CP = (F, A, I, G)$ with the same $F$, $I$, and $G$ as above and with $A$ a finite set of snap actions (or action schemata). A \emph{classical plan} $\pi = (a_1, \dots, a_n)$ is a sequence of (ground) actions. It is said to be \emph{valid} if and only if it is executable from the initial state and results in a final state that satisfies the goal $G$~\cite{ghallab_automated_2004}.}, which is then solved using any state-space search. The search extracts a classical plan solution to the planning problem and then checks if the associated temporal network is consistent (each snap action in the classical plan consists of a time point; duration constraints are imposed among $\aEnd{\alpha}$ and $\aStart{\alpha}$ time points, together with precedence constraints). If the temporal network is consistent, then a time-triggered plan has been found, and the search stops. Otherwise, the search continues by computing another classical plan until a temporally consistent plan is found or the problem is shown to have no solution. 

\paragraph{Mapped Temporal Task Planning}
\noindent Our approach uses a hierarchical temporal formulation that we call \textit{Mapped Temporal Task Planning} (MTTP). This is not intended as a new planning formalism; rather, it is a compact notation for the hierarchical structure we implement, and it is closely related to hierarchical planning models (e.g., HTN) where high-level tasks are decomposed into lower-level actions~\cite{ghallab_automated_2004,patra_deliberative_2021}. The problem is characterised by the tuple $TP=(F, DA, I, G, K, M)$, where:
\begin{itemize}
    \item $F$ is the set of all fluents;
    \item $DA = DA_H \cup DA_L$ is the set of durative actions, distinguished in high-level ($DA_H$) and low-level temporal actions ($DA_L$), with $DA_L \cap DA_H = \emptyset$;
    \item $I\subseteq F$ is the initial state;
    \item $G\subseteq F$ is the final state;
    \item $K\subseteq F$ represents the set of \emph{static} (grounding) predicates that do not change during execution;
    \item $M \subseteq DA_H \times \bigtimes_{i=1..N} DA_L^i$ with $N \ge 1$ denotes the set of mappings.
\end{itemize}

This formulation differs from standard temporal planning in three key aspects:

\begin{enumerate}
    \item The set $K\subseteq F$, which encapsulates \textit{grounding knowledge}, i.e., information assumed to remain invariant during plan execution (analogous to static predicates in PDDL). For instance, a predicate such as \verb|has_manipulator(agent1)| signifies that \verb|agent1| is endowed with a manipulator.
    
    \item The distinction between high-level actions ($DA_H$) and low-level actions ($DA_L$). Conceptually, high-level actions represent abstract tasks (e.g., \verb|moveBlock(BlockA, LocA, LocB)|, whereas low-level actions encode the concrete steps required to execute the tasks (e.g., \verb|grasp(Arm1, Block)|).
    
    \item The mappings $M \subseteq DA_H \times \bigtimes_{i=1..N} DA_L^i$
    which associate each high-level action with a (finite) sequence of low-level actions.
    A high-level durative action $\alpha$ is said to be an \textit{abstraction} of a sequence $\langle \alpha_1, ..., \alpha_N \rangle$ if the following property holds: If $\alpha \in DA_H$ is executable in a state $s_s$ and results in a state $s_d$, then for any states $s_s'$ and $s_d'$ such that $s_s \subseteq s_s'$ and $s_d \subseteq s_d'$, the sequence $\langle \alpha_1, ..., \alpha_N \rangle$ is executable in $s_s'$ and leads to $s_d'$.
    For any element $m = (\alpha, \langle \alpha_1, ..., \alpha_N \rangle) \in M$, we denote by $m(\alpha)$ the sequence $\langle \alpha_1, ..., \alpha_N \rangle$.
\end{enumerate}
We remark that this is a restricted form of hierarchical temporal planning~\cite{DBLP:journals/corr/abs-2306-07353} with only two hierarchical levels, namely high-level and low-level actions, and replaces general decomposition methods with an explicit mapping from each high-level action to its corresponding low-level refinement.

As in standard temporal planning, a feasible plan for mapped temporal task planning consists of a time-triggered plan such that the goal state is reached from the initial state under a consistent temporal schedule.

% Finally, as part of the set of fluents ($F$), we also define a set of resources $R\subseteq F$, which in this work describe the agents that can carry out a task. This will allow us later to set up a scheduling optimization problem to parallelize tasks for multiple agents and reduce the makespan of the plan.

To simplify notation, we will henceforth use the term ``action'' to refer to ``durative actions,'' unless explicitly stated otherwise.

%\subsection{Prolog}
\paragraph{Prolog}
Prolog is a logic-based programming language whose inference engine acts both as a knowledge representation layer and an execution mechanism. Unlike PDDL-based representations, this lets us query the \kb directly to evaluate action preconditions and enumerate feasible planning alternatives.

Prolog has a long history in task planning~\cite{dix_planning_2010} and has attracted renewed interest when combined with natural language processing, as this integration supports human-robot interaction~\cite{bitter_natural_2010,lally_natural_2011}.

Below, we briefly review Prolog’s semantics and operational principles; for details on Prolog semantics and the SWI-Prolog implementation, we refer the reader to the SWI-Prolog documentation\footnote{\label{fn:swiprolog}\url{https://www.swi-prolog.org/}}.
A key property of Prolog is that predicates are evaluated in the order in which they appear in the \kb, making that order semantically significant. Our framework exploits this to encode multiple planning alternatives directly in the \kbase, via actions with different preconditions or multiple definitions for the same predicates. When a query is executed, the interpreter explores predicates sequentially, generating different candidate actions and state transitions. For example, if the \kb contains \verb|available(agent2)| and \verb|available(agent1)| in this order, then the query \verb|available(X)| first assigns \verb|agent2| to \verb|X|; requesting another solution returns \verb|agent1|.
Another key feature of Prolog is its support for backtracking: when a search branch fails, the interpreter automatically retraces previous choices and explores alternatives. In our framework, this mechanism is used during planning to evaluate alternative action sequences and recover from infeasible branches. The planner relies on Prolog’s inference engine to ground predicates, check action preconditions, and systematically enumerate feasible alternatives, while the search strategy itself is a breadth-first exploration of the state space. Prolog’s backtracking does not control the global search order; it is used within each search node to efficiently generate and test alternative logical assignments and action instantiations.

\paragraph{Large Language Models}
Large Language Models (LLMs) are AI models for natural language processing, typically based on transformer networks~\cite{vaswani_attention_2017}. Transformers use self-attention to weigh each word in a sequence against all others, capturing long-range dependencies and contextual relationships. LLMs are trained on massive text corpora and often contain hundreds of billions of parameters~\cite{zhao_LLM_survey_2024,brown_language_2020,hoffmann_training_2022}, enabling broad generalization across language tasks.

However, LLMs mainly exploit statistical associations rather than genuine inferential reasoning, limiting their reliability on tasks requiring multi-step logical deduction, as discussed in Section~\ref{sec:relatedwork}. Structured prompting methods such as few-shot learning~\cite{huang_fewer_2024} and Chain-of-Thought prompting~\cite{wei_chain_2022} help steer LLMs toward domain-specific output formats. Our framework uses these techniques to generate well-formed Prolog facts and rules from natural language task descriptions, while offloading logical verification and plan generation to Prolog's inference engine.

%% file: sections/4-methodology.tex
\section{Knowledge Management System}\label{sec:kb}
\input{sections/4-methodology/a-kbgen}

\section{Plan Generation}\label{sec:plangen}
In this section, we describe how \frameworkname exploits the durative actions, mappings, and fluents defined in the \kb to generate a task plan for multiple agents. Planning is executed in four steps: 
\begin{enumerate*}
    \item Generation of a total-order (TO) plan, 
    \item Extraction of a partial-order (PO) plan,
    \item Generating and optimizing the STN exploiting parallelization, and 
    \item Extraction of the final behavior tree.
\end{enumerate*}

\subsection{Total-Order Plan Generation}\label{ssec:toplangen}
\input{sections/4-methodology/d-toplangen}
\subsection{Partial-Order Plan Generation}\label{ssec:poplangen}
\input{sections/4-methodology/e-poplangen}
\subsection{Partial-Order Temporal Plan Optimization}\label{ssec:poplanopt}
\input{sections/4-methodology/f-poplanopt}
\subsection{Behavior-Tree Generation and Execution}\label{ssec:bt}
\input{sections/4-methodology/g-bt}

%% file: sections/4-methodology/a-kbgen.tex
We will first discuss the Knowledge Management System (KMS) module, which is responsible for translating natural language descriptions into a structured Prolog \kb. By separating knowledge management from planning, the KMS allows the \kb to be validated and reused across different planning instances (\textbf{R1}, \textbf{R3}). As inputs, the KMS takes the natural language description of the environment and the agents' capabilities, and converts them into a Prolog \kb using an LLM.
The \kb contains all the necessary elements to define the mapped planning problem introduced in Section~\ref{ssec:bg_task-planning}.

As mentioned before, the \kbase contains high-level and low-level predicates. For this reason, the input descriptions are also split into \HL and \LL prompts. The former captures more abstract concepts (e.g., complex actions such as \verb|move_block| or objects in the environment). The latter captures more concrete aspects (e.g., the actions that can be carried out by the agents, such as \verb|move_arm| or actuator-specific predicates). An example of this division can be seen in Section~\ref{sssec:runegKMS}.

\begin{figure}
    \centering
\begin{minted}[fontsize=\scriptsize,breaklines]{Prolog}
hl_d_action(action_name(args),   % #$\alpha$%
    [preconditions_at_start],    % #$\pc{\aStart{\alpha}}$%
    [preconditions_at_end],      % #$\pc{\aEnd{\alpha}}$%
    [overalls],                  % #$\inv{\alpha}$%
    [effects_at_start],          % #$\eff{\aStart{\alpha}}$%
    [effects_at_end],            % #$\eff{\aEnd{\alpha}}$%
    [min_duration, max_duration] % #$(\delta_{min}(\alpha),\delta_{max}(\alpha))$%
).
\end{minted}
    \caption{An example of how a \HL durative action is encoded in the \kb.}
    \label{fig:hlDurativeActionExample}
\end{figure}

The \kbase is divided into the following parts:
\begin{itemize}
    \item General \kb ($K$): contains the grounding predicates, both for the \HL and \LL. These predicates describe parts of the scenario or of the environment that do not change during execution. For example, the predicate \verb|wheeled(a1)|, which states that robot \verb|a1| has wheels, should be part of the general \kb and not of the states. 
    
    \item Initial ($I$) state and final states ($G$): contain all the fluents that may change during the execution of the plan. This could be, for example, the position of objects in the environment. 
    
    \item High-level and low-level durative actions ($DA_H$ and $DA_L$): high-level actions are encoded as \verb|hl_d_action| predicates, as shown in Figure~\ref{fig:hlDurativeActionExample}, and capture abstract tasks whose preconditions may reference both \kb predicates and native Prolog predicates, enabling complex computations where needed. Low-level actions share the same structure but are encoded as \verb|ll_d_action| predicates and capture concrete, agent-specific behaviors such as individual arm movements or actuator commands.
    
    \item Mappings ($M$): contains a dictionary of \HL actions $DA_H$ and how they should be mapped to a sequence of \LL actions $DA_L$. As will become apparent in the following, the distinction between \HL and \LL actions induces a significant simplification in the planning phase.
\end{itemize}

The \kb generation is a multi-step iterative process comprising \emph{domain validation}, \emph{\HL \kb generation}, and \emph{\LL \kb generation} with interleaved consistency checks, each described in detail below

We would like to highlight that we employ the LLM using few-shot prompting and provide examples to the LLM to guide its ``reasoning''. This process improves adherence to the expected structure, but it does not guarantee correctness; therefore, the resulting Prolog is still iteratively validated and, when needed, corrected by the user. 

\subsection{Prompt Validation}
\label{ssec:KmsDomainValidation}
As a first step, the user provides the natural language descriptions for the \HL and \LL parts, and the \frameworkname validates the two descriptions to ensure the absence of conflicts. 
Specifically, it verifies that: (i) both descriptions share the same goal; (ii) objects remain consistent across \HL and \LL; and (iii) agents are capable of executing the required tasks.
 At this step, the LLM highlights potential inconsistencies, and since correctness cannot be guaranteed automatically, a human developer confirms or edits the output before proceeding. For instance, if a task requires an agent with a gripper but the \LL description contains no mention of robotic arms, a warning is generated.

\subsection{Knowledge-Base Generation}
\label{ssec:KmsKbGeneration}
After the descriptions have been validated, the framework moves to generate the \kb. For this step, an LLM is used to generate the required Prolog code and particular attention should be placed on the examples passed to the LLM at this time. Indeed, as we have mentioned before, the \kbase is highly structured, and the planner expects to have the different components written correctly. Few-shot prompting enables the LLM to know these details. 

In PLANTOR, \kb generation is performed sequentially, by decomposing the output into different components. For the high-level layer, the LLM is first asked to generate the general \kb $K$, then the initial and goal states $I$ and $G$, and finally the set of high-level durative actions $DA_H$. At each step, the previously generated components are provided back to the model, so that the next component can be generated consistently with the partial \kb already constructed.

The same strategy is then applied to the low-level layer. The previously generated high-level \kb is provided together with the low-level natural-language description, and the LLM is asked to extend the \kb with low-level static predicates, low-level initial and goal facts, low-level durative actions $DA_L$, and the mappings $M$ from high-level actions to low-level action sequences. This staged generation process reduces the burden on the model at each call and makes the intermediate artifacts easier to inspect and validate.

\subsection{Iterative Generation with Consistency Checks}
\label{ssec:KmsConsistencyChecks}

The generation of the \kb is an iterative process not only because it is executed in multiple stages, but also because each generated representation is validated before the pipeline proceeds. Whenever the LLM generates either the \HL or \LL \kb, a series of Prolog-based consistency checks is executed on the resulting representation. These checks can be applied independently to the \HL layer, the \LL layer, or to the complete \kb once both layers have been generated.

The checks identify as many violations as possible in a single pass, providing richer diagnostic feedback to the LLM. The resulting diagnostic messages are then provided to the LLM as feedback for subsequent repair or regeneration attempts. This 
validation-and-repair cycle can be repeated multiple times, with the intent of allowing the model to progressively correct inconsistencies and improve the generated \kb.

While additional iterations increase generation cost and latency, they also provide richer feedback and generally improve the structural quality and validity of the final domain model, as shown in Section~\ref{sssec:expKBGeneration}.

The checks focus on structural properties that are easy for the LLM to violate but that are required by the planner. For a list of checks that are currently implemented, please refer to Table~\ref{tab:kms-consistency-checks}. These checks do not guarantee semantic correctness with respect to the user's intended scenario, but they enforce the syntactic and structural assumptions required by the planner and the hierarchical representation (\textbf{R1}). Notably, constraints such as \HL to \LL mapping consistency, abstraction-layer separation, and static/dynamic predicate separation are naturally expressible as Prolog queries but cannot be captured directly in standard PDDL domain models, where parsers and validators focus on syntax, typing, and plan validity. By exploiting the Prolog \kb to detect these violations before planning commences, the framework identifies modeling inconsistencies earlier and provides targeted, human-readable diagnostics, making failures more informative than discovering them only at planning time (\textbf{R2}).

For example, when a mapping refers to a low-level action that was never defined, the checks report a message such as:
\begin{textboxerror}
\footnotesize
\texttt{Mapped actions not declared as low-level actions: [...]}
\end{textboxerror}
This feedback is deliberately aligned with the structure of the generated \kb: it identifies the violated consistency class and directly points to the problematic action or predicate.

\subsection{Running Example -- KMS}
\label{sssec:runegKMS}

We now introduce a running example, which will be used throughout this work to expose the interplay between the different components of the framework. 
This scenario is taken from the Blocks World domain~\cite{gupta_complexity_1992}, which is frequently used in task planning. In particular, in this scenario, we consider a table and some blocks, which may either be directly on the table or stacked on top of each other, as well as robotic arms, which move the blocks around. Each block is also associated with a position in the 2D space. 
In this particular example, we start from a situation in which we have three blocks, \verb|b1|, \verb|b2| and \verb|b3|, with \verb|b1| and \verb|b2| on the table in positions (1,3) and (3,1), respectively, and \verb|b3| placed on top of \verb|b1|. The goal is to move \verb|b1| to position (2,2), \verb|b2| in position (4,4) and \verb|b3| in position (1,1), all three on the table. An iconography of the example is shown in Figure~\ref{fig:running-example}.

\begin{figure*}[t]
    \centering
    \resizebox{0.95\textwidth}{!}{\input{figures/running-example}}
    \caption{A scheme representing the running example. Three blocks must be moved from their initial positions to new positions scattered on a table.}
    \label{fig:running-example}
\end{figure*}

While this is a trivial example, it highlights very well the capability of the KMS to generate complex predicates that can be used for planning, and it also shows the cooperative abilities of the framework. Indeed, even though using a single robotic arm yields a straightforward plan, coordinating multiple agents to achieve the common goal 
increases the planning complexity significantly.

We now focus on the \kbase generation. As an example, consider the two queries shown in Figure~\ref{fig:RunningExampleQuery}. These are used as query inputs to the KMS module, and hence to the whole \frameworkname framework.

\begin{figure*}[t]
\centering
\begin{minipage}{0.49\linewidth}
\begin{textbox}{High-level query}
\footnotesize
In this scenario, there are three blocks and two robotic agents. The blocks are initially placed as such: b1 is in position (1,3), b2 is in position (3,1) and b3 is on top of b1. The robotic agents can move the blocks from one position to another. In particular, each robotic arm can:
\begin{itemize}[leftmargin=2em]
    \item move a block from a position on the table, to another position on the table;
    \item move a block from a position on the table, to the top of another block;
    \item move a block from the top of a block, to a position on the table;
    \item move a block from the top of a block, to the top of another block.
\end{itemize}
Each action has a duration between 1 and 20 time units.
The goal of this scenario is to move b1 to position (2,2), b2 in position (4,4) and b3 in position (1,1) on the table. 
\end{textbox}
\end{minipage}
\hfill
\begin{minipage}{0.49\linewidth}
\begin{textbox}{Low-level query}
\footnotesize
In this scenario, there are three blocks that are in the same positions as described in the high-level query. The robotic agents are robotic arms and their end-effector is a two-finger gripper that can grasp objects. 
In particular, the following APIs are made available for the agents to move:
\begin{itemize}[leftmargin=2em]
    \item move arm(Arm, X1, Y1, X2, Y2), which allows the end-effector to be moved from one position to another;
    \item grip(Arm), which allows the gripper to be lowered onto a block and grip the block;
    \item release(Arm), which allows for lowering the block and release the gripper leaving the block in place
\end{itemize}
The grippers initially start from position (2,2) and (4,4), respectively. We do not care about their position at the end.
\end{textbox}
\end{minipage}
\caption{An example of a \HL and a \LL query.}
\label{fig:RunningExampleQuery}
\end{figure*}

As mentioned before, the first step is to do \emph{domain validation} on the input queries. In this case, when using GPT 5.2, both queries pass the test. For example, if we had changed the \HL query by removing the possibility of moving one block from on-top of other to the table, then the LLM would return the following error:

\begin{textboxerror}
\footnotesize
The agent’s allowed actions do not include moving a block from one block to a position on the table, which is required to place the block b3 in its final position.
\end{textboxerror}

\noindent Since the messages are generated in natural language, domain validation also allows the user to understand where the mistake is and correct it. 

Once the validation part succeeds, the framework moves on to the generation of the \HL \kb. In this example, we are using GPT5.2 with 3 iterations over the consistency checks. The final \kb generated by the model is shown in Figure~\ref{fig:hlkbrunningexample}. For this particular example, we present only the general \kbase ($K$), the initial ($I$) and final ($G$) states, and a single action out of the set of durative \HL actions $DA_H$ due to space limitations. 

\begin{figure*}[t]
    \centering
\begin{minipage}{\linewidth}
    \begin{minipage}{.25\linewidth}
        \begin{codebox}{prolog}{General KB}
pos(1,1).
pos(1,3).
pos(2,2).
pos(3,1).
pos(4,4).
block(b1).
block(b2).
block(b3).
agent(a1).
agent(a2).
        \end{codebox}
    \end{minipage}
    \hfill
    \begin{minipage}{.72\linewidth}
        \begin{minipage}{\linewidth}
        \begin{codebox}{prolog}{Initial state ($I$)}
init_state([
  ontable(b1), ontable(b2), on(b3, b1), at(b1,1,3), 
  at(b2,3,1), at(b3,1,3), clear(b2), clear(b3), 
  available(a1), available(a2))].        
        \end{codebox}
    \end{minipage}
        \\
    \begin{minipage}{\linewidth}
        \begin{codebox}{prolog}{Final state ($G$)}
goal_state([
  ontable(b1), ontable(b2), ontable(b3), at(b1,2,2),
  at(b2,4,4), at(b3,1,1), clear(b1), clear(b2),
  clear(b3), available(a1), available(a2)]).
        \end{codebox}
        \end{minipage}
    \end{minipage}
\end{minipage}
\vspace{-0.3cm}
\begin{codebox}{prolog}{Action example}
hl_d_action(
  move_table_to_table(Agent, Block, X1, Y1, X2, Y2),
  [ontable(Block), at(Block, X1, Y1), clear(Block), available(Agent), 
   agent(Agent), pos(X1,Y1), pos(X2,Y2), block(Block)],
  [neg(at(_, X2, Y2))],
  [],
  [del(available(Agent)), del(clear(Block)),
   del(ontable(Block)), del(at(Block, X1, Y1))],
  [add(available(Agent)), add(clear(Block)),
   add(ontable(Block)), add(at(Block, X2, Y2))],
  [1, 20]).
\end{codebox}
\caption{The results of running the KMS to generate the \HL \kb. Here are shown $K, I, G$ and one action of the set of durative \HL actions $DA_H$.}
\label{fig:hlkbrunningexample}
\end{figure*}

The resulting Prolog \HL \kb is human-readable as it is composed of truth predicates (in fulfilment of requirement \textbf{R2}).
The user at this point can make corrections to the first generated \kb, if needed, and finally, \frameworkname will also generate the \LL \kbase. In this case for space limitation, we show the changes made to the previous elements, one low-level action, and one mapping, Figure~\ref{fig:llkbrunningexample}. 

\begin{figure*}[t]
\centering
\begin{minipage}{\linewidth}
    \begin{minipage}{.27\linewidth}
        \begin{codebox}{prolog}{General KB}
pos(1,1).
pos(1,3).
pos(2,2).
pos(3,1).
pos(4,4).
block(b1).
block(b2).
block(b3).
agent(a1).
agent(a2).
ll_arm(a1).
ll_arm(a2).
ll_gripper(a1).
ll_gripper(a2).
        \end{codebox}
    \end{minipage}
    \hfill
    \begin{minipage}{.72\linewidth}
        \begin{minipage}{\linewidth}
        \begin{codebox}{prolog}{Initial state ($I$)}
init_state([
  ontable(b1), ontable(b2), on(b3, b1), at(b1,1,3),
  at(b2,3,1), at(b3,1,3), clear(b2), clear(b3),
  available(a1), available(a2), ll_arm_at(a1,2,2),
  ll_arm_at(a2,4,4), ll_gripper_state(a1, open),
  ll_gripper_state(a2, open)]).
        \end{codebox}
        \end{minipage}
        \\
        \begin{minipage}{\linewidth}
        \begin{codebox}{prolog}{Final state ($G$)}
goal_state([
  ontable(b1), ontable(b2), ontable(b3), at(b1,2,2),
  at(b2,4,4), at(b3,1,1), clear(b1), clear(b2),
  clear(b3), available(a1), available(a2), 
  ll_arm_at(a1,_,_), ll_arm_at(a2,_,_),
  ll_gripper_state(a1,_), ll_gripper_state(a2,_)]).
        \end{codebox}
        \end{minipage}
    \end{minipage}
\end{minipage}
\vspace{-0.3cm}
\begin{codebox}{prolog}{Action example}
ll_d_action(ll_move_arm(Arm, X1, Y1, X2, Y2),
  [ll_arm(Arm), ll_arm_at(Arm, X1, Y1)],
  [neg(ll_arm_at(_, X2, Y2))],
  [],
  [del(ll_arm_at(Arm, X1, Y1))],
  [add(ll_arm_at(Arm, X2, Y2))],
  [1, 2]).
\end{codebox}
\vspace{-0.35cm}
\begin{codebox}{prolog}{Mapping example}
mapping(
  move_table_to_table(Agent, _Block, X1, Y1, X2, Y2),
  [ll_move_arm(Agent, _, _, X1, Y1),
   ll_grip(Agent),
   ll_move_arm(Agent, X1, Y1, X2, Y2),
   ll_release(Agent)]).
\end{codebox}
\caption{The results of running the KMS to generate the \LL \kb starting from the \HL \kb of Figure~\ref{fig:hlkbrunningexample}. Here are shown the update $K, I, G$, one action of the set of durative \LL actions $DA_L$ and one mapping.}
\label{fig:llkbrunningexample}
\end{figure*}

Again, the user can correct possible errors (or refine the \kb) and then move on to the planning phase.

%% file: figures/running-example.tex
% \documentclass[tikz,border=2mm]{standalone}
% \begin{document}
% \usetikzlibrary{arrows.meta,calc}

\begin{tikzpicture}[
    x={(0.75cm,0.25cm)},
    y={(-0.55cm,0.35cm)},
    z={(0cm,0.75cm)},
    blockedge/.style={draw=black, thick},
    table/.style={draw=black!60, fill=gray!5},
    gridline/.style={draw=black!25},
    >=Latex
]
\footnotesize

\def\s{0.55}

\newcommand{\drawblock}[5]{%
    % #1 x, #2 y, #3 z, #4 label, #5 color

    % front face
    \draw[blockedge, fill=#5]
        (#1,#2,#3) --
        ++(\s,0,0) --
        ++(0,0,\s) --
        ++(-\s,0,0) -- cycle;

    % side face
    \draw[blockedge, fill=#5]
        (#1,#2,#3) --
        ++(0,\s,0) --
        ++(0,0,\s) --
        ++(0,-\s,0) -- cycle;

    % top face
    \draw[blockedge, fill=#5]
        (#1,#2,#3+\s) --
        ++(\s,0,0) --
        ++(0,\s,0) --
        ++(-\s,0,0) -- cycle;

    \node at (#1+0.25,#2,#3+0.2) {\scriptsize\texttt{#4}}; % This cannot be changed to \verb
}

\newcommand{\drawtable}{%
    \draw[table]
        (0.5,0.5,0) --
        (4.5,0.5,0) --
        (4.5,4.5,0) --
        (0.5,4.5,0) -- cycle;

    \foreach \i in {1,2,3,4} {
        \draw[gridline] (\i,0.5,0.01) -- (\i,4.5,0.01);
        \draw[gridline] (0.5,\i,0.01) -- (4.5,\i,0.01);
    }

    \foreach \i in {1,2,3,4} {
        \node[anchor=north] at (\i,0.25,0) {\i};
        \node[anchor=east]  at (0.25,\i,0) {\i};
    }

    \node[anchor=north] at (2.5,-0.35,0) {$x$};
    \node[anchor=east]  at (-0.45,2.,0) {$y$};
}

% ---------------- Initial state ----------------
\begin{scope}
    \begin{scope}[local bounding box=initialTable]
        \drawtable
    \end{scope}

    \node[anchor=south] at ([yshift=5mm]initialTable.north)
        {\textbf{Initial state}};

    \drawblock{0.725}{2.725}{0.02}{b1}{orange!75}
    \drawblock{0.725}{2.725}{0.57}{b3}{green!60}
    \drawblock{2.725}{0.725}{0.02}{b2}{cyan!50}
\end{scope}

% ---------------- Goal state ----------------
\begin{scope}[xshift=7.4cm]
    \begin{scope}[local bounding box=goalTable]
        \drawtable
    \end{scope}

    \node[anchor=south] at ([yshift=5mm,xshift=-2mm]goalTable.north)
        {\textbf{Goal state}};

    \drawblock{1.725}{1.725}{0.02}{b1}{orange!75}
    \drawblock{3.725}{3.725}{0.02}{b2}{cyan!50}
    \drawblock{0.725}{0.725}{0.02}{b3}{green!60}
\end{scope}

% ---------------- Arrow ----------------
\draw[->, very thick]
    (5,0.5,0.05) -- ++(1.2cm,0);
% \node at ($(5,1.5,0.05)+(1cm,0)$) {Plan};

\end{tikzpicture}

% \end{document}

%% file: sections/4-methodology/d-toplangen.tex
A total-order (TO) plan $\pi$ is a strictly sequential list of snap actions that drives the system from the initial state $I$ to a goal state in $G$. 

The algorithm used to extract a total-order plan is shown in Algorithm~\ref{alg:toplanning} in~\ref{ssec:app_toPlanAlgo} and consists of two sequential steps:
\begin{itemize}
    \item Identify a total-order plan $\pi_H$ for high-level actions:
\begin{equation}
    \pi_H =\langle a_1,\ldots,a_n\rangle, a_i\in \{\aStart{\alpha}, \aEnd{\alpha}\vert \alpha\in DA_H\},
    \label{eq:toplan}
\end{equation}
    \item Find the low-level total-order plan $\pi_L$ by mapping each high-level action to a sequence of actions from the low-level:
\begin{equation}
    \pi_L =\langle a_1,\ldots,a_n\rangle, a_i\in \{\aStart{\alpha}, \aEnd{\alpha}\vert \alpha\in DA_H\cup DA_L\},
\end{equation}
\end{itemize}

The makespan optimization is addressed separately in the scheduling step (Section~\ref{ssec:poplanopt}). 

We would like to highlight that during the second step, we do not remove the high-level actions from the total-order plan because, as we will see in Section~\ref{ssec:poplangen}, they will be used to define the causal relationships between the actions of the plan. 

Prolog's inference is used to ground lifted predicates, evaluate action preconditions, and query the \kb during the search. These queries are the main way in which logical reasoning supports the planning process in our framework.

Once a high-level total-order plan is found, we apply the mappings by re-simulating the plan from the initial state. Each high-level action is expanded into its low-level action sequence. The preconditions of the \LL actions are checked during the expansion to ensure that the low-level plan remains consistent w.r.t.\ the \kb (\textbf{R1}).

The total-order plan $\pi_L$ extracted from this function is a list of snap actions that are executed in sequence, where $t(a_i)$ denotes the start time of snap action $a_i$ as defined in Section~\ref{ssec:bg_task-planning}: %% added pointer for t(a_i) %%
\begin{equation*}
    \forall i \in \{0,\hdots \vert \pi_L\vert-1\}~t(a_i)<t(a_{i+1}),
    ~a_i \in \{\aStart{\alpha}, \aEnd{\alpha}\vert \alpha \in DA_L \cup DA_H\}
\end{equation*}

The TO plan search procedure is intentionally simple. More advanced planning strategies and heuristics from the literature could be integrated, but our planner is meant as an inspectable symbolic search layer over the generated Prolog \kbase, not as a competitor to state-of-the-art planners. For the considered robotic scenarios, where plans are short, this design is sufficient and preserves transparency and explainability. Moreover, current state-of-the-art planners offer limited support for PDDL, particularly for disjunctive and quantified preconditions (e.g., in the \verb|ll_move_arm| action of Figure~\ref{fig:llkbrunningexample}), whereas such features are naturally supported by the Prolog engine.

\subsubsection{Explainability}
A key advantage of total-order planning over a Prolog \kb is that the search is fully inspectable (\textbf{R2}). During plan generation, \frameworkname exposes the explored search graph, whose nodes are symbolic states and edges are candidate snap actions. The final plan is not just a sequence of actions, but a path in this graph.

This representation also enables local explanations of failed action applicability. Given a state $S$, a high-level action $\alpha_H$, and a phase $\tau \in \{start,end\}$, the planner reuses the search-time applicability checks to determine whether the action can be executed. If not, it identifies the violated condition, which may be:
\begin{enumerate*}[label=(\roman*)]
    \item the action is unknown in the \kb;
    \item an end action is requested while the action is not executing;
    \item overall conditions are violated;
    \item start or end preconditions are unsatisfied; or
    \item the action’s effects violate invariants of currently executing actions.
\end{enumerate*}

When preconditions fail, the explanation can also list the specific predicates causing the failure (e.g., a required predicate is missing in the current state, or no valid grounding exists in the KB). This lets the user distinguish between causes such as an incorrect initial state, a missing fact in the KB, an overly restrictive action definition, or a conflict with an executing action.

The search graph also explains why applicable actions may not appear in the final plan. The framework tracks generated successors that entered the frontier but were later discarded (e.g., because another node had lower cost). Explanations thus cover not only why the chosen plan is feasible, but also why alternatives were abandoned. Since traces are expressed using predicates from \kb, they naturally form a human-readable explanation interface.

\subsubsection{Running Example -- Total-Order Plan}
\label{sssec::runegTOPlan}

Let us consider again the \kb that we generated in Section~\ref{sssec:runegKMS}. We will now show how \frameworkname extracts the TO plan.

The algorithm starts from the initial state and explores possible other states by starting from the first action in the \kb (as previously mentioned, Prolog evaluates predicates in the order in which they appear in the \kb), which in this case is the one shown in Figure~\ref{fig:hlkbrunningexample}. 

The first step is to merge the overall conditions and the preconditions, as all of them must be true in the current state for the action to be applied. Since in this case the list of overalls is empty, the preconditions that we have to check are simply the start preconditions:

\begin{minted}[fontsize=\scriptsize,breaklines]{prolog}
ontable(Block), at(Block, X1, Y1), clear(Block), available(Agent), agent(Agent), block(Block), pos(X1, Y1), pos(X2, Y2)
\end{minted}

In this list of preconditions, there are predicates that must be grounded w.r.t. the general \kb $K$, such as \verb|pos(X1, Y1)|, and predicates that must be satisfied in the current state, such as \verb|ontable(Block)|. Our planner internally differentiates between these predicates by testing which ones must be grounded to $K$ and which must be compared to the current state $S$. For instance, the predicate \verb|pos(1,1)| is a possible grounding of \verb|pos(X1,Y1)| from the general \kb, while the assignment \verb|ontable(b1)| from the current (initial) state satisfies the precondition \verb|ontable(Block)|. 

It is worth noting that not only can we compare predicates w.r.t. the \kb and the current state, but we can also execute Prolog commands. For instance, if we wanted to make sure that the block is moved from one position to a \emph{different} one, then we could append the following predicates to the list of preconditions \verb|\+(X1 = X2, Y1 = Y2)|. 

A cornerstone of Prolog that we previously mentioned is its intrinsic ability to test multiple options. In this particular instance, we can see that \verb|pos\2| has multiple possible groundings in $K$: \verb|pos(1,1), pos(2,2), pos(3,1)|. This is a fundamental characteristic of the planner as it allows it to find multiple states when applying an action based on the different groundings of the predicates. 

Once the preconditions have been verified, the planner applies the corresponding effects to generate a new state $S_i$ starting from the current state $S$. For each combination of groundings in the preconditions, a different state is added to the OPEN list of the planner. If it is already present, then the planner checks whether the new node has a lower cost than the one already present and in case, it discards this new state as there is a shorter plan that allows for reaching that state.

Generally speaking, for each current state $S$, the planner checks whether $S$ satisfies the starting preconditions of any durative action in $DA_H$ and also checks if any action can be terminated by verifying whether (only) the end preconditions are satisfied in the current state and the effects can be applied.

When the planner finishes this first step, it either reports that no plan could be found or outputs and stores the total-order \HL plan $\pi_H$, which in this case is:

\begin{minted}[fontsize=\scriptsize]{text}
[1] start(move_onblock_to_table(a2,b3,b1,1,3,1,1))
[2] start(move_table_to_table(a1,b2,3,1,4,4))
[3] end(move_onblock_to_table(a2,b3,b1,1,3,1,1))
[4] start(move_table_to_table(a2,b1,1,3,2,2))
[5] end(move_table_to_table(a1,b2,3,1,4,4))
[6] end(move_table_to_table(a2,b1,1,3,2,2))
\end{minted}

At this point, the algorithm takes the mappings and applies them to the previous plan. For instance, consider the mapping for \verb|move_table_to_table| (Figure~\ref{fig:llkbrunningexample}), and the mapping for \verb|move_block_to_table|:
\begin{minted}[fontsize=\scriptsize,]{prolog}
mapping(
  move_onblock_to_table(Agent, _Block1, _Block2, X1, Y1, X2, Y2),
  [
    ll_move_arm(Agent, _, _, X1, Y1),
    ll_grip(Agent),
    ll_move_arm(Agent, X1, Y1, X2, Y2),
    ll_release(Agent)
  ]).
\end{minted}

The planner then applies them and finds the final total-order \LL plan $\pi_L$:

\begin{minted}[fontsize=\scriptsize]{text}
[0] start(move_onblock_to_table(a2,b3,b1,1,3,1,1))
[1] start(ll_move_arm(a2,4,4,1,3))
[2] end(ll_move_arm(a2,4,4,1,3))
[3] start(ll_grip(a2))
[4] end(ll_grip(a2))
[5] start(ll_move_arm(a2,1,3,1,1))
[6] end(ll_move_arm(a2,1,3,1,1))
[7] start(ll_release(a2))
[8] end(ll_release(a2))
[9] start(move_table_to_table(a1,b2,3,1,4,4))
[10] start(ll_move_arm(a1,2,2,3,1))
[11] end(ll_move_arm(a1,2,2,3,1))
[12] start(ll_grip(a1))
[13] end(ll_grip(a1))
[14] start(ll_move_arm(a1,3,1,4,4))
[15] end(ll_move_arm(a1,3,1,4,4))
[16] start(ll_release(a1))
[17] end(ll_release(a1))
[18] end(move_onblock_to_table(a2,b3,b1,1,3,1,1))
[19] start(move_table_to_table(a2,b1,1,3,2,2))
[20] start(ll_move_arm(a2,1,1,1,3))
[21] end(ll_move_arm(a2,1,1,1,3))
[22] start(ll_grip(a2))
[23] end(ll_grip(a2))
[24] start(ll_move_arm(a2,1,3,2,2))
[25] end(ll_move_arm(a2,1,3,2,2))
[26] start(ll_release(a2))
[27] end(ll_release(a2))
[28] end(move_table_to_table(a1,b2,3,1,4,4))
[29] end(move_table_to_table(a2,b1,1,3,2,2))
\end{minted}

%% file: sections/4-methodology/e-poplangen.tex
The total-order plan $\pi_L$ extracted in the previous step is correct but overly constrained: it sequences all actions regardless of whether some could be executed in parallel. To identify opportunities for concurrent execution, we analyze $\pi_L$ in search of all causal relationships between actions, represented as \emph{enablers}.

Following the standard notion of causal support in partial-order planning~\cite{ghallab_automated_2004}, we call an earlier action $a_j$ an \emph{enabler} of $a_i$ if it establishes one of the conditions required for $a_i$ to execute. Specifically, $a_j$ may enable $a_i$ either by adding a fluent required by a positive precondition, or by deleting a fluent whose absence is required by a negative precondition:

\begin{equation}
\begin{array}{rl}
     a_j \in \ach{a_i} \iff & t(a_j) < t(a_i) \wedge \\
                            & \left((\exists l\,\vert\, l\in \pc{a_i} \wedge add(l)\in \eff{a_j}) \vee \right.\\
                            & \left.\,\,(\exists l \,\vert\, \lnot l\in \pc{a_i} \wedge del(l)\in \eff{a_j})\right)
\end{array}
\label{eq:enablers}
\end{equation}

Besides the enablers added corresponding to the classical definition, we also enforce the following \emph{hierarchical precedence constraints}:
\begin{itemize}
    \item For every high-level durative action $\alpha_i$, the termination action $\aEnd{\alpha_i}$ depends on the completion of all actions introduced by its refinement $m(\alpha_i)$. Therefore, when $\alpha_i$ is expanded into a sequence of lower-level actions, every action in the mapping is added as an enabler of $\aEnd{\alpha_i}$. For example, if $m(\alpha_i)=\{\alpha_j,\alpha_k\}$,     the resulting total-order plan is $\{\aStart{\alpha_i},\aStart{\alpha_j},\aEnd{\alpha_j},\aStart{\alpha_k},\aEnd{\alpha_k},\aEnd{\alpha_i}\}$. Since $\alpha_i$ can terminate only after all refined actions have completed, each action in the mapping induces a precedence constraint toward $\aEnd{\alpha_i}$:
    \begin{equation}
        \bigwedge_{a\in m(\alpha_i)} a\in \ach{\aEnd{\alpha_i}}.
        \label{eq:constraint5}
    \end{equation}

    \item Conversely, every action generated by the refinement of a high-level action must occur after the execution of its corresponding start action. Hence, the start action $\aStart{\alpha_i}$ is added as an enabler of all actions in the mapping $m(\alpha_i)$. Referring to the previous example, $\aStart{\alpha_i}$ is an enabler of $\aStart{\alpha_j}$, $\aEnd{\alpha_j}$, $\aStart{\alpha_k}$, and $\aEnd{\alpha_k}$. This precedence relation is formalized as:
    \begin{equation}
        \bigwedge_{a\in m(\alpha_i)} \aStart{\alpha_i} \in \ach{a}.
        \label{eq:constraint4}
    \end{equation}
\end{itemize}

The algorithm that manages this extraction is shown in Algorithm~\ref{alg:poplanning} in~\ref{ssec:app_enablerExtractionAlgo}. 

\subsubsection{Running Example -- Partial-Order Plan}
\label{sssec:PORunEx}
Starting from the TO plan $\pi_L$ derived in Section~\ref{sssec::runegTOPlan}, we now extract the enabler relationships that will support parallel execution. For instance, the snap action 19, \verb|start(move_table_to_table(a2,b1,1,3,2,2))|, has the predicates \verb|clear(Block2)| and \verb|available(Agent)| as preconditions (among the others), which are satisfied only once action 0 and action 18 have applied their effects, respectively. The algorithm establishes that $\{a_0,a_{18}\}$ are an enablers of $a_{19}$.

Instead, we can see that snap action 9 \verb|start(move_table_to_table(a1,b2,3,1,4,4))| does not conflict with other actions in terms of requiring actions to ``enable'' it, and hence it is independent of the other high-level moves.

After this step, the planner returns the enablers for the actions, shown in the squared brackets in Figure~\ref{fig:enablersRunningExample}:

\begin{figure}[htp]
    \centering
    \begin{minted}[fontsize=\scriptsize]{text}
[0] start(move_onblock_to_table(a2,b3,b1,1,3,1,1)) <= []
[1] start(ll_move_arm(a2,4,4,1,3)) <= [0]
[2] end(ll_move_arm(a2,4,4,1,3)) <= [0,1]
[3] start(ll_grip(a2)) <= [0,2]
[4] end(ll_grip(a2)) <= [0,3]
[5] start(ll_move_arm(a2,1,3,1,1)) <= [0,2,4]
[6] end(ll_move_arm(a2,1,3,1,1)) <= [0,5]
[7] start(ll_release(a2)) <= [0,4,6]
[8] end(ll_release(a2)) <= [0,7]
[9] start(move_table_to_table(a1,b2,3,1,4,4)) <= []
[10] start(ll_move_arm(a1,2,2,3,1)) <= [9]
[11] end(ll_move_arm(a1,2,2,3,1)) <= [9,10]
[12] start(ll_grip(a1)) <= [9,11]
[13] end(ll_grip(a1)) <= [9,12]
[14] start(ll_move_arm(a1,3,1,4,4)) <= [9,11,13]
[15] end(ll_move_arm(a1,3,1,4,4)) <= [1,9,14]
[16] start(ll_release(a1)) <= [9,13,15]
[17] end(ll_release(a1)) <= [9,16]
[18] end(move_onblock_to_table(a2,b3,b1,1,3,1,1)) <= [0,1,2,3,4,5,6,7,8]
[19] start(move_table_to_table(a2,b1,1,3,2,2)) <= [0,18]
[20] start(ll_move_arm(a2,1,1,1,3)) <= [1,6,8,19]
[21] end(ll_move_arm(a2,1,1,1,3)) <= [5,19,20]
[22] start(ll_grip(a2)) <= [8,19,21]
[23] end(ll_grip(a2)) <= [19,22]
[24] start(ll_move_arm(a2,1,3,2,2)) <= [2,19,21,23]
[25] end(ll_move_arm(a2,1,3,2,2)) <= [10,19,24]
[26] start(ll_release(a2)) <= [4,19,23,25]
[27] end(ll_release(a2)) <= [19,26]
[28] end(move_table_to_table(a1,b2,3,1,4,4)) <= [9,10,11,12,13,14,15,16,17]
[29] end(move_table_to_table(a2,b1,1,3,2,2)) <= [19,20,21,22,23,24,25,26,27]
    \end{minted}
    \caption{The result of extracting the enablers of the actions from the total-order plan $\pi_L$ for the example shown in Section~\ref{sssec:runegKMS}.}
    \label{fig:enablersRunningExample}
\end{figure}

%% file: sections/4-methodology/f-poplanopt.tex
The last part of the planning module is the temporal optimization module. Starting from the partial-order plan extracted in Section~\ref{ssec:poplangen}, the module builds a simple temporal network. In the implementation, each snap action in the partial-order plan is represented as a time-point node, while two artificial nodes, \verb|INIT| and \verb|END|, are added to anchor the beginning and the completion of the temporal plan. Causal links and ordering relations extracted from the partial-order plan are translated into temporal precedence constraints, while duration bounds are added between the corresponding start and end snap actions.

We solve the STN timestamp assignment as a linear optimization problem using a linear solver. By default, timestamps are considered real numbers, but optionally, integer timestamps can be enforced, in which case the model becomes a mixed-integer linear problem. The structure of the optimization problem is described next.

\paragraph{Decision variables}
For each individual snap action $a \in \pi_L$ we introduce a timestamp variable $t_a$. The initial node is fixed at zero:
\begin{equation}
    t_{\verb|INIT|} = 0.
\end{equation}
The timestamp of the final node, corresponding to the ending of the execution, is $t_{\verb|END|}$, hence it must be: 
\begin{equation}
    \forall a \in \pi_L, t_a \leq t_{\verb|END|}.
\end{equation}

% The implementation supports continuous time variables by default, with an optional integer-time mode when required by the selected OR-Tools backend.

\paragraph{Precedence constraints}
For each hierarchical precedence constraint or causal edge $(a_i \rightarrow a_j)$ in the partial-order plan, we impose:
\begin{equation}
    t_{a_j} \ge t_{a_i} + \epsilon,
    \label{eq:precedence}
\end{equation}
where $\epsilon > 0$ is a small minimum separation enforcing strict ordering (as common practice in temporal planning~\cite{coles_temporal_2017}); 
% in the current implementation, $\epsilon$ is set to a fixed default but can be configured by the user. 
This includes causal links derived from enablers and additional ordering constraints introduced during partial-order extraction.

\paragraph{Duration constraints}
For each durative action whose start and end snap actions are identified, the planner parses the corresponding duration bounds $[d_a^{\min}, d_a^{\max}]$ from the Prolog output and adds them:
\begin{equation}
    d_a^{\min} \le t_{a^{\text{end}}} - t_{a^{\text{start}}} \le d_a^{\max}.
    \label{eq:duration}
\end{equation}
These binary constraints are encoded as STN edges between the start and end time points of the action.

\paragraph{Concurrency model}
Actions are allowed to overlap unless the partial-order plan contains a causal or ordering relation between them, or unless their temporal bounds indirectly constrain their timestamps (\textbf{R4}). Any required serialization must therefore be captured before STN construction, through the causal and hierarchical precedence constraints extracted from the symbolic plan.

\paragraph{Objective}
The optimization minimizes the timestamp of the final anchor node:
\begin{equation}
    \min t_{\verb|END|}.
    \label{eq:end_time}
\end{equation}
Since all action nodes are constrained to occur between \verb|INIT| and \verb|END|, minimizing $t_{\verb|END|}$ corresponds to minimizing the completion time of the temporal plan.

Before running the optimization part, we perform a consistency check on the STN to ensure that the temporal constraints added to the network allow for a feasible assignment. As mentioned in Section~\ref{sec:background}, this can be done by ensuring that the network does not have negative cycles.

The solver returns a feasible timestamp assignment for the STN or reports infeasibility. In the latter case, the framework reports the conflicting constraints, allowing the user to revise the duration bounds or the underlying \kb. The resulting STN stores the optimized timestamps and provides the temporal structure used by the downstream execution representation.

\subsubsection{Partial-Order Temporal Plan Optimization -- Running Example}
\label{sssec:PORunExample}
Let's consider the enablers scenario shown in Figure~\ref{fig:enablersRunningExample}. A key observation is that action $a_{9}$ has no causal dependency from the previous actions and does not influence later different \HL actions: this task is logically independent from the other two high-level actions, involving a different block and a different agent. 

After solving the STN optimization problem, we obtain a time-triggered plan, shown in Figure~\ref{fig:runningExampleSTNOptimized}, of \LL actions that can be converted into a behavior tree and then executed~\cite{zapf_constructing_2024}. This plan improves over the previous total-order representation because it removes unnecessary sequencing constraints between independent actions. As a result, the completion time is reduced while preserving all causal dependencies and duration constraints. The optimized STN makes explicit the parallelism that was only implicit in the partial-order plan and produces a schedule with a shorter makespan.

\begin{figure*}[t]
    \centering
    \resizebox{0.8\textwidth}{!}{\input{figures/stn_optimized}}
    \caption{Optimized STN schedule showing the timestamps for the low-level actions. Independent actions are executed in parallel where causal and duration constraints allow, reducing the overall plan duration while preserving the required ordering dependencies.}
    \label{fig:runningExampleSTNOptimized}
\end{figure*}

%% file: figures/stn_optimized.tex
\begin{tikzpicture}
\begin{axis}[
    width=15cm,
    height=9cm,
    xmin=0,
    xmax=8,
    ymin=0,
    ymax=12,
    xlabel={Time},
    ylabel={Scheduled actions},
    xtick={0,1,2,3,4,5,6,7,8},
    xticklabel={\pgfmathprintnumber[fixed,precision=5]{\tick}},
    ytick={0.5,1.5,2.5,3.5,4.5,5.5,6.5,7.5,8.5,9.5,10.5,11.5},
    yticklabels={
        {$a_{26}$},
        {$a_{24}$},
        {$a_{22}$},
        {$a_{20}$},
        {$a_{16}$},
        {$a_{7}$},
        {$a_5$},
        {$a_{14}$},
        {$a_3$},
        {$a_{12}$},
        {$a_1$},
        {$a_{10}$}
        % {},
        % {$a_{24}$},
        % {$a_{22}$},
        % {$a_{20}$},
        % {$a_{16}$},
        % {$a_{7}$},
        % {$a_5$},
        % {$a_9$},
        % {$a_3$},
        % {$a_{12}$},
        % {},
        % {$a_{10}$}
    },
    grid=both,
    major grid style={draw=gray!25},
    minor grid style={draw=gray!15},
    tick align=outside,
    axis line style={draw=none},
    tick style={draw=none},
    enlarge y limits=false
]

\draw[bar] (axis cs:0,11.15) rectangle (axis cs:1,11.85);
\draw[bar] (axis cs:0,10.15) rectangle (axis cs:1,10.85);

\draw[bar] (axis cs:1,9.15) rectangle (axis cs:2,9.85);
\draw[bar] (axis cs:1,8.15) rectangle (axis cs:2,8.85);

\draw[bar] (axis cs:2,7.15) rectangle (axis cs:3,7.85);
\draw[bar] (axis cs:2,6.15) rectangle (axis cs:3,6.85);

\draw[bar] (axis cs:3,5.15) rectangle (axis cs:4,5.85);
\draw[bar] (axis cs:3,4.15) rectangle (axis cs:5,4.85);

\draw[bar] (axis cs:4,3.15) rectangle (axis cs:5,3.85);

\draw[bar] (axis cs:5,2.15) rectangle (axis cs:6,2.85);

\draw[bar] (axis cs:6,1.15) rectangle (axis cs:7,1.85);

\draw[bar] (axis cs:7,0.15) rectangle (axis cs:8,0.85);

\end{axis}
\end{tikzpicture}

%% file: sections/4-methodology/g-bt.tex
\newcommand{\seq}[0]{\protect\writings{\texttt{SEQUENCE}}} % These cannot be changed with \verb
\newcommand{\parr}[0]{\protect\writings{\texttt{PARALLEL}}}

% In this section, we first introduce how to convert from a STN to a \bt, and then we provide some details regarding the implementation. 

% \subsubsection{\bt Generation}\label{sssec:btgen}

The optimized plan is converted into a Behavior Tree, which provides the executable representation dispatched to the robotic middleware (\textbf{R5}). The conversion from STN to \bt is taken from~\cite{zapf_constructing_2024}. We summarize it here and refer the reader to the original article for a complete treatment.

% Recall that in the STN constructed in Section~\ref{ssec:poplanopt}, a node with multiple parents cannot execute until all parents have completed, while a node with multiple children triggers parallel execution. These two properties drive the structure of the extracted \bt.

A \btree \cite{colledanchise_behavior_2017} is a hierarchical structure that, starting from the root, recursively ticks its nodes until the last leaf is reached. A \emph{tick} is a signal sent to a node, requesting it to perform its function. We consider nodes of the following types:
\begin{itemize}
    \item \emph{action}: a node representing an action to be executed;
    \item \emph{control}: a node that is either a \seq or a \parr node, determining how its children are ticked, either sequentially or in parallel, respectively;
    \item \emph{condition}: a node that checks whether a condition is satisfied before allowing execution to proceed.
\end{itemize}
% In literature, there are also other kind of nodes, e.g. \emph{decorator} nodes, but all of them are not used to perform the conversion (see \cite{zapf_constructing_2024} for further details)

The algorithm converts the STN to a \bt starting from the fictitious \verb|init| node. For every node in the STN, it checks:
\begin{itemize}
    \item If it has only one child, a \seq node is introduced; otherwise, a \parr node is used.
    \item If it has more than one parent, the node must wait for all parents to have completed before being executed.
\end{itemize}

%% file: sections/5-experiments.tex
In this section, we first present the implementation details of the framework, followed by a description of the experiments conducted and the results obtained. We then discuss the scalability of \frameworkname before concluding with a final discussion on the proposed framework. The experimental evaluation has been added to a Zenodo repository~\cite{saccon_2026_21630757}.

\subsection{Implementation Details}
\label{ssec:implementation}

\input{sections/5-experiments/a-implementation_details}

\subsection{Scenarios Description}
\label{ssec:scenariosDescription}

\input{sections/5-experiments/b-scenarios_description}

\subsection{Knowledge Management System Evaluation} 
\label{ssec:KBGenValidation}
\input{sections/5-experiments/c-kb_generation}

\subsection{Planner}
\label{ssec:planRes}
\input{sections/5-experiments/d-planner}

\subsection{Real-World Experiment}
\label{ssec:realLifExperiment}
\input{sections/5-experiments/e-real_experiment}

%% file: sections/5-experiments/a-implementation_details.tex
The framework is implemented as three main software components: the knowledge-management system, the symbolic planner with temporal optimization, and the execution layer based on \bts. This separation keeps the generated \kb, the planning algorithms, and the execution representation inspectable and independently testable.

\paragraph{KMS}
As already explained, the \kb is written in Prolog to exploit its inference capabilities. This representation allows the framework to define predicates and construct queries over facts, action preconditions, effects, and mappings. The generated \kbase is loaded directly inside the planner and must therefore follow the expected Prolog structure.

The LLM models are guided using examples for few-shot prompting. They included general formatting examples for the \kb structure (Figure~\ref{fig:genExample}, in~\ref{ssec:app_examplesFewShotPrompting}) and durative actions (Figure~\ref{fig:actionExample}), together with scenario-specific examples based on Blocks World (Figure~\ref{fig:blocksExample}) and an AGV navigation setting. The few-shot examples used for \kb generation were intentionally not drawn from the Grippers domain, hence, the Grippers experiments evaluate whether the models can transfer the expected Prolog representation to a domain involving different objects, constraints, and forms of multi-agent interaction, rather than merely adapting an in-domain example. After generation, the output is parsed into the corresponding Prolog sections and checked with the consistency predicates described in Section~\ref{ssec:KmsConsistencyChecks}. 

For reproducibility, the LLM calls were configured to minimize sampling variability whenever the backend allowed it. In particular, deterministic decoding was used by setting the temperature to 0.0. For the locally executed open-weight models, we set \verb|top_p=1.0|, both frequency and presence penalties to 0.0, and the random seed to 42. The maximum number of generated tokens was set to 8192 for Qwen 3.6 and to 16384 for Llama 3.3, while Qwen was also configured with disabled thinking traces and stop sequences for user/system markers. The Claude and GPT models accessed through Azure were run with a maximum output length of 8192 tokens and temperature 0.0. 

\paragraph{Planner}
The planner is implemented using SWI-Prolog and Python 3. SWI-Prolog loads the generated \kb, grounds lifted predicates, checks action applicability, and extracts feasible total-order plans (\textbf{R1}). The same Prolog layer applies the high-level to low-level mappings and computes the enablers used to obtain the partial-order plan.

Python is used for graph construction and temporal reasoning. The partial-order plan is represented with NetworkX\footnote{\label{fn:networkx}\url{https://networkx.org/}} and converted into an STN.

The STN is solved and optimized using the OR-Tools linear solver interface$^{\ref{fn:or-tools}}$. In the default configuration, timestamps are continuous variables and the GLOP backend is used. The implemented objective minimizes the timestamp of the final \verb|END| node subject to the ordering and duration constraints encoded in the STN.

\paragraph{\Btrees}
The framework loads a \bt from an XML file and instantiates it using \verb|py_trees|\footnote{\label{fn:pytrees}\url{https://github.com/splintered-reality/py_trees}}. The current implementation supports sequence and parallel control nodes, along with custom action nodes. 
%Action nodes act as the interface between the symbolic execution structure and the robot-side execution module: when ticked, they send execution requests through a ZMQ interface (a lightweight asynchronous messaging library) and return success or failure according to the response.

This XML-based representation keeps the execution layer decoupled from the planner. It also makes the generated \bt inspectable and portable (\textbf{R2}, \textbf{R5}): the same structure can be executed by the current Python implementation or adapted to ROS2-oriented \bt engines such as PlanSys2~\cite{martin_plansys2_2021} or BehaviorTree.CPP\footnote{\label{fn:behaviortreeCPP}\url{https://github.com/BehaviorTree/BehaviorTree.CPP}}.

\paragraph{Experimental Setup}
The experiments were conducted on two domains: Blocks World and Grippers, with multiple problem instances for each. The evaluated LLMs included both remote API-served models and locally executed open-weight models.

The remote models were GPT 5.2 and GPT 5.4 Mini from the OpenAI GPT family, and Claude Opus 4.6 and Claude Sonnet 4.6 from the Anthropic Claude family, all accessed through the Microsoft Azure portal rather than the respective provider APIs directly. GPT 5.4 Mini was included to assess whether a smaller remote model could still generate usable \kbs. The open-weight models were \verb|Qwen3.6-35B-A3B|~\cite{qwen36_35b_a3b}, \verb|Mixtral-8x7B-Instruct-v0.1|~\cite{jiang2024mixtralexperts}, and \verb|Llama-3.3-70B-Instruct|~\cite{grattafiori2024llama3herdmodels}, selected to evaluate locally deployable alternatives to proprietary remote models. Qwen and Mixtral are distributed under Apache-2.0 licenses; Llama~3.3 is open-weight but released under Meta's custom Llama~3.3 Community License. In the following, we refer to these models as \textit{Qwen~3.6}, \textit{Mixtral}, and \textit{Llama~3.3}, respectively. %The evaluated models are summarized in Table~\ref{tab:models}.

% \begin{table}[t]
%     \centering
%     \small
%     \renewcommand{\arraystretch}{1.2}
%     \begin{tabular}{lll}
%         \toprule
%         \textbf{Model} & \textbf{Type} & \textbf{Backend} \\
%         \midrule
%         GPT 5.2         & Hosted      & Microsoft Azure \\
%         GPT 5.4 Mini    & Hosted      & Microsoft Azure \\
%         Claude Sonnet 4.6 & Hosted    & Microsoft Azure \\
%         Claude Opus 4.6 & Hosted      & Microsoft Azure \\
%         Llama~3.3       & Open-weight & HuggingFace \\
%         Qwen~3.6        & Open-weight & HuggingFace \\
%         Mixtral         & Open-weight & HuggingFace \\
%         \bottomrule
%     \end{tabular}
%     \caption{LLMs evaluated in the experiments.}
%     \label{tab:models}
% \end{table}

Open-weight models were run on an HPC cluster node running Rocky Linux~9.3 on an AMD Epyc~7643 CPU platform with 1~TB of system memory, using two Nvidia A100-SXM4-80GB GPUs for local HuggingFace inference. The Python environment comprised Python~3.12.3, PyTorch~2.10.0, CUDA~12.8, and Transformers~5.6.0.dev0. Remote models were accessed through the Azure backend and did not use local GPU resources.

Symbolic planning, partial-order extraction, STN construction, temporal optimization, and BT-related processing were executed locally on a PC with an AMD Ryzen 7700X, 64GB of DDR5 memory and, as software, running Python~3.9.23 and SWI-Prolog~9.2.9.

%% file: sections/5-experiments/b-scenarios_description.tex
The experimental queries cover two standard task-planning domains: Blocks World~\cite{gupta_complexity_1992} and Grippers~\cite{mcdermott_pddl_1998,mcdermott_1998_2000}. In accordance with our framework, each instance is described at two abstraction levels: the high-level (HL) query defines objects, initial and goal states, agents, and abstract durative actions, while the low-level (LL) query refines those actions into robot primitives and HL-to-LL mappings.

\paragraph{Blocks world}
The Blocks World benchmark contains six tabletop manipulation instances, summarized in Table~\ref{tab:blocks-world-scenarios},~\ref{ssec:app_BWScenarios}. The first four cover standard rearrangement and stacking tasks using the basic move templates already present in the examples: table-to-table, table-to-block, block-to-table, and block-to-block. The last two add domain-specific structure: arch construction, where a longer architrave must be placed on two supports, and a container case, where blocks must be removed from bowls before stacking. The LL descriptions model agents as robotic arms with grippers and refine block moves into arm motion and gripper close/open primitives. 

\paragraph{Grippers}
The Grippers benchmark contains six ball-transport instances between Room~A and Room~B, summarized in Table~\ref{tab:grippers-scenarios},~\ref{ssec:app_GScenarios}. The scenarios range from one robot with two grippers to two-robot variants with wide-door motion, narrow-door resource constraints, mandatory intermediate delivery, cooperative heavy-ball carrying, and doorway handoffs. These prompts test move, pick-up, drop, and handoff actions together with gripper occupancy, robot-location, door-capacity, and coordinated-action constraints. The LL descriptions refine these actions into arm poses, ground-placement poses, handoff poses, and gripper state changes.

Together, the scenarios test whether generated Prolog KBs are both syntactically valid and semantically compatible with planning, temporal-resource constraints, and HL-to-LL mappings.

%% file: sections/5-experiments/c-kb_generation.tex
The tests for the generation of the \kb are divided in three parts. A first one is for the validation check carried out at the beginning (shown in Table~\ref{tab:validationRes}), a second one for the generation of the high-level (HL) \kb (Table~\ref{tab:hlRes}), and a final one for the generation of the low-level (LL) \kb (Table~\ref{tab:llRes}). 

\subsubsection{Domain Validation} 
\label{sssec:expDomainValidation}

\newcommand{\wh}[0]{$X_{H}$}
\newcommand{\wl}[0]{$X_{L}$}

\begin{table*}[t]
    \centering
    
    \caption{Results of the domain-validation consistency check on the input queries. Each column corresponds to one validation example from the Blocks World or Grippers domains. A \cm~indicates that the model correctly classified the query as valid or invalid, identifying the inconsistency when present. \wh~indicates a failure in validating the HL prompt; in this case, the subsequent combined HL/LL validation was not considered. \wl~indicates that the HL prompt was validated correctly, but the model failed to validate the combined HL/LL prompt.}
    \label{tab:validationRes}
    
    \setlength{\tabcolsep}{4pt}
    \renewcommand{\arraystretch}{1.05}
    \footnotesize
    
    \begin{tabular}{rcccccccccccccccc}
        \toprule
                    & \multicolumn{9}{c}{Blocks World} & \multicolumn{7}{c}{Grippers} \\
        \cmidrule(lr){2-10}\cmidrule(lr){11-17}
        
        Model          &  1  & 2.a & 2.b & 3.a & 3.b &  4  & 5.a & 5.b &  6  & 1.a & 1.b &  2  &  3  &  4  &  5  &  6  \\
        \midrule
        Opus 4.6       & \cm & \cm & \cm & \cm & \cm & \cm & \cm & \cm & \cm & \cm & \cm & \cm & \cm & \cm & \cm & \cm \\
        Sonnet 4.6     & \cm & \cm & \cm & \cm & \cm & \cm & \cm & \cm & \cm & \cm & \cm & \wl & \cm & \cm & \wl & \cm \\
        GPT 5.2        & \cm & \cm & \cm & \cm & \cm & \cm & \cm & \cm & \cm & \cm & \cm & \cm & \cm & \cm & \cm & \cm \\
        GPT 5.4 Mini   & \cm & \wl & \cm & \cm & \cm & \cm & \cm & \wh & \cm & \cm & \cm & \cm & \cm & \cm & \wl & \cm \\
        Qwen 3.6       & \cm & \cm & \wh & \cm & \cm & \cm & \cm & \cm & \cm & \cm & \cm & \cm & \cm & \cm & \wl & \cm \\
        Llama 3.3      & \cm & \cm & \cm & \cm & \cm & \cm & \wh & \cm & \cm & \cm & \cm & \cm & \cm & \cm & \cm & \cm \\
        Mixtral        & \cm & \cm & \wh & \cm & \wl & \cm & \cm & \wh & \cm & \cm & \cm & \cm & \cm & \cm & \cm & \cm \\
        \bottomrule
    \end{tabular}
\end{table*}

Before generating a \kb, the model is asked to evaluate whether the scenario is consistent for a domain validation (Section~\ref{ssec:KmsDomainValidation}). The check is performed first on the HL description and then on the combined HL/LL description. This step is useful for detecting contradictory statements in the prompts.

As shown in Table~\ref{tab:validationRes}, the validation experiments also include a set of intentionally incorrect queries, introduced to test whether the model can detect inconsistent or unrealisable specifications:
\begin{itemize}
    \item Instance 2.b of the Blocks World scenario negates the expected final position of block~B4: the correct query requires B4 to be on the table at position~(10,10), while the incorrect query states that B4 is not on the table at that position.
    \item Instance 3.b of the Blocks World scenario introduces a mismatch between the high-level and low-level descriptions: the high-level query specifies two agents, whereas the incorrect low-level query provides only one arm.
    \item Instance 5.b of the Blocks World scenario changes the architrave dimensions in the arch-construction task: the correct query defines the architrave as a longer block acting as an architrave, while the incorrect query states that all blocks have size~1x1x1, not fitting the gap between the pillars.
    \item Instance 1.b of the Grippers scenario removes the low-level manipulation capabilities: the high-level query requires the robot to pick up, carry, and drop balls using its grippers, while the incorrect low-level query describes a wheeled robot with no arms or grippers, making those actions impossible to refine.
\end{itemize}

The results in Table~\ref{tab:validationRes} show that the validation step is effective for checking both the internal consistency of a scenario and the compatibility between abstraction levels. The strongest models, Opus~4.6 and GPT 5.2, classify all validation queries correctly, including the intentionally inconsistent ones. The other models commit some minor mistakes in erroneously detecting inconsistencies in the durations of the actions. Interestingly, all of Mixtral’s errors are false negatives on intentionally incorrect Blocks World instances, accepting 2.b, 3.b, and 5.b as valid. This pattern suggests a permissive validation behavior, where the model accepts the user’s specification as coherent despite explicit contradictions or abstraction-level mismatches. This behavior is related to \emph{sycophancy}, i.e., the tendency of language models to align with a user’s point of view even when it is objectively incorrect~\cite{sharma2024towards,malmqvist2025sycophancy,wang2026truth,10825538}. For more details, refer to~\ref{ssec:app_domainValidationInsights}.

\subsubsection{\kb Generation} 
\label{sssec:expKBGeneration}

The \kb generation represents a more complex step for a generative model. In order to correctly evaluate this step, we first describe the process of checking the output of the LLMs, which consisted in the following steps:
\begin{enumerate}
    \item The framework extracts the correct parts from the output and writes them into a Prolog file containing the \kb;
    \item We include the file in the planner and test whether it finds a plan for the HL \kb;
    \item If the planner cannot find a plan, then we enable the explainability functions that help the user understand why a problem is not solvable and fix the \kb. 
    \item The same process as step 3. is repeated also for the LL \kb, after fixing possible mistakes made by the model during the generation of the HL \kb.
\end{enumerate}

As mentioned in Section~\ref{ssec:KmsConsistencyChecks}, the framework can iterate multiple times on the generation of the \kb, each time applying the consistency checks in Table~\ref{tab:kms-consistency-checks} and, if any fails, providing a feedback to the LLM in order to correct errors. For this experiment, we set the iteration value to 3, hence \frameworkname could ask to correct the failed consistency checks to the LLM up to 3 times. 

When \frameworkname fails to find a plan from a \kb, this must be manually corrected. To evaluate how bad the \kb is, we count the number of logical changes that had to be make for each component of the \kb. For instance, if the LLM did not model correctly a precondition for the actions, then we consider that as one error even though it may be necessary to change multiple actions by adding said precondition.

Notice that, while we ran experiments also using Mixtral, we do not report them here as the model could not produce any feasible \kbase. The generated \kbs are still available online. 

\begin{table*}[t]
    \centering
    
    \caption{Results for the generation of the \HL \kbase using the models listed on the left. A \cm~indicates that the model's output was completely correct, while a tuple of numbers indicates how many logical changes had to be made for each component of the \kb ($K$, $I$, $G$, $DA_H$).}
    \label{tab:hlRes}
    
    \setlength{\tabcolsep}{4pt}
    \renewcommand{\arraystretch}{1.05}
    \footnotesize
    
    \begin{tabular}{cccccccc}
        \toprule
        & 
        & Opus 4.6
        & \rotatebox[origin=c]{0}{Sonnet 4.6}
        & \rotatebox[origin=c]{0}{GPT 5.2}
        & \rotatebox[origin=c]{0}{GPT 5.4 Mini} 
        & Qwen 3.6 
        & Llama 3.3
        \\
        
        \midrule
    
        \multirow{6}{*}{\rotatebox[origin=c]{90}{Blocks World}}
        & 1 & \cm       & \cm       & \cm       & \cm       & \cm       & \cm       \\
        & 2 & \cm       & \cm       & \cm       & \cm       & \cm       & (0,0,1,2) \\
        & 3 & (0,0,0,1) & (0,0,0,1) & (0,0,0,1) & (3,0,0,1) & (0,0,0,1) & (0,1,0,1) \\
        & 4 & \cm       & \cm       & \cm       & \cm       & \cm       & \cm       \\
        & 5 & \cm       & (1,0,1,0) & \cm       & \cm       & (1,0,0,0) & (1,1,1,0) \\
        & 6 & \cm       & \cm       & \cm       & \cm       & (0,0,1,0) & \cm       \\
        
        \midrule
    
        \multirow{6}{*}{\rotatebox[origin=c]{90}{Grippers}}
        & 1 & \cm       & \cm       & \cm       & (0,0,0,1) & \cm       & \cm       \\
        & 2 & \cm       & (0,0,0,2) & \cm       & (0,2,2,2) & \cm       & \cm       \\
        & 3 & \cm       & \cm       & \cm       & (0,0,0,3) & (0,0,0,1) & (0,0,1,2) \\
        & 4 & \cm       & \cm       & (0,0,0,1) & (0,0,0,5) & (0,0,0,1) & (0,0,0,2) \\
        & 5 & \cm       & (0,0,0,1) & (1,0,0,1) & (2,0,0,2) & (1,0,0,4) & (1,0,0,4) \\
        & 6 & \cm       & (0,0,0,4) & \cm       & (0,1,1,8) & (0,0,0,2) & \cm       \\
    
        \bottomrule
    \end{tabular}
\end{table*}

\paragraph{High-Level \kb Generation} Overall, the results show that most models are able to correctly generate the static part of the \HL \kbase, while the action model remains the main source of errors. Across the reported runs, 54 out of 78 required logical changes concerning the \HL action set $DA_H$, compared to 11 for $K$, 5 for $I$, and 8 for $G$. This suggests that LLMs usually identify the relevant objects, locations, and goal facts, but may struggle to encode the transition semantics needed by the planner.

Opus 4.6 is the most reliable model in this setting, requiring only one action-level correction over all 12 instances. GPT 5.2 is the second strongest model, with 9 completely correct \HL \kbases. 

GPT 5.4 Mini was the weakest model among the reported ones for \HL \kb generation. It produced a completely correct \kb in only 5 out of 12 instances, and all of these successful cases were in Blocks World. The gap is more evident in Grippers, where this LLM requires corrections in every instance, mostly in $DA_H$. This is expected because the Grippers variants introduce door constraints, special balls, and multi-robot cooperation, which must be represented through precise action preconditions and effects. This result is particularly informative because it cannot be attributed to in-domain few-shot examples: the examples provided to the LLMs were based on Blocks World and AGV navigation, not on Grippers. Moreover, they not only did not include any Grippers-domain instance, but also did not contain examples involving mutual-exclusion constraints such as narrow doors or objects requiring specialized forms of cooperation. This is also reflected in the repair loop, where GPT 5.4 Mini frequently reached the maximum number of consistency-repair iterations, especially in the Grippers domain. Refer to~\ref{ssec:app_hlKBGenerationInsights} for additional insights on the results.

\paragraph{Low-Level \kb Generation}
Once we obtained and fixed the problems with the \HL \kbases, we feed them back to the LLMs to generate the \LL \kbs. As before, also in this case we allowed the models up to 3 iterations to fix possible wrong consistency checks.

\begin{table*}[t]
    \centering
    
    \caption{Results for the generation of the LL \kbase (\kb) using the models listed on the left. A \checkmark~indicates that the model's output was completely correct, while a tuple denotes incorrect output. Each number within the tuple indicates how many changes to that part of the \kb were necessary: $(K, I, G, DA_H, DA_L, M)$.}
    \label{tab:llRes}
    
    \setlength{\tabcolsep}{3pt}
    \renewcommand{\arraystretch}{1.05}
    \footnotesize
    
    \begin{tabular}{cccccccc}
        \toprule
        &   & Opus 4.6      & Sonnet 4.6    & GPT 5.2       & GPT 5.4       & Qwen 3.6      & Llama 3.3 \\
        &   &               &               &               & Mini          &               &           \\
    
        \midrule
    
        \multirow{6}{*}{\rotatebox[origin=c]{90}{Blocks World}}
        & 1 & \cm           & \cm           & \cm           & (0,0,0,0,0,1) & (1,0,0,0,0,0) & (1,0,0,0,0,0) \\
        & 2 & \cm           & \cm           & \cm           & (0,0,0,0,0,1) & (1,0,0,0,0,0) & (1,0,0,0,0,2) \\
        & 3 & (0,0,0,0,1,0) & (0,0,0,0,1,0) & \cm           & \cm           & (1,0,0,0,1,0) & (1,0,0,0,1,0) \\
        & 4 & \cm           & \cm           & (1,0,0,0,0,0) & (0,0,0,0,0,1) & (1,0,0,0,0,0) & (1,0,0,0,0,0) \\
        & 5 & \cm           & (1,1,0,0,0,0) & \cm           & \cm           & (1,0,0,0,0,0) & (1,0,0,0,0,0) \\
        & 6 & (0,0,0,0,1,0) & (0,0,0,0,1,0) & (0,0,0,0,1,0) & \cm           & (1,0,0,0,1,0) & (1,0,0,0,1,0) \\
        
        \midrule
    
        \multirow{6}{*}{\rotatebox[origin=c]{90}{Grippers}}
        & 1 & (0,0,0,0,0,2) & (0,0,1,0,0,2) & \cm           & (0,0,0,0,3,0) & (0,1,1,0,1,0) & (0,1,1,0,3,1) \\
        & 2 & \cm           & \cm           & \cm           & (0,1,1,0,4,0) & (0,1,0,0,3,1) & (0,0,0,0,4,1) \\
        & 3 & (0,0,0,0,0,1) & \cm           & \cm           & (0,1,1,0,5,8) & (0,1,1,0,4,2) & (0,2,2,0,3,5) \\
        & 4 & \cm           & \cm           & \cm           & (0,1,1,0,2,2) & (0,1,0,0,6,1) & (0,0,0,0,8,1) \\
        & 5 & \cm           & (0,0,0,0,2,2) & (1,0,0,0,1,2) & (0,0,0,0,2,1) & (0,1,1,0,6,0) & (0,1,1,0,1,3) \\
        & 6 & \cm           & \cm           & \cm           & (0,0,0,0,0,2) & (0,0,0,0,3,0) & (0,0,0,0,0,7) \\
    
        \bottomrule
    \end{tabular}
\end{table*}

The results in Table~\ref{tab:llRes} show that the \LL generation is substantially more difficult than the \HL one, but also that the difficulty is not uniform across domains. Across all reported experiments, 27 out of 72 \LL \kbases were generated without any manual correction. Out of 158 logical errors, 70 concern the low-level action set $DA_L$ and 49 concern the mapping $M$ between \HL and \LL actions.

In the Blocks World domain, the number of required changes is low. Most of these edits are not conceptual changes to the manipulation model, but missing static position facts in $K$. 

In particular, Qwen 3.6 and Llama 3.3 repeatedly omit low-level arm home positions, such as \verb|pos(4,4)| and \verb|pos(5,5)|, or task-specific positions such as \verb|pos(7.5,6)| for the architrave placement. While this produces an invalid \LL \kb, the correction is trivial: adding the missing \verb|pos/2| facts is usually sufficient. This explains why, in Blocks World, 14 out of 29 corrections concern $K$, while only 9 concern $DA_L$ and 5 concern $M$.

The Grippers domain is considerably harder. In this domain, 13 out of 36 \LL \kbases are completely correct, and 105 out of 129 corrections concern either $DA_L$ or $M$. This is expected because the \LL model is not just a short sequence of arm motions, but it must maintain the state of agents and objects, while also preserving the intended abstraction relation with the \HL actions. The most striking error consists in the models using \HL state predicates inside \LL actions, even though they have been instructed against in the examples. Without this separation, the planner can derive a high-level plan but the low-level state is not independently updated by the refinement actions. In particular, Qwen 3.6 comes out as the model that is most susceptible to this kind of mistake. Another frequent Grippers error concerns the representation of possession: in several instances, the \kbases represented gripping actions without deleting the empty state of the gripper, or vice versa. Others move a robot while failing to carry the low-level ball state with it. The mapping component $M$ is the second main source of errors. Refer to~\ref{ssec:app_llKBGenerationInsights} for more details on the experiment results.

GPT 5.2 is the strongest model in the low-level experiments, producing 9 completely correct knowledge-bases out of 12. Opus 4.6 and Sonnet 4.6 follow with 8 and 7 completely correct knowledge-bases, respectively. The remaining errors are mostly small mapping or naming issues rather than failures to generate the correct low-level transition model. GPT 5.4 Mini performs well on the simpler Blocks World cases but fails considerably in Grippers, where it requires 35 logical changes. Qwen 3.6 and Llama 3.3 never produce a completely correct \LL \kb in the reported runs. Their files are often close enough to be repaired, but they repeatedly miss static \LL facts in Blocks World and confuse high-level and low-level state predicates in Grippers. Llama 3.3 also shows more mapping-sequence errors, especially in cases requiring a final home pose or coordinated manipulation.

Overall, the \LL generation results suggest that LLMs are reasonably good at identifying the primitive actions required and the mapping between the \HL and the \LL. The main reliability issue is maintaining the abstraction boundary between the \HL and \LL state spaces, using the wrong predicates in the \LL actions. The main conclusion is therefore that \LL \kb generation benefits from the same LLM-based synthesis used for \HL \kb generation, but it requires a stronger guidance. This is an expected outcome given the much more complex nature of threading the needle between the two levels of abstraction. A more specific set of examples on these features or a fine-tuning of the models should allow for a reduction of the mistakes made by the models. 

All in all, the scale of the generated \kbases (Table~\ref{tab:predNumber},~\ref{ssec:app_tabPredicatesKB}) makes the benefit of LLM-based generation apparent. A corrected \LL \kb contains, on average, 207.3 non-comment Prolog lines, 54.0 state facts, and 14.4 action or mapping schemas; in the Grippers domain, this increases to 257.1 lines and 16.9 schemas on average, with the largest generated \kb reaching 561 non-comment Prolog lines. Writing and debugging these \kbs manually would require encoding hundreds of facts, action conditions, effects, and refinement steps. Even when manual corrections are necessary, the LLM usually produces most of this structure correctly, leaving the user to repair a comparatively small number of logical modeling errors rather than author the full \kb from scratch.

\paragraph{Consistency checks}

\begin{table*}[t]
    \centering
    \caption{Most frequent failed \HL and \LL consistency checks during the repair loop. Appearances counts how many failed repair iterations reported each problem.}
    \label{tab:consistencyFailures}
    \footnotesize
        \begin{tabular}{clc}
        \toprule
        Layer & Failed check & Appearances \\
        \midrule
        \multirow{6}{*}{\rotatebox[origin=c]{90}{High-Level}} 
        & Declared facts never used & 54 \\
        & Effects break action invariants & 5 \\
        & Undefined predicates in actions & 3 \\
        & Contradictory add/delete effects & 2 \\
        & Deleted facts never restored & 10 \\
        & Unreachable action preconditions & 3 \\
        \midrule
        \multirow{9}{*}{\rotatebox[origin=c]{90}{Low-level}} 
        & Duration bounds do not cover mapped \LL actions & 107 \\
        & Unreachable action preconditions & 77 \\
        & Added facts never required & 2 \\
        & Declared facts never used & 50 \\
        & Undefined predicates in actions & 15 \\
        & Effects break action invariants & 3 \\
        & \LL actions modify \HL predicates & 28 \\
        & Static predicates changed by effects & 9 \\
        & Predicates shared by static \kb and states & 7 \\
        \bottomrule
    \end{tabular}
\end{table*}

In Table~\ref{tab:consistencyFailures}, we can see that the most frequent automatic \HL consistency failures concern inconsistent action effects and facts that are declared in the static part of the domain but never used by any state, precondition, or effect. This typically happens when the model introduces plausible domain facts such as rooms, doors, or robot types, but does not integrate them into the transition model. At the \LL layer, the most frequent failures are different: most repair iterations report mappings whose \HL duration bounds do not cover the sum of the mapped \LL action durations, followed by unreachable \LL action preconditions. The \LL checks also expose abstraction violations, where \LL actions modify \HL predicates, and static predicates that are changed by effects or reused as state predicates. These checks are useful for catching many malformed transition systems and invalid refinements, but they do not capture every semantic error. Some \kbases pass the structural checks while still requiring manual correction because they permit the wrong plan or omit a domain-specific action as shown in Table~\ref{tab:repairIterations},~\ref{ssec:app_repairLoopResults}. This table shows that the repair loop is more frequently activated for \LL generation than for \HL generation, especially in the Grippers domain. This is consistent with the failure categories in Table~\ref{tab:consistencyFailures}: \LL generation introduces additional constraints from action durations, mappings, and abstraction separation, which are not present in the \HL case. The repeated occurrence of entries equal to 3 indicates that several \LL \kbases reached the maximum number of allowed repair iterations, particularly for GPT 5.4 Mini, Qwen 3.6, and Llama 3.3 on the Grippers instances.

At the same time, the underlined entries show that the repair loop should not be interpreted as a complete correctness guarantee. Some \kbases still required manual correction even after no failed consistency check was reported, while others reached the maximum number of repair iterations and remained incorrect. This confirms that the automatic checks are useful for removing many structural errors, but semantic errors in action definitions or refinements can remain undetected, albeit more consistency checks depending on the considered scenarios can be added.

%% file: sections/5-experiments/d-planner.tex
In this section, we evaluate the performance of the planner using the low-level KBs generated and manually corrected in Section~\ref{sssec:expKBGeneration}. We emphasize that our objective is not to claim that the proposed planner matches state-of-the-art task planners, but rather to assess whether it provides a feasible planning component within the integrated framework. We divide this section into the three main steps for the planner: 
\begin{enumerate*}[label=(\roman*)]
    \item high-level and low-level total-order plan generation (\textbf{R1}),
    \item enablers extraction, and 
    \item STN generation and optimization (\textbf{R4}).
\end{enumerate*}

\begin{figure*}[htp]
    \centering
    \begin{subfigure}{\textwidth}
        \centering
        \caption{Total-order and low-level-order timings -- Blocks World}
        \label{fig:topobw}
        \input{figures/experiments/timings_to_po_bw}
    \end{subfigure}
    \vspace{0.1cm}\\
    \begin{subfigure}{\textwidth}
        \centering
        \caption{Total-order and low-level-order timings -- Grippers}
        \label{fig:topog}
        \input{figures/experiments/timings_to_po_g}
    \end{subfigure}
    \caption{Average time required to generate the high-level total-order plan and to expand it into the corresponding low-level total-order plan. Instances in which the planner could not find a solution within the timeout are not reported.}
    \label{fig:toPoTimesFig}
\end{figure*}

Figure~\ref{fig:toPoTimesFig} shows the average time over 10 runs required to compute the high-level total-order plan and to expand it into the corresponding low-level total-order plan. Table~\ref{tab:toPoTimesTable} (\ref{ssec:app_toTimes}) reports the same results together with the standard deviation. The planner was given a timeout of 660 seconds for each run. We also used a patience value of 2: if the planner timed out twice consecutively for the same knowledge-base, we considered the corresponding search space too large and stopped evaluating that case. For readability, Figure~\ref{fig:toPoTimesFig} reports only successful runs, while Table~\ref{tab:toPoTimesTable} marks unsolved cases with \texttt{X}.

\begin{figure*}
    \centering
    \resizebox{\textwidth}{!}{\input{figures/experiments/timings_normalized_to.tex}}
    \caption{Normalized total-order planning times for the successful runs. Times are normalized within each benchmark instance (w.r.t. the instance average) to compare the relative behavior of the KBs generated by different models. Timed-out cases are omitted from the computations.}
    \label{fig:toNormalizedTimesFig}
\end{figure*}

The results show that the planning pipeline times are dominated by the total-order high-level plan search. Once a high-level total-order plan is found, the low-level expansion step is comparatively inexpensive, since it only applies the mappings encoded in the KB and checks the consistency of the resulting low-level sequence. This behavior is expected: the total-order planner explores the symbolic state space, while the low-level expansion operates on an already selected sequence of high-level actions.

The hardest cases are those in which the generated \kb induces a large number of possible groundings or the ones in which it weakly constrains the applicable actions. In these cases, the breadth-first search may explore a large portion of the state space before reaching a solution and may therefore exceed the timeout requirement. These failures should not be interpreted as failures of the LLM: different generated \kbs, even after manual correction, can still encode the same task with slightly different action preconditions, effects, or static predicates, and these modelling choices can substantially affect the branching factor of the planner.

Planning times vary substantially across model-generated \kbs for some instances, although no model is consistently fastest across all scenarios. This indicates that relatively small differences in predicate ordering, action preconditions, and transition models can strongly affect the branching factor of the breadth-first search. This is expected, since the prompting procedure asks the models to produce correct and inspectable KBs, not planner-optimized ones. Figure~\ref{fig:toNormalizedTimesFig} supports this observation by comparing the normalized planning times across instances. Apart from a few outliers, the models exhibit similar behavior, indicating that the main driver of planning time is the structure of the domain instance rather than the specific LLM used to generate the KB. The normalized values were obtained by dividing each result by the average of that instance.

Figure~\ref{fig:timesEnablers} and Figure~\ref{fig:timesSTNGenOpt} report the time required to extract the enablers from the low-level total-order plan and the time required to construct and optimize the STN, respectively. They show that these steps are negligible when compared with total-order high-level plan search. The cost of both steps depends mainly on the length of the generated plan, rather than on the full symbolic search space.

Overall, these results confirm that the proposed planner is sufficient as a feasible planning component inside \frameworkname. The system can transform corrected LLM-generated \kbs into low-level temporal plans and expose the intermediate artifacts needed for inspection. At the same time, the experiments also highlight the main scalability limitation of the current implementation: the total-order symbolic search remains the bottleneck. This is consistent with the deliberately simple planner used in this work, whose goal is integration and inspectability rather than competitive performance with state-of-the-art planners.

%% file: figures/experiments/timings_to_po_bw.tex
  % legend style={at={(0.98,0.98)}, anchor=north east, draw=black!25, fill=white, fill opacity=0.88, text opacity=1, font=\scriptsize},
  % legend cell align={left},

% Requires: \usepackage{pgfplots}
% Recommended: \pgfplotsset{compat=1.18}
\definecolor{mplBlue}{HTML}{1F77B4}
\definecolor{mplOrange}{HTML}{FF7F0E}
\definecolor{mplGreen}{HTML}{2CA02C}
\definecolor{mplRed}{HTML}{D62728}
\definecolor{mplPurple}{HTML}{9467BD}
\begin{tikzpicture}
\begin{axis}[
  width=0.96\textwidth,
  height=0.42\textwidth,
  xlabel={\scriptsize KB},
  ylabel={\scriptsize Mean time (s)},
  xmin=-0.5, xmax=28.5,
  ymode=log,
  log origin=infty,
  ymin=1e-05, ymax=1000,
  ytickten={-5,-4,-3,-2,-1,0,1,2,3},
  minor y tick num=8,
  major tick length=2.6pt,
  minor tick length=1.8pt,
  minor tick style={black!70, line width=0.35pt},
  xtick={0,1,2,3,4,5,6,7,8,9,10,11,12,13,14,15,16,17,18,19,20,21,22,23,24,25,26,27,28},
  xticklabels={{1 -- Opus 4.6},{1 -- Sonnet 4.6},{1 -- GPT5.2},{1 -- GPT5.4 Mini},{1 -- Qwen},{1 -- Llama},{2 -- Opus 4.6},{2 -- Sonnet 4.6},{2 -- GPT5.2},{2 -- Qwen},{2 -- Llama},{3 -- Opus 4.6},{3 -- Sonnet 4.6},{3 -- GPT5.2},{3 -- GPT5.4 Mini},{3 -- Qwen},{3 -- Llama},{4 -- Opus 4.6},{4 -- Sonnet 4.6},{4 -- GPT5.2},{4 -- GPT5.4 Mini},{4 -- Qwen},{4 -- Llama},{6 -- Opus 4.6},{6 -- Sonnet 4.6},{6 -- GPT5.2},{6 -- GPT5.4 Mini},{6 -- Qwen},{6 -- Llama}},
  tick align=outside,
  tick pos=left,
  x tick label style={rotate=90, anchor=east, font=\scriptsize},
  y tick label style={font=\scriptsize},
  label style={font=\small},
  ymajorgrids=true,
  grid style={dotted, gray!45},
  minor grid style={dotted, gray!25},
  legend style={at={(0.98,0.98)}, anchor=north east, draw=black!25, fill=white, fill opacity=0.88, text opacity=1, font=\scriptsize},
  legend cell align={left},
  axis line style={black!70},
  clip=false
]
\addplot[ybar, area legend, bar width=0.3256, fill=mplBlue, draw=black, line width=0.4pt] coordinates {(-0.185,0.299616) (0.815,0.195559) (1.815,0.0503652) (2.815,0.198308) (3.815,0.281305) (4.815,0.197928) (5.815,533.52) (6.815,342.062) (7.815,545.556) (8.815,25.2076) (9.815,50.5961) (10.815,11.4265) (11.815,11.4505) (12.815,1.87071) (13.815,306.617) (14.815,404.339) (15.815,192.856) (16.815,1.0553) (17.815,1.04399) (18.815,1.04383) (19.815,0.891305) (20.815,1.04594) (21.815,0.867242) (22.815,0.494357) (23.815,0.220944) (24.815,0.398089) (25.815,0.0428484) (26.815,0.416475) (27.815,0.312912)};
\addlegendentry{Total order}
\addplot[ybar, area legend, bar width=0.3256, fill=mplOrange, draw=black, line width=0.4pt] coordinates {(0.185,8.38041e-05) (1.185,7.78437e-05) (2.185,5.91755e-05) (3.185,7.56741e-05) (4.185,7.82251e-05) (5.185,8.40425e-05) (6.185,0.000162339) (7.185,0.000142288) (8.185,0.000156927) (9.185,0.00016346) (10.185,0.000147533) (11.185,0.000163651) (12.185,0.000162959) (13.185,0.000128889) (14.185,0.000138927) (15.185,0.000164986) (16.185,0.000168371) (17.185,0.00016551) (18.185,0.000136256) (19.185,0.000140953) (20.185,0.000149155) (21.185,0.000162148) (22.185,0.000150299) (23.185,8.88348e-05) (24.185,0.000106263) (25.185,8.86917e-05) (26.185,9.18627e-05) (27.185,9.61542e-05) (28.185,0.000116205)};
\addlegendentry{Low-level order}
\draw[black, line width=0.8pt] (axis cs:-0.185,0.246828) -- (axis cs:-0.185,0.352404);
\draw[black, line width=0.8pt] (axis cs:-0.23,0.246828) -- (axis cs:-0.14,0.246828);
\draw[black, line width=0.8pt] (axis cs:-0.23,0.352404) -- (axis cs:-0.14,0.352404);
\draw[black, line width=0.8pt] (axis cs:0.815,0.193452) -- (axis cs:0.815,0.197667);
\draw[black, line width=0.8pt] (axis cs:0.77,0.193452) -- (axis cs:0.86,0.193452);
\draw[black, line width=0.8pt] (axis cs:0.77,0.197667) -- (axis cs:0.86,0.197667);
\draw[black, line width=0.8pt] (axis cs:1.815,0.047303) -- (axis cs:1.815,0.0534273);
\draw[black, line width=0.8pt] (axis cs:1.77,0.047303) -- (axis cs:1.86,0.047303);
\draw[black, line width=0.8pt] (axis cs:1.77,0.0534273) -- (axis cs:1.86,0.0534273);
\draw[black, line width=0.8pt] (axis cs:2.815,0.193713) -- (axis cs:2.815,0.202902);
\draw[black, line width=0.8pt] (axis cs:2.77,0.193713) -- (axis cs:2.86,0.193713);
\draw[black, line width=0.8pt] (axis cs:2.77,0.202902) -- (axis cs:2.86,0.202902);
\draw[black, line width=0.8pt] (axis cs:3.815,0.276404) -- (axis cs:3.815,0.286207);
\draw[black, line width=0.8pt] (axis cs:3.77,0.276404) -- (axis cs:3.86,0.276404);
\draw[black, line width=0.8pt] (axis cs:3.77,0.286207) -- (axis cs:3.86,0.286207);
\draw[black, line width=0.8pt] (axis cs:4.815,0.191024) -- (axis cs:4.815,0.204831);
\draw[black, line width=0.8pt] (axis cs:4.77,0.191024) -- (axis cs:4.86,0.191024);
\draw[black, line width=0.8pt] (axis cs:4.77,0.204831) -- (axis cs:4.86,0.204831);
\draw[black, line width=0.8pt] (axis cs:5.815,518.453) -- (axis cs:5.815,548.586);
\draw[black, line width=0.8pt] (axis cs:5.77,518.453) -- (axis cs:5.86,518.453);
\draw[black, line width=0.8pt] (axis cs:5.77,548.586) -- (axis cs:5.86,548.586);
\draw[black, line width=0.8pt] (axis cs:6.815,334.62) -- (axis cs:6.815,349.504);
\draw[black, line width=0.8pt] (axis cs:6.77,334.62) -- (axis cs:6.86,334.62);
\draw[black, line width=0.8pt] (axis cs:6.77,349.504) -- (axis cs:6.86,349.504);
\draw[black, line width=0.8pt] (axis cs:7.815,530.717) -- (axis cs:7.815,560.395);
\draw[black, line width=0.8pt] (axis cs:7.77,530.717) -- (axis cs:7.86,530.717);
\draw[black, line width=0.8pt] (axis cs:7.77,560.395) -- (axis cs:7.86,560.395);
\draw[black, line width=0.8pt] (axis cs:8.815,24.4168) -- (axis cs:8.815,25.9983);
\draw[black, line width=0.8pt] (axis cs:8.77,24.4168) -- (axis cs:8.86,24.4168);
\draw[black, line width=0.8pt] (axis cs:8.77,25.9983) -- (axis cs:8.86,25.9983);
\draw[black, line width=0.8pt] (axis cs:9.815,49.4353) -- (axis cs:9.815,51.7569);
\draw[black, line width=0.8pt] (axis cs:9.77,49.4353) -- (axis cs:9.86,49.4353);
\draw[black, line width=0.8pt] (axis cs:9.77,51.7569) -- (axis cs:9.86,51.7569);
\draw[black, line width=0.8pt] (axis cs:10.815,11.1378) -- (axis cs:10.815,11.7152);
\draw[black, line width=0.8pt] (axis cs:10.77,11.1378) -- (axis cs:10.86,11.1378);
\draw[black, line width=0.8pt] (axis cs:10.77,11.7152) -- (axis cs:10.86,11.7152);
\draw[black, line width=0.8pt] (axis cs:11.815,11.152) -- (axis cs:11.815,11.749);
\draw[black, line width=0.8pt] (axis cs:11.77,11.152) -- (axis cs:11.86,11.152);
\draw[black, line width=0.8pt] (axis cs:11.77,11.749) -- (axis cs:11.86,11.749);
\draw[black, line width=0.8pt] (axis cs:12.815,1.86046) -- (axis cs:12.815,1.88095);
\draw[black, line width=0.8pt] (axis cs:12.77,1.86046) -- (axis cs:12.86,1.86046);
\draw[black, line width=0.8pt] (axis cs:12.77,1.88095) -- (axis cs:12.86,1.88095);
\draw[black, line width=0.8pt] (axis cs:13.815,301.524) -- (axis cs:13.815,311.709);
\draw[black, line width=0.8pt] (axis cs:13.77,301.524) -- (axis cs:13.86,301.524);
\draw[black, line width=0.8pt] (axis cs:13.77,311.709) -- (axis cs:13.86,311.709);
\draw[black, line width=0.8pt] (axis cs:14.815,398.484) -- (axis cs:14.815,410.195);
\draw[black, line width=0.8pt] (axis cs:14.77,398.484) -- (axis cs:14.86,398.484);
\draw[black, line width=0.8pt] (axis cs:14.77,410.195) -- (axis cs:14.86,410.195);
\draw[black, line width=0.8pt] (axis cs:15.815,189.37) -- (axis cs:15.815,196.341);
\draw[black, line width=0.8pt] (axis cs:15.77,189.37) -- (axis cs:15.86,189.37);
\draw[black, line width=0.8pt] (axis cs:15.77,196.341) -- (axis cs:15.86,196.341);
\draw[black, line width=0.8pt] (axis cs:16.815,1.01931) -- (axis cs:16.815,1.09129);
\draw[black, line width=0.8pt] (axis cs:16.77,1.01931) -- (axis cs:16.86,1.01931);
\draw[black, line width=0.8pt] (axis cs:16.77,1.09129) -- (axis cs:16.86,1.09129);
\draw[black, line width=0.8pt] (axis cs:17.815,1.01937) -- (axis cs:17.815,1.0686);
\draw[black, line width=0.8pt] (axis cs:17.77,1.01937) -- (axis cs:17.86,1.01937);
\draw[black, line width=0.8pt] (axis cs:17.77,1.0686) -- (axis cs:17.86,1.0686);
\draw[black, line width=0.8pt] (axis cs:18.815,1.01875) -- (axis cs:18.815,1.06891);
\draw[black, line width=0.8pt] (axis cs:18.77,1.01875) -- (axis cs:18.86,1.01875);
\draw[black, line width=0.8pt] (axis cs:18.77,1.06891) -- (axis cs:18.86,1.06891);
\draw[black, line width=0.8pt] (axis cs:19.815,0.802762) -- (axis cs:19.815,0.979848);
\draw[black, line width=0.8pt] (axis cs:19.77,0.802762) -- (axis cs:19.86,0.802762);
\draw[black, line width=0.8pt] (axis cs:19.77,0.979848) -- (axis cs:19.86,0.979848);
\draw[black, line width=0.8pt] (axis cs:20.815,1.01901) -- (axis cs:20.815,1.07288);
\draw[black, line width=0.8pt] (axis cs:20.77,1.01901) -- (axis cs:20.86,1.01901);
\draw[black, line width=0.8pt] (axis cs:20.77,1.07288) -- (axis cs:20.86,1.07288);
\draw[black, line width=0.8pt] (axis cs:21.815,0.849154) -- (axis cs:21.815,0.88533);
\draw[black, line width=0.8pt] (axis cs:21.77,0.849154) -- (axis cs:21.86,0.849154);
\draw[black, line width=0.8pt] (axis cs:21.77,0.88533) -- (axis cs:21.86,0.88533);
\draw[black, line width=0.8pt] (axis cs:22.815,0.487421) -- (axis cs:22.815,0.501294);
\draw[black, line width=0.8pt] (axis cs:22.77,0.487421) -- (axis cs:22.86,0.487421);
\draw[black, line width=0.8pt] (axis cs:22.77,0.501294) -- (axis cs:22.86,0.501294);
\draw[black, line width=0.8pt] (axis cs:23.815,0.195751) -- (axis cs:23.815,0.246138);
\draw[black, line width=0.8pt] (axis cs:23.77,0.195751) -- (axis cs:23.86,0.195751);
\draw[black, line width=0.8pt] (axis cs:23.77,0.246138) -- (axis cs:23.86,0.246138);
\draw[black, line width=0.8pt] (axis cs:24.815,0.380582) -- (axis cs:24.815,0.415595);
\draw[black, line width=0.8pt] (axis cs:24.77,0.380582) -- (axis cs:24.86,0.380582);
\draw[black, line width=0.8pt] (axis cs:24.77,0.415595) -- (axis cs:24.86,0.415595);
\draw[black, line width=0.8pt] (axis cs:25.815,0.0424763) -- (axis cs:25.815,0.0432204);
\draw[black, line width=0.8pt] (axis cs:25.77,0.0424763) -- (axis cs:25.86,0.0424763);
\draw[black, line width=0.8pt] (axis cs:25.77,0.0432204) -- (axis cs:25.86,0.0432204);
\draw[black, line width=0.8pt] (axis cs:26.815,0.403136) -- (axis cs:26.815,0.429814);
\draw[black, line width=0.8pt] (axis cs:26.77,0.403136) -- (axis cs:26.86,0.403136);
\draw[black, line width=0.8pt] (axis cs:26.77,0.429814) -- (axis cs:26.86,0.429814);
\draw[black, line width=0.8pt] (axis cs:27.815,0.303095) -- (axis cs:27.815,0.32273);
\draw[black, line width=0.8pt] (axis cs:27.77,0.303095) -- (axis cs:27.86,0.303095);
\draw[black, line width=0.8pt] (axis cs:27.77,0.32273) -- (axis cs:27.86,0.32273);
\draw[black, line width=0.8pt] (axis cs:0.185,7.74104e-05) -- (axis cs:0.185,9.01979e-05);
\draw[black, line width=0.8pt] (axis cs:0.14,7.74104e-05) -- (axis cs:0.23,7.74104e-05);
\draw[black, line width=0.8pt] (axis cs:0.14,9.01979e-05) -- (axis cs:0.23,9.01979e-05);
\draw[black, line width=0.8pt] (axis cs:1.185,7.4341e-05) -- (axis cs:1.185,8.13463e-05);
\draw[black, line width=0.8pt] (axis cs:1.14,7.4341e-05) -- (axis cs:1.23,7.4341e-05);
\draw[black, line width=0.8pt] (axis cs:1.14,8.13463e-05) -- (axis cs:1.23,8.13463e-05);
\draw[black, line width=0.8pt] (axis cs:2.185,4.8725e-05) -- (axis cs:2.185,6.9626e-05);
\draw[black, line width=0.8pt] (axis cs:2.14,4.8725e-05) -- (axis cs:2.23,4.8725e-05);
\draw[black, line width=0.8pt] (axis cs:2.14,6.9626e-05) -- (axis cs:2.23,6.9626e-05);
\draw[black, line width=0.8pt] (axis cs:3.185,7.00128e-05) -- (axis cs:3.185,8.13353e-05);
\draw[black, line width=0.8pt] (axis cs:3.14,7.00128e-05) -- (axis cs:3.23,7.00128e-05);
\draw[black, line width=0.8pt] (axis cs:3.14,8.13353e-05) -- (axis cs:3.23,8.13353e-05);
\draw[black, line width=0.8pt] (axis cs:4.185,6.09968e-05) -- (axis cs:4.185,9.54535e-05);
\draw[black, line width=0.8pt] (axis cs:4.14,6.09968e-05) -- (axis cs:4.23,6.09968e-05);
\draw[black, line width=0.8pt] (axis cs:4.14,9.54535e-05) -- (axis cs:4.23,9.54535e-05);
\draw[black, line width=0.8pt] (axis cs:5.185,7.99758e-05) -- (axis cs:5.185,8.81093e-05);
\draw[black, line width=0.8pt] (axis cs:5.14,7.99758e-05) -- (axis cs:5.23,7.99758e-05);
\draw[black, line width=0.8pt] (axis cs:5.14,8.81093e-05) -- (axis cs:5.23,8.81093e-05);
\draw[black, line width=0.8pt] (axis cs:6.185,0.000155187) -- (axis cs:6.185,0.000169491);
\draw[black, line width=0.8pt] (axis cs:6.14,0.000155187) -- (axis cs:6.23,0.000155187);
\draw[black, line width=0.8pt] (axis cs:6.14,0.000169491) -- (axis cs:6.23,0.000169491);
\draw[black, line width=0.8pt] (axis cs:7.185,0.00013472) -- (axis cs:7.185,0.000149857);
\draw[black, line width=0.8pt] (axis cs:7.14,0.00013472) -- (axis cs:7.23,0.00013472);
\draw[black, line width=0.8pt] (axis cs:7.14,0.000149857) -- (axis cs:7.23,0.000149857);
\draw[black, line width=0.8pt] (axis cs:8.185,0.000148222) -- (axis cs:8.185,0.000165632);
\draw[black, line width=0.8pt] (axis cs:8.14,0.000148222) -- (axis cs:8.23,0.000148222);
\draw[black, line width=0.8pt] (axis cs:8.14,0.000165632) -- (axis cs:8.23,0.000165632);
\draw[black, line width=0.8pt] (axis cs:9.185,0.000153354) -- (axis cs:9.185,0.000173565);
\draw[black, line width=0.8pt] (axis cs:9.14,0.000153354) -- (axis cs:9.23,0.000153354);
\draw[black, line width=0.8pt] (axis cs:9.14,0.000173565) -- (axis cs:9.23,0.000173565);
\draw[black, line width=0.8pt] (axis cs:10.185,0.000143458) -- (axis cs:10.185,0.000151609);
\draw[black, line width=0.8pt] (axis cs:10.14,0.000143458) -- (axis cs:10.23,0.000143458);
\draw[black, line width=0.8pt] (axis cs:10.14,0.000151609) -- (axis cs:10.23,0.000151609);
\draw[black, line width=0.8pt] (axis cs:11.185,0.000158586) -- (axis cs:11.185,0.000168715);
\draw[black, line width=0.8pt] (axis cs:11.14,0.000158586) -- (axis cs:11.23,0.000158586);
\draw[black, line width=0.8pt] (axis cs:11.14,0.000168715) -- (axis cs:11.23,0.000168715);
\draw[black, line width=0.8pt] (axis cs:12.185,0.000157133) -- (axis cs:12.185,0.000168785);
\draw[black, line width=0.8pt] (axis cs:12.14,0.000157133) -- (axis cs:12.23,0.000157133);
\draw[black, line width=0.8pt] (axis cs:12.14,0.000168785) -- (axis cs:12.23,0.000168785);
\draw[black, line width=0.8pt] (axis cs:13.185,0.000126965) -- (axis cs:13.185,0.000130814);
\draw[black, line width=0.8pt] (axis cs:13.14,0.000126965) -- (axis cs:13.23,0.000126965);
\draw[black, line width=0.8pt] (axis cs:13.14,0.000130814) -- (axis cs:13.23,0.000130814);
\draw[black, line width=0.8pt] (axis cs:14.185,0.000133311) -- (axis cs:14.185,0.000144542);
\draw[black, line width=0.8pt] (axis cs:14.14,0.000133311) -- (axis cs:14.23,0.000133311);
\draw[black, line width=0.8pt] (axis cs:14.14,0.000144542) -- (axis cs:14.23,0.000144542);
\draw[black, line width=0.8pt] (axis cs:15.185,0.000152934) -- (axis cs:15.185,0.000177037);
\draw[black, line width=0.8pt] (axis cs:15.14,0.000152934) -- (axis cs:15.23,0.000152934);
\draw[black, line width=0.8pt] (axis cs:15.14,0.000177037) -- (axis cs:15.23,0.000177037);
\draw[black, line width=0.8pt] (axis cs:16.185,0.000159291) -- (axis cs:16.185,0.000177452);
\draw[black, line width=0.8pt] (axis cs:16.14,0.000159291) -- (axis cs:16.23,0.000159291);
\draw[black, line width=0.8pt] (axis cs:16.14,0.000177452) -- (axis cs:16.23,0.000177452);
\draw[black, line width=0.8pt] (axis cs:17.185,0.000162499) -- (axis cs:17.185,0.000168521);
\draw[black, line width=0.8pt] (axis cs:17.14,0.000162499) -- (axis cs:17.23,0.000162499);
\draw[black, line width=0.8pt] (axis cs:17.14,0.000168521) -- (axis cs:17.23,0.000168521);
\draw[black, line width=0.8pt] (axis cs:18.185,0.000132505) -- (axis cs:18.185,0.000140008);
\draw[black, line width=0.8pt] (axis cs:18.14,0.000132505) -- (axis cs:18.23,0.000132505);
\draw[black, line width=0.8pt] (axis cs:18.14,0.000140008) -- (axis cs:18.23,0.000140008);
\draw[black, line width=0.8pt] (axis cs:19.185,0.000135956) -- (axis cs:19.185,0.00014595);
\draw[black, line width=0.8pt] (axis cs:19.14,0.000135956) -- (axis cs:19.23,0.000135956);
\draw[black, line width=0.8pt] (axis cs:19.14,0.00014595) -- (axis cs:19.23,0.00014595);
\draw[black, line width=0.8pt] (axis cs:20.185,0.000144275) -- (axis cs:20.185,0.000154035);
\draw[black, line width=0.8pt] (axis cs:20.14,0.000144275) -- (axis cs:20.23,0.000144275);
\draw[black, line width=0.8pt] (axis cs:20.14,0.000154035) -- (axis cs:20.23,0.000154035);
\draw[black, line width=0.8pt] (axis cs:21.185,0.000156059) -- (axis cs:21.185,0.000168238);
\draw[black, line width=0.8pt] (axis cs:21.14,0.000156059) -- (axis cs:21.23,0.000156059);
\draw[black, line width=0.8pt] (axis cs:21.14,0.000168238) -- (axis cs:21.23,0.000168238);
\draw[black, line width=0.8pt] (axis cs:22.185,0.000143216) -- (axis cs:22.185,0.000157382);
\draw[black, line width=0.8pt] (axis cs:22.14,0.000143216) -- (axis cs:22.23,0.000143216);
\draw[black, line width=0.8pt] (axis cs:22.14,0.000157382) -- (axis cs:22.23,0.000157382);
\draw[black, line width=0.8pt] (axis cs:23.185,8.67557e-05) -- (axis cs:23.185,9.09139e-05);
\draw[black, line width=0.8pt] (axis cs:23.14,8.67557e-05) -- (axis cs:23.23,8.67557e-05);
\draw[black, line width=0.8pt] (axis cs:23.14,9.09139e-05) -- (axis cs:23.23,9.09139e-05);
\draw[black, line width=0.8pt] (axis cs:24.185,0.000101076) -- (axis cs:24.185,0.00011145);
\draw[black, line width=0.8pt] (axis cs:24.14,0.000101076) -- (axis cs:24.23,0.000101076);
\draw[black, line width=0.8pt] (axis cs:24.14,0.00011145) -- (axis cs:24.23,0.00011145);
\draw[black, line width=0.8pt] (axis cs:25.185,8.39566e-05) -- (axis cs:25.185,9.34269e-05);
\draw[black, line width=0.8pt] (axis cs:25.14,8.39566e-05) -- (axis cs:25.23,8.39566e-05);
\draw[black, line width=0.8pt] (axis cs:25.14,9.34269e-05) -- (axis cs:25.23,9.34269e-05);
\draw[black, line width=0.8pt] (axis cs:26.185,8.98869e-05) -- (axis cs:26.185,9.38385e-05);
\draw[black, line width=0.8pt] (axis cs:26.14,8.98869e-05) -- (axis cs:26.23,8.98869e-05);
\draw[black, line width=0.8pt] (axis cs:26.14,9.38385e-05) -- (axis cs:26.23,9.38385e-05);
\draw[black, line width=0.8pt] (axis cs:27.185,9.31404e-05) -- (axis cs:27.185,9.9168e-05);
\draw[black, line width=0.8pt] (axis cs:27.14,9.31404e-05) -- (axis cs:27.23,9.31404e-05);
\draw[black, line width=0.8pt] (axis cs:27.14,9.9168e-05) -- (axis cs:27.23,9.9168e-05);
\draw[black, line width=0.8pt] (axis cs:28.185,0.000110141) -- (axis cs:28.185,0.000122269);
\draw[black, line width=0.8pt] (axis cs:28.14,0.000110141) -- (axis cs:28.23,0.000110141);
\draw[black, line width=0.8pt] (axis cs:28.14,0.000122269) -- (axis cs:28.23,0.000122269);
\end{axis}
\end{tikzpicture}

%% file: figures/experiments/timings_to_po_g.tex
% Requires: \usepackage{pgfplots}
% Recommended: \pgfplotsset{compat=1.18}
\definecolor{mplBlue}{HTML}{1F77B4}
\definecolor{mplOrange}{HTML}{FF7F0E}
\definecolor{mplGreen}{HTML}{2CA02C}
\definecolor{mplRed}{HTML}{D62728}
\definecolor{mplPurple}{HTML}{9467BD}
\begin{tikzpicture}
\begin{axis}[
  width=0.96\textwidth,
  height=0.42\textwidth,
  xlabel={\scriptsize KB},
  ylabel={\scriptsize Mean time (s)},
  xmin=-0.5, xmax=35.5,
  ymode=log,
  log origin=infty,
  ymin=0.0001, ymax=10000,
  ytickten={-4,-3,-2,-1,0,1,2,3},
  minor y tick num=8,
  major tick length=2.6pt,
  minor tick length=1.8pt,
  minor tick style={black!70, line width=0.35pt},
  xtick={0,1,2,3,4,5,6,7,8,9,10,11,12,13,14,15,16,17,18,19,20,21,22,23,24,25,26,27,28,29,30,31,32,33,34,35},
  xticklabels={{1 -- Opus 4.6},{1 -- Sonnet 4.6},{1 -- GPT5.2},{1 -- GPT5.4 Mini},{1 -- Qwen},{1 -- Llama},{2 -- Opus 4.6},{2 -- Sonnet 4.6},{2 -- GPT5.2},{2 -- GPT5.4 Mini},{2 -- Qwen},{2 -- Llama},{3 -- Opus 4.6},{3 -- Sonnet 4.6},{3 -- GPT5.2},{3 -- GPT5.4 Mini},{3 -- Qwen},{3 -- Llama},{4 -- Opus 4.6},{4 -- Sonnet 4.6},{4 -- GPT5.2},{4 -- GPT5.4 Mini},{4 -- Qwen},{4 -- Llama},{5 -- Opus 4.6},{5 -- Sonnet 4.6},{5 -- GPT5.2},{5 -- GPT5.4 Mini},{5 -- Qwen},{5 -- Llama},{6 -- Opus 4.6},{6 -- Sonnet 4.6},{6 -- GPT5.2},{6 -- GPT5.4 Mini},{6 -- Qwen},{6 -- Llama}},
  tick align=outside,
  tick pos=left,
  x tick label style={rotate=90, anchor=east, font=\scriptsize},
  y tick label style={font=\scriptsize},
  label style={font=\small},
  ymajorgrids=true,
  grid style={dotted, gray!45},
  minor grid style={dotted, gray!25},
  legend style={at={(0.02,0.98)}, anchor=north west, draw=black!25, fill=white, fill opacity=0.88, text opacity=1, font=\scriptsize},
  legend cell align={left},
  axis line style={black!70},
  clip=false
]
\addplot[ybar, area legend, bar width=0.3256, fill=mplBlue, draw=black, line width=0.4pt] coordinates {(-0.185,0.156105) (0.815,0.16629) (1.815,0.183808) (2.815,2.45363) (3.815,0.143412) (4.815,0.246694) (5.815,35.2801) (6.815,37.908) (7.815,35.2407) (8.815,37.3384) (9.815,26.4839) (10.815,31.8165) (11.815,84.6565) (12.815,148.558) (13.815,157.605) (14.815,77.6321) (15.815,134.276) (16.815,118.695) (17.815,23.7966) (18.815,25.4595) (19.815,28.66) (20.815,21.6864) (21.815,25.2426) (22.815,26.2028) (23.815,1.13469) (24.815,1.03968) (25.815,11.3286) (26.815,0.928423) (27.815,0.88954) (28.815,1.12441) (29.815,32.585) (30.815,27.1489) (31.815,72.4745) (32.815,3.30654) (33.815,33.7885) (34.815,31.1359)};
\addlegendentry{Total order}
\addplot[ybar, area legend, bar width=0.3256, fill=mplOrange, draw=black, line width=0.4pt] coordinates {(0.185,0.000281143) (1.185,0.000236773) (2.185,0.000292397) (3.185,0.000241089) (4.185,0.00023756) (5.185,0.00020504) (6.185,0.000457931) (7.185,0.000426245) (8.185,0.000414109) (9.185,0.000389576) (10.185,0.000368643) (11.185,0.000343513) (12.185,0.000540471) (13.185,0.000562) (14.185,0.000520039) (15.185,0.000581765) (16.185,0.000439954) (17.185,0.000437832) (18.185,0.000466895) (19.185,0.000573444) (20.185,0.000510526) (21.185,0.000545835) (22.185,0.00050149) (23.185,0.00046196) (24.185,0.000427675) (25.185,0.000324011) (26.185,0.000335717) (27.185,0.000366187) (28.185,0.000278115) (29.185,0.000291967) (30.185,0.000823474) (31.185,0.000715184) (32.185,0.0006814) (33.185,0.000774336) (34.185,0.000476313) (35.185,0.000547934)};
\addlegendentry{Low-level order}
\draw[black, line width=0.8pt] (axis cs:-0.185,0.153496) -- (axis cs:-0.185,0.158715);
\draw[black, line width=0.8pt] (axis cs:-0.23,0.153496) -- (axis cs:-0.14,0.153496);
\draw[black, line width=0.8pt] (axis cs:-0.23,0.158715) -- (axis cs:-0.14,0.158715);
\draw[black, line width=0.8pt] (axis cs:0.815,0.164962) -- (axis cs:0.815,0.167619);
\draw[black, line width=0.8pt] (axis cs:0.77,0.164962) -- (axis cs:0.86,0.164962);
\draw[black, line width=0.8pt] (axis cs:0.77,0.167619) -- (axis cs:0.86,0.167619);
\draw[black, line width=0.8pt] (axis cs:1.815,0.182576) -- (axis cs:1.815,0.18504);
\draw[black, line width=0.8pt] (axis cs:1.77,0.182576) -- (axis cs:1.86,0.182576);
\draw[black, line width=0.8pt] (axis cs:1.77,0.18504) -- (axis cs:1.86,0.18504);
\draw[black, line width=0.8pt] (axis cs:2.815,2.36391) -- (axis cs:2.815,2.54336);
\draw[black, line width=0.8pt] (axis cs:2.77,2.36391) -- (axis cs:2.86,2.36391);
\draw[black, line width=0.8pt] (axis cs:2.77,2.54336) -- (axis cs:2.86,2.54336);
\draw[black, line width=0.8pt] (axis cs:3.815,0.142388) -- (axis cs:3.815,0.144436);
\draw[black, line width=0.8pt] (axis cs:3.77,0.142388) -- (axis cs:3.86,0.142388);
\draw[black, line width=0.8pt] (axis cs:3.77,0.144436) -- (axis cs:3.86,0.144436);
\draw[black, line width=0.8pt] (axis cs:4.815,0.234447) -- (axis cs:4.815,0.258941);
\draw[black, line width=0.8pt] (axis cs:4.77,0.234447) -- (axis cs:4.86,0.234447);
\draw[black, line width=0.8pt] (axis cs:4.77,0.258941) -- (axis cs:4.86,0.258941);
\draw[black, line width=0.8pt] (axis cs:5.815,33.6647) -- (axis cs:5.815,36.8955);
\draw[black, line width=0.8pt] (axis cs:5.77,33.6647) -- (axis cs:5.86,33.6647);
\draw[black, line width=0.8pt] (axis cs:5.77,36.8955) -- (axis cs:5.86,36.8955);
\draw[black, line width=0.8pt] (axis cs:6.815,37.411) -- (axis cs:6.815,38.4049);
\draw[black, line width=0.8pt] (axis cs:6.77,37.411) -- (axis cs:6.86,37.411);
\draw[black, line width=0.8pt] (axis cs:6.77,38.4049) -- (axis cs:6.86,38.4049);
\draw[black, line width=0.8pt] (axis cs:7.815,33.5465) -- (axis cs:7.815,36.935);
\draw[black, line width=0.8pt] (axis cs:7.77,33.5465) -- (axis cs:7.86,33.5465);
\draw[black, line width=0.8pt] (axis cs:7.77,36.935) -- (axis cs:7.86,36.935);
\draw[black, line width=0.8pt] (axis cs:8.815,35.7163) -- (axis cs:8.815,38.9605);
\draw[black, line width=0.8pt] (axis cs:8.77,35.7163) -- (axis cs:8.86,35.7163);
\draw[black, line width=0.8pt] (axis cs:8.77,38.9605) -- (axis cs:8.86,38.9605);
\draw[black, line width=0.8pt] (axis cs:9.815,25.5794) -- (axis cs:9.815,27.3885);
\draw[black, line width=0.8pt] (axis cs:9.77,25.5794) -- (axis cs:9.86,25.5794);
\draw[black, line width=0.8pt] (axis cs:9.77,27.3885) -- (axis cs:9.86,27.3885);
\draw[black, line width=0.8pt] (axis cs:10.815,30.5034) -- (axis cs:10.815,33.1296);
\draw[black, line width=0.8pt] (axis cs:10.77,30.5034) -- (axis cs:10.86,30.5034);
\draw[black, line width=0.8pt] (axis cs:10.77,33.1296) -- (axis cs:10.86,33.1296);
\draw[black, line width=0.8pt] (axis cs:11.815,81.7229) -- (axis cs:11.815,87.5901);
\draw[black, line width=0.8pt] (axis cs:11.77,81.7229) -- (axis cs:11.86,81.7229);
\draw[black, line width=0.8pt] (axis cs:11.77,87.5901) -- (axis cs:11.86,87.5901);
\draw[black, line width=0.8pt] (axis cs:12.815,133.549) -- (axis cs:12.815,163.567);
\draw[black, line width=0.8pt] (axis cs:12.77,133.549) -- (axis cs:12.86,133.549);
\draw[black, line width=0.8pt] (axis cs:12.77,163.567) -- (axis cs:12.86,163.567);
\draw[black, line width=0.8pt] (axis cs:13.815,150.727) -- (axis cs:13.815,164.483);
\draw[black, line width=0.8pt] (axis cs:13.77,150.727) -- (axis cs:13.86,150.727);
\draw[black, line width=0.8pt] (axis cs:13.77,164.483) -- (axis cs:13.86,164.483);
\draw[black, line width=0.8pt] (axis cs:14.815,74.2376) -- (axis cs:14.815,81.0266);
\draw[black, line width=0.8pt] (axis cs:14.77,74.2376) -- (axis cs:14.86,74.2376);
\draw[black, line width=0.8pt] (axis cs:14.77,81.0266) -- (axis cs:14.86,81.0266);
\draw[black, line width=0.8pt] (axis cs:15.815,128.833) -- (axis cs:15.815,139.719);
\draw[black, line width=0.8pt] (axis cs:15.77,128.833) -- (axis cs:15.86,128.833);
\draw[black, line width=0.8pt] (axis cs:15.77,139.719) -- (axis cs:15.86,139.719);
\draw[black, line width=0.8pt] (axis cs:16.815,115.527) -- (axis cs:16.815,121.863);
\draw[black, line width=0.8pt] (axis cs:16.77,115.527) -- (axis cs:16.86,115.527);
\draw[black, line width=0.8pt] (axis cs:16.77,121.863) -- (axis cs:16.86,121.863);
\draw[black, line width=0.8pt] (axis cs:17.815,23.0435) -- (axis cs:17.815,24.5496);
\draw[black, line width=0.8pt] (axis cs:17.77,23.0435) -- (axis cs:17.86,23.0435);
\draw[black, line width=0.8pt] (axis cs:17.77,24.5496) -- (axis cs:17.86,24.5496);
\draw[black, line width=0.8pt] (axis cs:18.815,24.5733) -- (axis cs:18.815,26.3456);
\draw[black, line width=0.8pt] (axis cs:18.77,24.5733) -- (axis cs:18.86,24.5733);
\draw[black, line width=0.8pt] (axis cs:18.77,26.3456) -- (axis cs:18.86,26.3456);
\draw[black, line width=0.8pt] (axis cs:19.815,27.2811) -- (axis cs:19.815,30.0388);
\draw[black, line width=0.8pt] (axis cs:19.77,27.2811) -- (axis cs:19.86,27.2811);
\draw[black, line width=0.8pt] (axis cs:19.77,30.0388) -- (axis cs:19.86,30.0388);
\draw[black, line width=0.8pt] (axis cs:20.815,19.3149) -- (axis cs:20.815,24.0578);
\draw[black, line width=0.8pt] (axis cs:20.77,19.3149) -- (axis cs:20.86,19.3149);
\draw[black, line width=0.8pt] (axis cs:20.77,24.0578) -- (axis cs:20.86,24.0578);
\draw[black, line width=0.8pt] (axis cs:21.815,24.5333) -- (axis cs:21.815,25.9519);
\draw[black, line width=0.8pt] (axis cs:21.77,24.5333) -- (axis cs:21.86,24.5333);
\draw[black, line width=0.8pt] (axis cs:21.77,25.9519) -- (axis cs:21.86,25.9519);
\draw[black, line width=0.8pt] (axis cs:22.815,25.3803) -- (axis cs:22.815,27.0253);
\draw[black, line width=0.8pt] (axis cs:22.77,25.3803) -- (axis cs:22.86,25.3803);
\draw[black, line width=0.8pt] (axis cs:22.77,27.0253) -- (axis cs:22.86,27.0253);
\draw[black, line width=0.8pt] (axis cs:23.815,1.09605) -- (axis cs:23.815,1.17334);
\draw[black, line width=0.8pt] (axis cs:23.77,1.09605) -- (axis cs:23.86,1.09605);
\draw[black, line width=0.8pt] (axis cs:23.77,1.17334) -- (axis cs:23.86,1.17334);
\draw[black, line width=0.8pt] (axis cs:24.815,0.985968) -- (axis cs:24.815,1.09339);
\draw[black, line width=0.8pt] (axis cs:24.77,0.985968) -- (axis cs:24.86,0.985968);
\draw[black, line width=0.8pt] (axis cs:24.77,1.09339) -- (axis cs:24.86,1.09339);
\draw[black, line width=0.8pt] (axis cs:25.815,10.7877) -- (axis cs:25.815,11.8695);
\draw[black, line width=0.8pt] (axis cs:25.77,10.7877) -- (axis cs:25.86,10.7877);
\draw[black, line width=0.8pt] (axis cs:25.77,11.8695) -- (axis cs:25.86,11.8695);
\draw[black, line width=0.8pt] (axis cs:26.815,0.904029) -- (axis cs:26.815,0.952817);
\draw[black, line width=0.8pt] (axis cs:26.77,0.904029) -- (axis cs:26.86,0.904029);
\draw[black, line width=0.8pt] (axis cs:26.77,0.952817) -- (axis cs:26.86,0.952817);
\draw[black, line width=0.8pt] (axis cs:27.815,0.882786) -- (axis cs:27.815,0.896294);
\draw[black, line width=0.8pt] (axis cs:27.77,0.882786) -- (axis cs:27.86,0.882786);
\draw[black, line width=0.8pt] (axis cs:27.77,0.896294) -- (axis cs:27.86,0.896294);
\draw[black, line width=0.8pt] (axis cs:28.815,1.0788) -- (axis cs:28.815,1.17002);
\draw[black, line width=0.8pt] (axis cs:28.77,1.0788) -- (axis cs:28.86,1.0788);
\draw[black, line width=0.8pt] (axis cs:28.77,1.17002) -- (axis cs:28.86,1.17002);
\draw[black, line width=0.8pt] (axis cs:29.815,31.3954) -- (axis cs:29.815,33.7747);
\draw[black, line width=0.8pt] (axis cs:29.77,31.3954) -- (axis cs:29.86,31.3954);
\draw[black, line width=0.8pt] (axis cs:29.77,33.7747) -- (axis cs:29.86,33.7747);
\draw[black, line width=0.8pt] (axis cs:30.815,26.1501) -- (axis cs:30.815,28.1476);
\draw[black, line width=0.8pt] (axis cs:30.77,26.1501) -- (axis cs:30.86,26.1501);
\draw[black, line width=0.8pt] (axis cs:30.77,28.1476) -- (axis cs:30.86,28.1476);
\draw[black, line width=0.8pt] (axis cs:31.815,64.2719) -- (axis cs:31.815,80.6771);
\draw[black, line width=0.8pt] (axis cs:31.77,64.2719) -- (axis cs:31.86,64.2719);
\draw[black, line width=0.8pt] (axis cs:31.77,80.6771) -- (axis cs:31.86,80.6771);
\draw[black, line width=0.8pt] (axis cs:32.815,3.17812) -- (axis cs:32.815,3.43496);
\draw[black, line width=0.8pt] (axis cs:32.77,3.17812) -- (axis cs:32.86,3.17812);
\draw[black, line width=0.8pt] (axis cs:32.77,3.43496) -- (axis cs:32.86,3.43496);
\draw[black, line width=0.8pt] (axis cs:33.815,32.3871) -- (axis cs:33.815,35.19);
\draw[black, line width=0.8pt] (axis cs:33.77,32.3871) -- (axis cs:33.86,32.3871);
\draw[black, line width=0.8pt] (axis cs:33.77,35.19) -- (axis cs:33.86,35.19);
\draw[black, line width=0.8pt] (axis cs:34.815,29.6623) -- (axis cs:34.815,32.6095);
\draw[black, line width=0.8pt] (axis cs:34.77,29.6623) -- (axis cs:34.86,29.6623);
\draw[black, line width=0.8pt] (axis cs:34.77,32.6095) -- (axis cs:34.86,32.6095);
\draw[black, line width=0.8pt] (axis cs:0.185,0.000274442) -- (axis cs:0.185,0.000287844);
\draw[black, line width=0.8pt] (axis cs:0.14,0.000274442) -- (axis cs:0.23,0.000274442);
\draw[black, line width=0.8pt] (axis cs:0.14,0.000287844) -- (axis cs:0.23,0.000287844);
\draw[black, line width=0.8pt] (axis cs:1.185,0.000234279) -- (axis cs:1.185,0.000239268);
\draw[black, line width=0.8pt] (axis cs:1.14,0.000234279) -- (axis cs:1.23,0.000234279);
\draw[black, line width=0.8pt] (axis cs:1.14,0.000239268) -- (axis cs:1.23,0.000239268);
\draw[black, line width=0.8pt] (axis cs:2.185,0.000246535) -- (axis cs:2.185,0.000338258);
\draw[black, line width=0.8pt] (axis cs:2.14,0.000246535) -- (axis cs:2.23,0.000246535);
\draw[black, line width=0.8pt] (axis cs:2.14,0.000338258) -- (axis cs:2.23,0.000338258);
\draw[black, line width=0.8pt] (axis cs:3.185,0.000234166) -- (axis cs:3.185,0.000248011);
\draw[black, line width=0.8pt] (axis cs:3.14,0.000234166) -- (axis cs:3.23,0.000234166);
\draw[black, line width=0.8pt] (axis cs:3.14,0.000248011) -- (axis cs:3.23,0.000248011);
\draw[black, line width=0.8pt] (axis cs:4.185,0.0002332) -- (axis cs:4.185,0.000241921);
\draw[black, line width=0.8pt] (axis cs:4.14,0.0002332) -- (axis cs:4.23,0.0002332);
\draw[black, line width=0.8pt] (axis cs:4.14,0.000241921) -- (axis cs:4.23,0.000241921);
\draw[black, line width=0.8pt] (axis cs:5.185,0.000201277) -- (axis cs:5.185,0.000208803);
\draw[black, line width=0.8pt] (axis cs:5.14,0.000201277) -- (axis cs:5.23,0.000201277);
\draw[black, line width=0.8pt] (axis cs:5.14,0.000208803) -- (axis cs:5.23,0.000208803);
\draw[black, line width=0.8pt] (axis cs:6.185,0.000446291) -- (axis cs:6.185,0.00046957);
\draw[black, line width=0.8pt] (axis cs:6.14,0.000446291) -- (axis cs:6.23,0.000446291);
\draw[black, line width=0.8pt] (axis cs:6.14,0.00046957) -- (axis cs:6.23,0.00046957);
\draw[black, line width=0.8pt] (axis cs:7.185,0.000417539) -- (axis cs:7.185,0.00043495);
\draw[black, line width=0.8pt] (axis cs:7.14,0.000417539) -- (axis cs:7.23,0.000417539);
\draw[black, line width=0.8pt] (axis cs:7.14,0.00043495) -- (axis cs:7.23,0.00043495);
\draw[black, line width=0.8pt] (axis cs:8.185,0.000397) -- (axis cs:8.185,0.000431218);
\draw[black, line width=0.8pt] (axis cs:8.14,0.000397) -- (axis cs:8.23,0.000397);
\draw[black, line width=0.8pt] (axis cs:8.14,0.000431218) -- (axis cs:8.23,0.000431218);
\draw[black, line width=0.8pt] (axis cs:9.185,0.000380916) -- (axis cs:9.185,0.000398236);
\draw[black, line width=0.8pt] (axis cs:9.14,0.000380916) -- (axis cs:9.23,0.000380916);
\draw[black, line width=0.8pt] (axis cs:9.14,0.000398236) -- (axis cs:9.23,0.000398236);
\draw[black, line width=0.8pt] (axis cs:10.185,0.000357795) -- (axis cs:10.185,0.000379491);
\draw[black, line width=0.8pt] (axis cs:10.14,0.000357795) -- (axis cs:10.23,0.000357795);
\draw[black, line width=0.8pt] (axis cs:10.14,0.000379491) -- (axis cs:10.23,0.000379491);
\draw[black, line width=0.8pt] (axis cs:11.185,0.000332851) -- (axis cs:11.185,0.000354176);
\draw[black, line width=0.8pt] (axis cs:11.14,0.000332851) -- (axis cs:11.23,0.000332851);
\draw[black, line width=0.8pt] (axis cs:11.14,0.000354176) -- (axis cs:11.23,0.000354176);
\draw[black, line width=0.8pt] (axis cs:12.185,0.000524707) -- (axis cs:12.185,0.000556235);
\draw[black, line width=0.8pt] (axis cs:12.14,0.000524707) -- (axis cs:12.23,0.000524707);
\draw[black, line width=0.8pt] (axis cs:12.14,0.000556235) -- (axis cs:12.23,0.000556235);
\draw[black, line width=0.8pt] (axis cs:13.185,0.000531786) -- (axis cs:13.185,0.000592214);
\draw[black, line width=0.8pt] (axis cs:13.14,0.000531786) -- (axis cs:13.23,0.000531786);
\draw[black, line width=0.8pt] (axis cs:13.14,0.000592214) -- (axis cs:13.23,0.000592214);
\draw[black, line width=0.8pt] (axis cs:14.185,0.000487547) -- (axis cs:14.185,0.00055253);
\draw[black, line width=0.8pt] (axis cs:14.14,0.000487547) -- (axis cs:14.23,0.000487547);
\draw[black, line width=0.8pt] (axis cs:14.14,0.00055253) -- (axis cs:14.23,0.00055253);
\draw[black, line width=0.8pt] (axis cs:15.185,0.000559221) -- (axis cs:15.185,0.00060431);
\draw[black, line width=0.8pt] (axis cs:15.14,0.000559221) -- (axis cs:15.23,0.000559221);
\draw[black, line width=0.8pt] (axis cs:15.14,0.00060431) -- (axis cs:15.23,0.00060431);
\draw[black, line width=0.8pt] (axis cs:16.185,0.000355836) -- (axis cs:16.185,0.000524072);
\draw[black, line width=0.8pt] (axis cs:16.14,0.000355836) -- (axis cs:16.23,0.000355836);
\draw[black, line width=0.8pt] (axis cs:16.14,0.000524072) -- (axis cs:16.23,0.000524072);
\draw[black, line width=0.8pt] (axis cs:17.185,0.000431142) -- (axis cs:17.185,0.000444522);
\draw[black, line width=0.8pt] (axis cs:17.14,0.000431142) -- (axis cs:17.23,0.000431142);
\draw[black, line width=0.8pt] (axis cs:17.14,0.000444522) -- (axis cs:17.23,0.000444522);
\draw[black, line width=0.8pt] (axis cs:18.185,0.000455878) -- (axis cs:18.185,0.000477912);
\draw[black, line width=0.8pt] (axis cs:18.14,0.000455878) -- (axis cs:18.23,0.000455878);
\draw[black, line width=0.8pt] (axis cs:18.14,0.000477912) -- (axis cs:18.23,0.000477912);
\draw[black, line width=0.8pt] (axis cs:19.185,0.000540625) -- (axis cs:19.185,0.000606264);
\draw[black, line width=0.8pt] (axis cs:19.14,0.000540625) -- (axis cs:19.23,0.000540625);
\draw[black, line width=0.8pt] (axis cs:19.14,0.000606264) -- (axis cs:19.23,0.000606264);
\draw[black, line width=0.8pt] (axis cs:20.185,0.00049456) -- (axis cs:20.185,0.000526492);
\draw[black, line width=0.8pt] (axis cs:20.14,0.00049456) -- (axis cs:20.23,0.00049456);
\draw[black, line width=0.8pt] (axis cs:20.14,0.000526492) -- (axis cs:20.23,0.000526492);
\draw[black, line width=0.8pt] (axis cs:21.185,0.000518809) -- (axis cs:21.185,0.000572862);
\draw[black, line width=0.8pt] (axis cs:21.14,0.000518809) -- (axis cs:21.23,0.000518809);
\draw[black, line width=0.8pt] (axis cs:21.14,0.000572862) -- (axis cs:21.23,0.000572862);
\draw[black, line width=0.8pt] (axis cs:22.185,0.000486968) -- (axis cs:22.185,0.000516012);
\draw[black, line width=0.8pt] (axis cs:22.14,0.000486968) -- (axis cs:22.23,0.000486968);
\draw[black, line width=0.8pt] (axis cs:22.14,0.000516012) -- (axis cs:22.23,0.000516012);
\draw[black, line width=0.8pt] (axis cs:23.185,0.000403483) -- (axis cs:23.185,0.000520437);
\draw[black, line width=0.8pt] (axis cs:23.14,0.000403483) -- (axis cs:23.23,0.000403483);
\draw[black, line width=0.8pt] (axis cs:23.14,0.000520437) -- (axis cs:23.23,0.000520437);
\draw[black, line width=0.8pt] (axis cs:24.185,0.000422041) -- (axis cs:24.185,0.000433309);
\draw[black, line width=0.8pt] (axis cs:24.14,0.000422041) -- (axis cs:24.23,0.000422041);
\draw[black, line width=0.8pt] (axis cs:24.14,0.000433309) -- (axis cs:24.23,0.000433309);
\draw[black, line width=0.8pt] (axis cs:25.185,0.000312861) -- (axis cs:25.185,0.00033516);
\draw[black, line width=0.8pt] (axis cs:25.14,0.000312861) -- (axis cs:25.23,0.000312861);
\draw[black, line width=0.8pt] (axis cs:25.14,0.00033516) -- (axis cs:25.23,0.00033516);
\draw[black, line width=0.8pt] (axis cs:26.185,0.000327295) -- (axis cs:26.185,0.000344139);
\draw[black, line width=0.8pt] (axis cs:26.14,0.000327295) -- (axis cs:26.23,0.000327295);
\draw[black, line width=0.8pt] (axis cs:26.14,0.000344139) -- (axis cs:26.23,0.000344139);
\draw[black, line width=0.8pt] (axis cs:27.185,0.000356929) -- (axis cs:27.185,0.000375445);
\draw[black, line width=0.8pt] (axis cs:27.14,0.000356929) -- (axis cs:27.23,0.000356929);
\draw[black, line width=0.8pt] (axis cs:27.14,0.000375445) -- (axis cs:27.23,0.000375445);
\draw[black, line width=0.8pt] (axis cs:28.185,0.00023349) -- (axis cs:28.185,0.000322741);
\draw[black, line width=0.8pt] (axis cs:28.14,0.00023349) -- (axis cs:28.23,0.00023349);
\draw[black, line width=0.8pt] (axis cs:28.14,0.000322741) -- (axis cs:28.23,0.000322741);
\draw[black, line width=0.8pt] (axis cs:29.185,0.0002856) -- (axis cs:29.185,0.000298334);
\draw[black, line width=0.8pt] (axis cs:29.14,0.0002856) -- (axis cs:29.23,0.0002856);
\draw[black, line width=0.8pt] (axis cs:29.14,0.000298334) -- (axis cs:29.23,0.000298334);
\draw[black, line width=0.8pt] (axis cs:30.185,0.000803391) -- (axis cs:30.185,0.000843557);
\draw[black, line width=0.8pt] (axis cs:30.14,0.000803391) -- (axis cs:30.23,0.000803391);
\draw[black, line width=0.8pt] (axis cs:30.14,0.000843557) -- (axis cs:30.23,0.000843557);
\draw[black, line width=0.8pt] (axis cs:31.185,0.000692644) -- (axis cs:31.185,0.000737724);
\draw[black, line width=0.8pt] (axis cs:31.14,0.000692644) -- (axis cs:31.23,0.000692644);
\draw[black, line width=0.8pt] (axis cs:31.14,0.000737724) -- (axis cs:31.23,0.000737724);
\draw[black, line width=0.8pt] (axis cs:32.185,0.000642834) -- (axis cs:32.185,0.000719967);
\draw[black, line width=0.8pt] (axis cs:32.14,0.000642834) -- (axis cs:32.23,0.000642834);
\draw[black, line width=0.8pt] (axis cs:32.14,0.000719967) -- (axis cs:32.23,0.000719967);
\draw[black, line width=0.8pt] (axis cs:33.185,0.000744145) -- (axis cs:33.185,0.000804526);
\draw[black, line width=0.8pt] (axis cs:33.14,0.000744145) -- (axis cs:33.23,0.000744145);
\draw[black, line width=0.8pt] (axis cs:33.14,0.000804526) -- (axis cs:33.23,0.000804526);
\draw[black, line width=0.8pt] (axis cs:34.185,0.000462459) -- (axis cs:34.185,0.000490166);
\draw[black, line width=0.8pt] (axis cs:34.14,0.000462459) -- (axis cs:34.23,0.000462459);
\draw[black, line width=0.8pt] (axis cs:34.14,0.000490166) -- (axis cs:34.23,0.000490166);
\draw[black, line width=0.8pt] (axis cs:35.185,0.000528546) -- (axis cs:35.185,0.000567321);
\draw[black, line width=0.8pt] (axis cs:35.14,0.000528546) -- (axis cs:35.23,0.000528546);
\draw[black, line width=0.8pt] (axis cs:35.14,0.000567321) -- (axis cs:35.23,0.000567321);
\end{axis}
\end{tikzpicture}

%% file: figures/experiments/timings_normalized_to.tex
% Requires: \usepackage{pgfplots}
% Recommended: \pgfplotsset{compat=1.18}
\begin{tikzpicture}
\begin{axis}[
  width=0.96\textwidth,
  height=0.42\textwidth,
  xlabel={Instance},
  ylabel={Normalized time},
  xmin=-0.5, xmax=10.5,
  ymin=-0.5, ymax=4.92675,
  xtick={0,1,2,3,4,5,6,7,8,9,10},
  ytick={0,1,2,3,4},
  xticklabels={{Blocks World 1},{Blocks World 2},{Blocks World 3},{Blocks World 4},{Blocks World 6},{Grippers 1},{Grippers 2},{Grippers 3},{Grippers 4},{Grippers 5},{Grippers 6}},
  tick align=outside,
  tick pos=left,
  x tick label style={rotate=90, anchor=east, font=\scriptsize},
  y tick label style={font=\scriptsize},
  label style={font=\small},
  ymajorgrids=true,
  grid style={dotted, gray!45},
  legend style={at={(1.02,0.98)}, anchor=north west, draw=black!25, fill=white, fill opacity=0.88, text opacity=1, font=\scriptsize},
  legend cell align={left},
  axis line style={black!70},
  clip=false
]
\addplot[gray!35, thin, forget plot] coordinates {(0,-0.5) (0,4.92055)};
\addplot[gray!35, thin, forget plot] coordinates {(1,-0.5) (1,4.92055)};
\addplot[gray!35, thin, forget plot] coordinates {(2,-0.5) (2,4.92055)};
\addplot[gray!35, thin, forget plot] coordinates {(3,-0.5) (3,4.92055)};
\addplot[gray!35, thin, forget plot] coordinates {(4,-0.5) (4,4.92055)};
\addplot[gray!35, thin, forget plot] coordinates {(5,-0.5) (5,4.92055)};
\addplot[gray!35, thin, forget plot] coordinates {(6,-0.5) (6,4.92055)};
\addplot[gray!35, thin, forget plot] coordinates {(7,-0.5) (7,4.92055)};
\addplot[gray!35, thin, forget plot] coordinates {(8,-0.5) (8,4.92055)};
\addplot[gray!35, thin, forget plot] coordinates {(9,-0.5) (9,4.92055)};
\addplot[gray!35, thin, forget plot] coordinates {(10,-0.5) (10,4.92055)};
\addplot[black!50, dashed, forget plot] coordinates {(-0.5,1) (10.5,1)};
\addplot+[only marks, mark=*, color=blue!70!black, mark options={solid, draw=black, line width=0.25pt}, mark size=2.2pt] coordinates {(-0.045,0.247074) (0.955,1.82224) (1.955,0.0120878) (2.955,1.05302) (3.955,1.26671) (4.955,0.329214) (5.955,1.03615) (6.955,1.31079) (7.955,1.13845) (8.955,4.13318) (9.955,2.16947)};
\addlegendentry{GPT5.2}
\addplot+[only marks, mark=square*, color=red!75!black, mark options={solid, draw=black, line width=0.25pt}, mark size=2.2pt] coordinates {(-0.027,0.972826) (1.973,1.98124) (2.973,0.899156) (3.973,0.136342) (4.973,4.39464) (5.973,1.09782) (6.973,0.645659) (7.973,0.861437) (8.973,0.338731) (9.973,0.0989787)};
\addlegendentry{GPT5.4 Mini}
\addplot+[only marks, mark=triangle*, color=teal!75!black, mark options={solid, draw=black, line width=0.25pt}, mark size=2.2pt] coordinates {(-0.009,0.959344) (0.991,1.14254) (1.991,0.0739886) (2.991,1.05318) (3.991,0.703037) (4.991,0.297839) (5.991,1.11457) (6.991,1.23554) (7.991,1.01131) (8.991,0.379323) (9.991,0.81268)};
\addlegendentry{Sonnet 4.6}
\addplot+[only marks, mark=diamond*, color=orange!85!black, mark options={solid, draw=black, line width=0.25pt}, mark size=2.2pt] coordinates {(0.009,1.46981) (1.009,1.78203) (2.009,0.0738339) (3.009,1.0646) (4.009,1.57303) (5.009,0.279596) (6.009,1.03731) (7.009,0.70408) (8.009,0.945259) (9.009,0.413988) (10.009,0.975408)};
\addlegendentry{Opus 4.6}
\addplot+[only marks, mark=pentagon*, color=purple!75!black, mark options={solid, draw=black, line width=0.25pt}, mark size=2.2pt] coordinates {(0.027,1.37998) (1.027,0.0841969) (2.027,2.61269) (3.027,1.05516) (4.027,1.32521) (5.027,0.256862) (6.027,0.778681) (7.027,1.11676) (8.027,1.0027) (9.027,0.324545) (10.027,1.01143)};
\addlegendentry{Qwen}
\addplot+[only marks, mark=otimes*, color=brown!80!black, mark options={solid, draw=black, line width=0.25pt}, mark size=2.2pt] coordinates {(0.045,0.970964) (1.045,0.168998) (2.045,1.24616) (3.045,0.874881) (4.045,0.995676) (5.045,0.441847) (6.045,0.935468) (7.045,0.987174) (8.045,1.04084) (9.045,0.410236) (10.045,0.932029)};
\addlegendentry{Llama}
\end{axis}
\end{tikzpicture}

%% file: sections/5-experiments/e-real_experiment.tex
\begin{figure*}[t]
    \centering
    \includegraphics[width=0.48\linewidth]{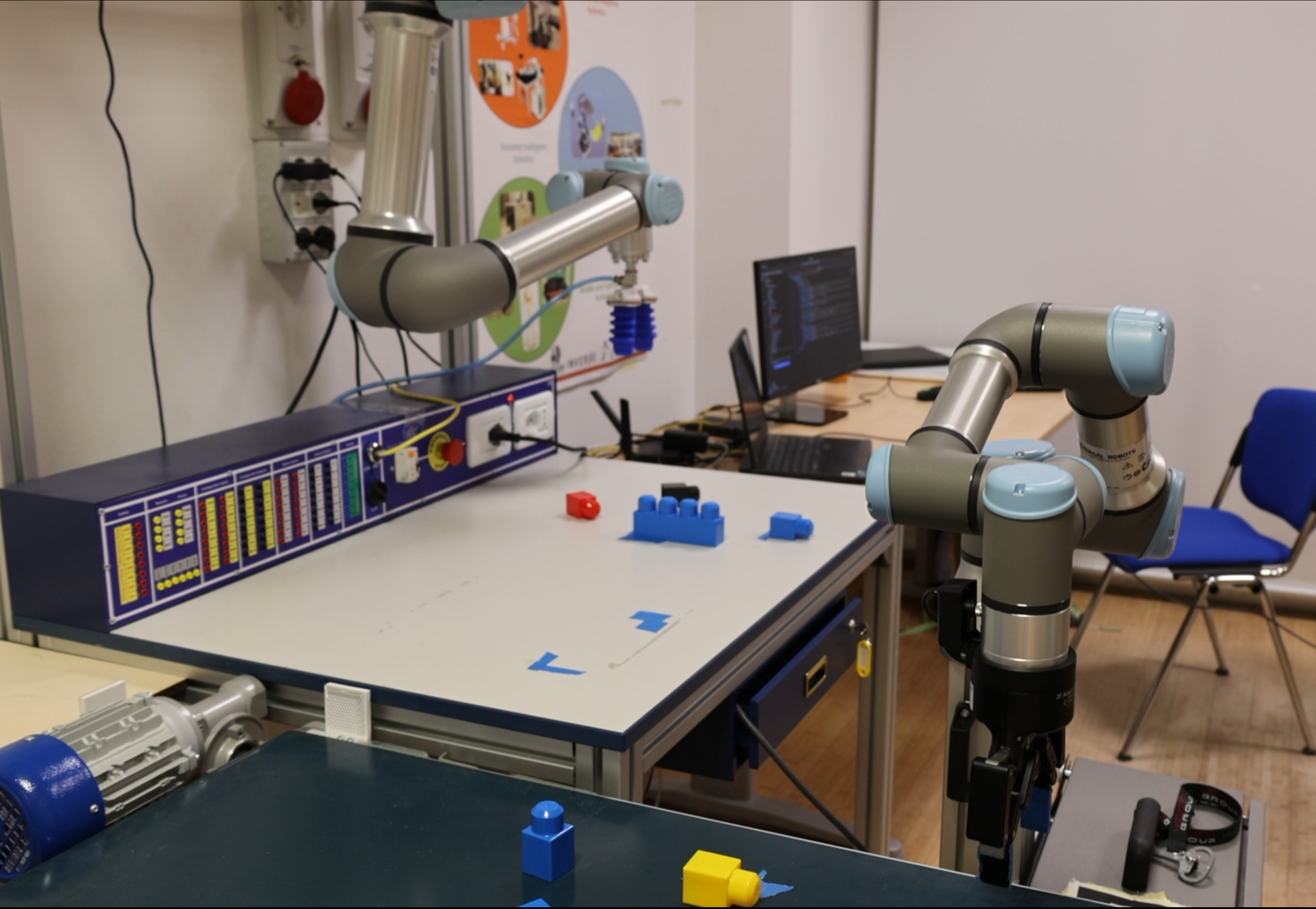}
    % \hfill
    \includegraphics[width=0.48\linewidth]{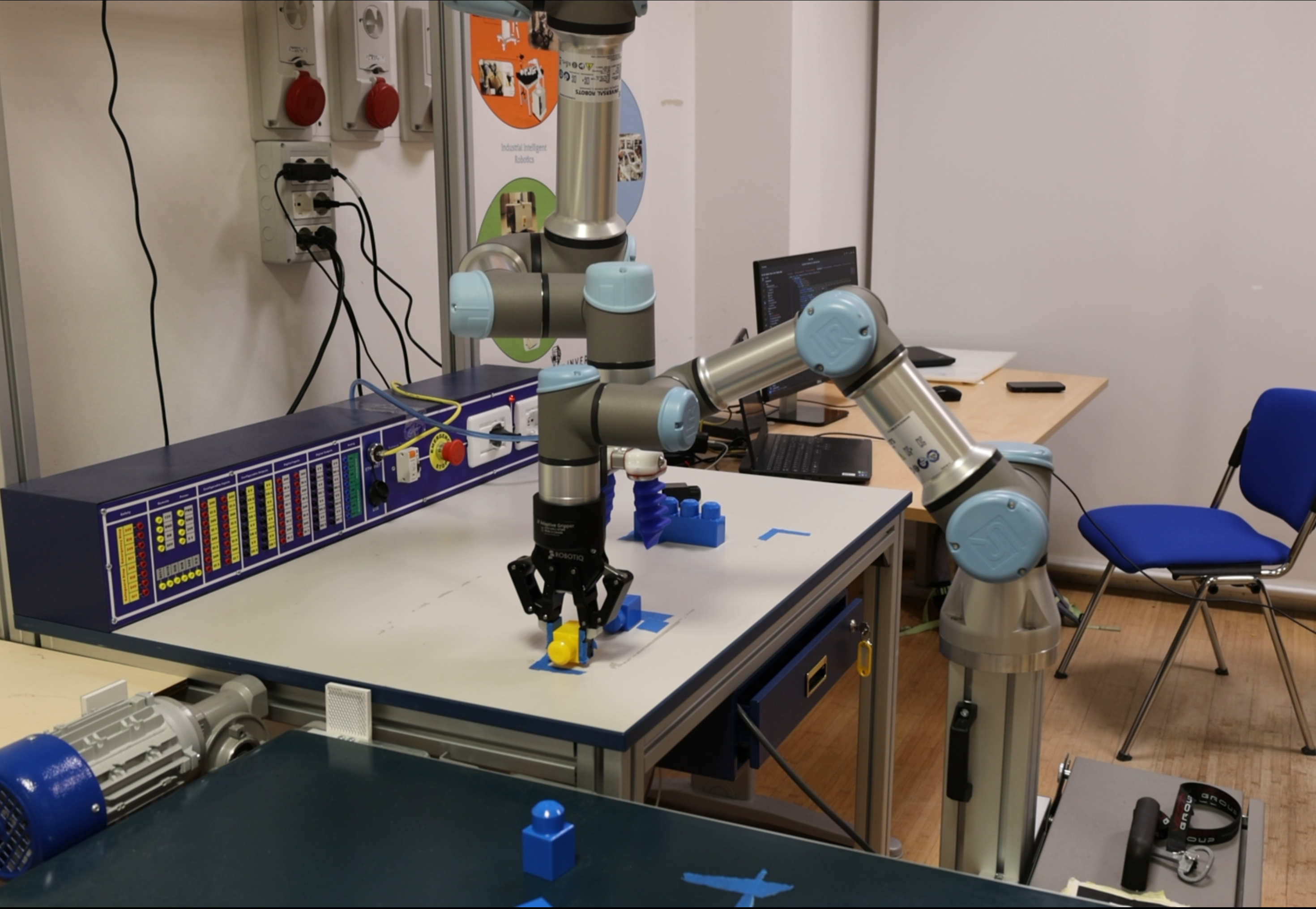}\\
    \includegraphics[width=0.48\linewidth]{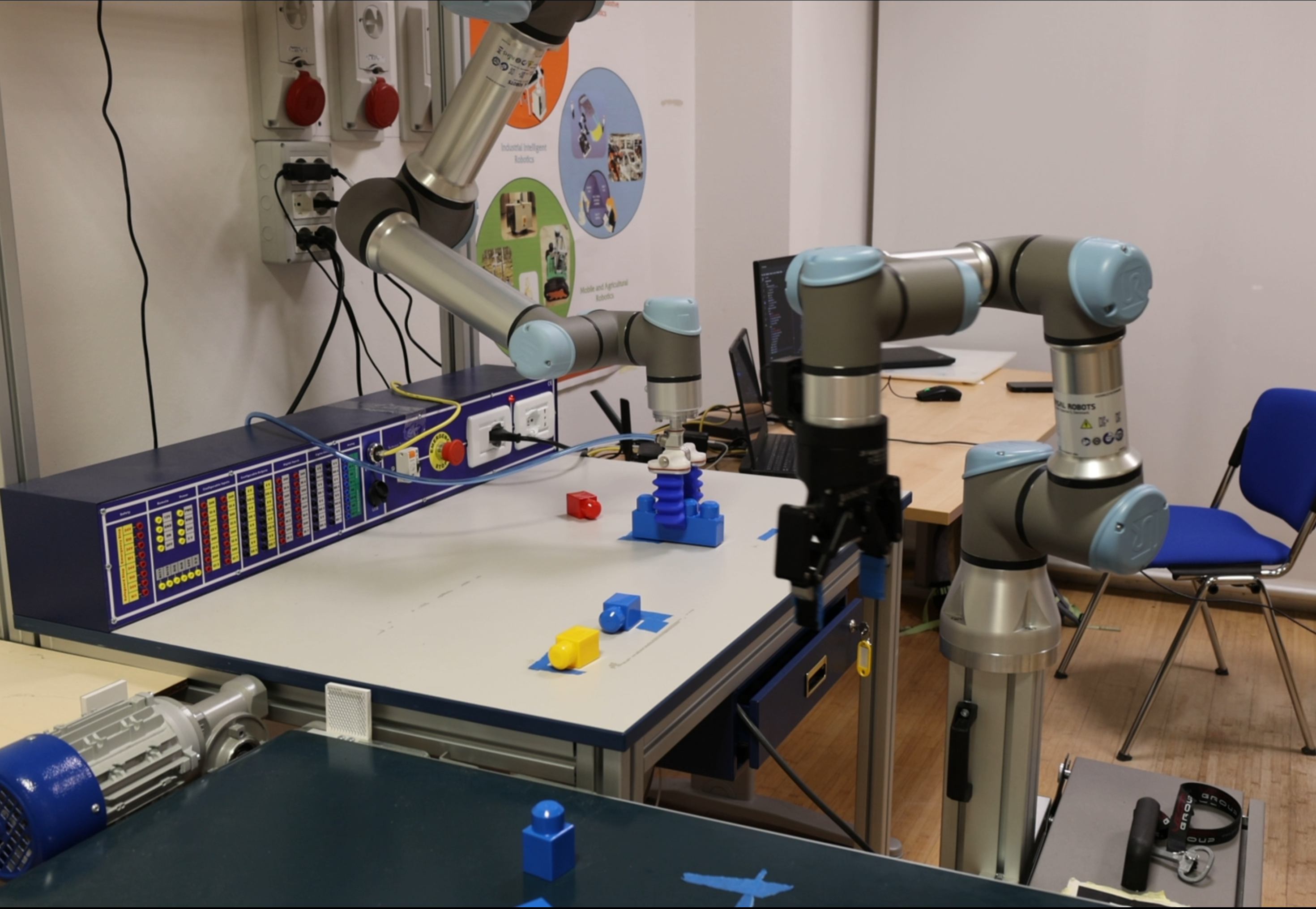}
    % \hfill
    \includegraphics[width=0.48\linewidth]{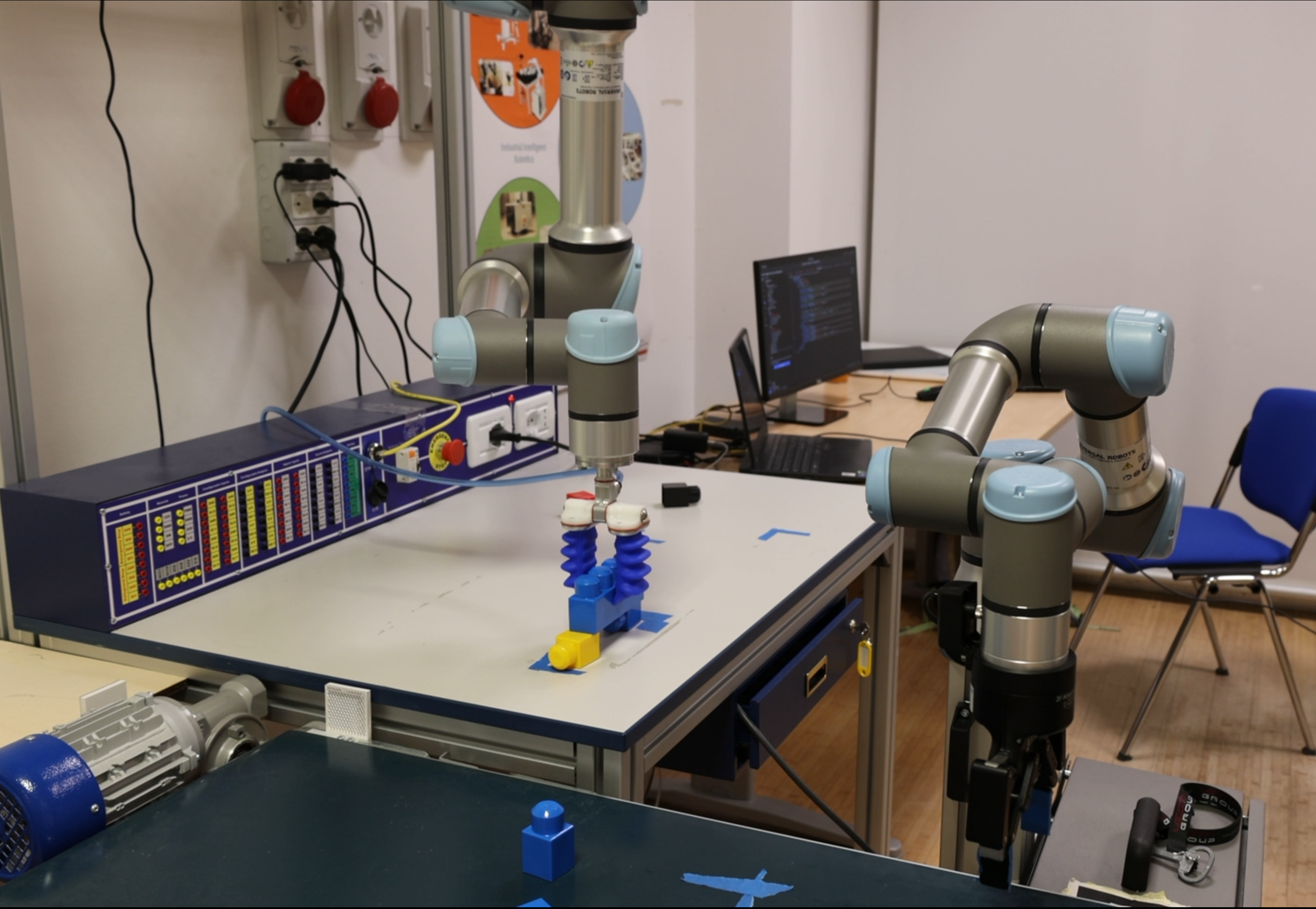}
    \caption{The real-world experiment we carried out using two Universal Robots arms. Images show the initial position of blocks and arms (top-left), the placement of the two blocks as pillars (top-right), the picking of the architrave (bottom-left), and finally the placement of the architrave (bottom-right).}
    \label{fig:rlexp}
\end{figure*}

Finally, we present a real-world experiment run on two robots from Universal Robots, a UR3 and a UR5e, equipped respectively with a 2f85 gripper from Robotiq and an mGrip gripper from Soft Robotics. The scenario consisted in having the two robots cooperate to correctly build an arch made of three blocks: one block per pillar plus one architrave. In Figure~\ref{fig:rlexp}, it is possible to see the setup used.

We positioned the blocks on the table and provided a query similar to the one used for instance 5 of Blocks World, changing only the coordinates of the blocks. The low-level actions corresponded to the robots’ APIs and were executed through ROS2.

We used \frameworkname to extract a \bt, which was parsed by BehaviorTree.ROS2$^{\ref{fn:behaviortreeCPP}}$ to enable direct communication with the ROS2 servers connected to the robots. The robots executed the plan and assembled the arch. This experiment is a qualitative demonstration of end-to-end integration (\textbf{R5}).

To showcase the robustness of the framework, we tested the \bt generation on four variations from the initial prompts used for the physical algorithm. We used Claude Opus 4.6 for the KMS module and report briefly here the statistics. 

All the prompts clearly mentioned the concept of \verb|area|, i.e., either the table or the infeed platform. The main errors made by Opus 4.6 consisted in not explicitly generating a predicate in $K$ for \verb|area|, even though it still used the concept of area correctly when specifying the positions of the blocks by using the predicate \verb|at(Object, Area, X, Y)| in the initial and final states. Interestingly, this error occurred in four out of five tests (the four variations plus the real-world experiment) and in one it did not cause any problems in the generation of the plan since it was correctly used in the preconditions of the action. The model often made another error in the generation of the high-level action \verb|move_block|, confusing the initial area of the block, the target area and the area of the agent. 

Contrary to the experiments shown in Section~\ref{sssec:expKBGeneration}, the mappings contained no independent errors; the apparent mapping errors were inherited from the confusion between area predicates in the high-level actions. Similarly, the errors committed in the generation of the low-level actions were due to a misspelled predicate for the position of the arms, which in 2 instances was generated as \verb|ll_ur5et| instead of \verb|ll_arm_at|. It is worth mentioning that this is not actually an error: the planner would still be able to find a plan, but it is a harmless naming irregularity of the LLM. 

Finding the total order plan took around 380 s on average across the 5 instances, with a minimum of 107 s and a maximum of 1217 s. As previously mentioned, this time variation is to be expected even for prompts that are quite similar, as the LLM is not instructed to generate a \kb that is optimized for planning. The enablers extraction, STN generation and optimization, and the BT conversion took 4 ms, 38 ms and 0.8 ms, respectively, on average. Unlike the benchmark experiments in Section~\ref{ssec:planRes}, these five runs were not subject to the 660 s timeout.

%% file: sections/7-conclusion.tex
\subsection{Discussion and Limitations}
\label{ssec:discussionFutureWork}

The experimental results presented in Section~\ref{sec:experiments} show that the proposed framework can translate natural-language task descriptions into executable Prolog \kbases and generate feasible plans for the considered multi-robot scenarios. In particular, the results suggest that the combination of LLM-based knowledge-base generation, symbolic reasoning, and constraint-based scheduling is a viable direction for producing plans that are both executable and inspectable.

At the same time, the current implementation of the planner should be interpreted as a proof of concept rather than as a fully optimized planning system. The most computationally demanding step is the generation of the high-level total-order plan, as shown in Table~\ref{tab:toPoTimesTable}. This is expected, since the search space grows rapidly with the number of objects, predicates, and possible action instantiations. Our current planner deliberately prioritizes integration with the Prolog \kbase and transparency of the reasoning process over raw performance. More sophisticated planning strategies, heuristics, or compilations to state-of-the-art planners could significantly improve scalability.

A second limitation concerns the evaluation of the two-level \kbase structure. The separation between high-level and low-level knowledge is designed to support reuse across different robotic platforms: the same high-level \kbase can in principle be kept while modifying only the low-level actions and mappings. However, this benefit has not yet been evaluated quantitatively. A dedicated study measuring how much manual effort is saved when transferring the same task description across different robot configurations is left for future work.

The scheduling step also opens opportunities that are not fully exploited in this work. At present, the optimization focuses primarily on minimizing makespan while preserving causal and temporal constraints. However, the same formulation could be extended to encode richer resource-allocation objectives, such as assigning actions to robots based on proximity, availability, energy consumption, expected execution cost, or reliability. This would make the framework more useful in realistic multi-agent deployments, where minimizing time is only one of several relevant objectives.

Future work will therefore focus on improving the planning module in two main directions. First, we plan to investigate the integration of more efficient planning backends, for instance through a compilation from Prolog to PDDL and the use of planners such as OPTIC~\cite{benton_temporal_2012} or FastDownward~\cite{helmert_fast_2006}. This could improve performance, although it may reduce the direct inspectability currently provided by the Prolog-based search. Second, we plan to extend the scheduling optimization with additional resource and allocation constraints, so that the final plan can account for both temporal efficiency and robot-specific execution preferences.

\paragraph{Task and Motion Planning Limitations}

Another important limitation is the current separation between task planning and motion planning. The framework generates symbolic task plans and translates them into executable behavior trees, but it does not explicitly reason about geometric feasibility, collision avoidance, kinematic constraints, grasp feasibility, or path planning. As a consequence, a plan may be valid at the symbolic level but still fail during execution because one of its actions cannot be physically realized by the robot in the current environment.

Addressing this limitation requires a tighter integration with Task and Motion Planning (TAMP)~\cite{garrett_integrated_2021}. In future work, we plan to extend the framework with action models that expose the parameters required for physical execution, such as poses, grasps, trajectories, and motion constraints. One possible direction is to represent low-level actions using parametric Dynamic Movement Primitives~\cite{saveriano_2023} or Kernelized Movement Primitives~\cite{silverio2019uncertainty}. These could be learned from demonstrations and then instantiated during planning, providing a bridge between symbolic action selection and continuous robot motion.

This integration would also help address a second practical issue: the estimation of action durations. In the current implementation, the scheduling problem requires lower and upper duration bounds for each action, which are manually provided by the user. These bounds are intentionally conservative, but this can lead to suboptimal schedules. Learning or estimating action durations from motion-level execution data would make the temporal optimization more accurate and would improve the quality of the generated plans.

\paragraph{Limitations of LLM-Based Knowledge-Base Generation}

The use of LLMs for \kbase generation is effective, but it remains a source of possible errors. Even advanced models can introduce incorrect predicates, inconsistent action descriptions, missing mappings, or unnecessary preconditions. The Prolog-based validation checks mitigate this issue by detecting many structural inconsistencies before planning starts, but they do not guarantee that the generated \kbase fully captures the intended semantics of the task. Human supervision therefore remains necessary, especially when the generated domain is used for real robotic execution.

A further limitation is that LLM-generated \kbases are not always optimized for planning efficiency. In Prolog, the order of predicates in action preconditions can substantially affect the number of possible groundings explored by the planner. For example, checking a state-dependent predicate before a broad static predicate can reduce unnecessary branching. However, an LLM may generate a semantically correct but inefficient ordering. Similarly, it may omit useful constraints, such as requiring the initial and final positions of a movement action to be different, unless this is explicitly stated in the prompt.

Future work should therefore investigate methods for making LLM-generated \kbases more planning-oriented. Possible directions include fine-tuning models on domain-specific examples, using retrieval-based prompting with validated \kbase fragments, or introducing an additional repair stage that optimizes the generated Prolog representation for planner efficiency rather than only for syntactic and semantic consistency.

\paragraph{Real-World Applicability}

Despite these limitations, the proposed framework has significant potential for real-world robotic applications. Its main advantage is that it does not use the LLM as a direct planner or controller. Instead, the LLM is used to generate an intermediate symbolic representation that can be inspected, validated, corrected, and queried before execution. This makes the approach particularly suitable for domains where reliability and traceability are important, such as robotic assembly, warehouse automation, and multi-agent manipulation tasks.

For practical deployment, however, additional robustness mechanisms are needed. In particular, future versions of the framework should support uncertainty handling, probabilistic reasoning~\cite{saccon_automated_2026} (for multiple agents), execution monitoring, and real-time re-planning. These capabilities would allow the system to react to unexpected failures, perception errors, or environmental changes, further reducing the gap between symbolic planning and robust robot execution.

\subsection{Conclusions}
\label{ssec:conclusions}

This paper presented PLANTOR, a framework that combines large language models, logic programming, and temporal planning to generate executable multi-robot plans from natural-language task descriptions. The central idea is to use LLMs not as black-box planners, but as tools for generating structured Prolog \kbases. These \kbases are then validated, queried, and used by a symbolic planning pipeline that produces temporally optimized plans suitable for parallel execution.

The framework separates knowledge generation from plan generation. This separation improves transparency, since the generated \kbase, intermediate plans, causal dependencies, and scheduling constraints can all be inspected by a human user. It also supports reuse, since the high-level task representation can be separated from robot-specific low-level actions and mappings. Finally, the optimized plan is translated into a behavior tree, enabling deployment on robotic platforms through standard execution mechanisms.

The experimental evaluation shows that LLMs can generate useful Prolog \kbases with limited human supervision, and that the proposed consistency checks reduce the number of structural errors before planning. The planning pipeline successfully generates feasible plans and exploits parallelism across multiple robotic agents. These results support the broader claim that LLMs and symbolic methods are complementary: LLMs provide flexibility in interpreting natural-language descriptions, while logic programming and constraint-based optimization provide structure, inspectability, and correctness guarantees.

At the same time, several limitations remain. The current planner is not optimized for large-scale domains, the benefits of \kbase reuse require a more systematic evaluation, and the framework does not yet integrate motion-level feasibility or online re-planning. Addressing these limitations will be essential for deploying the approach in more complex and uncertain robotic environments. Overall, the results indicate that LLM-aided knowledge-base generation, when combined with symbolic planning and temporal optimization, is a promising direction for building robotic systems that are adaptable, explainable, and practically executable.

%% file: sections/8-appendix.tex
\clearpage
\appendix\section{Consistency Checks}\label{sec:app_consistencyChecks}
\input{sections/8-appendix/consistency_checks}

\clearpage
\section{Planner Algorithms}\label{sec:app_plannerAlgorithms}
\input{sections/8-appendix/algorithms}

\clearpage
\section{Scenario Descriptions}\label{sec:app_scenarioDescriptions}
\input{sections/8-appendix/scenarios_descriptions}

\clearpage
\section{Knowledge-Base Generation Experiments}\label{sec:app_kbGenerationExperiments}
\input{sections/8-appendix/kb_generation_experiments}

\clearpage
\section{Planner Experiments}\label{sec:app_plannerExperiments}
\input{sections/8-appendix/planner_experiments}

\clearpage

%% file: sections/8-appendix/consistency_checks.tex
This appendix reports the consistency checks used during the iterative \kbase generation and repair process described in Section~\ref{ssec:KmsConsistencyChecks}. The checks are intended to verify structural properties required by the planner, such as the correct use of predicates, the separation between static and dynamic facts, and the consistency of high-level to low-level mappings. They should not be interpreted as a complete semantic validation of the generated domain, but rather as automatic diagnostics that help identify common modeling errors before planning is attempted.

\begin{table}[h]
    \centering
    \small
    \renewcommand{\arraystretch}{1.35}
    \begin{tabular}{p{0.28\linewidth}p{0.64\linewidth}}
        \toprule
        \textbf{Check} & \textbf{Purpose} \\
        \midrule
        Precondition satisfiability &
        Ensures that each precondition list is internally consistent, e.g., it does not contain both \verb|p| and \verb|neg(p)|. \\

        Effect consistency &
        Ensures that a single start or end effect list does not both add and delete the same fluent. \\

        Durative lifecycle consistency &
        Ensures that start effects do not invalidate end preconditions or overall invariants of the same durative action. \\

        Predicate definition and usage &
        Ensures that predicates used in preconditions or effects belong to the inferred domain signature, and reports unused static predicates. \\

    Goal reachability &
        Ensures that goal predicates can be reached and that deleted 
        predicates can be reintroduced by at least one action effect. \\

    Predicate necessity &
        Ensures that added predicates are required somewhere in the 
        domain and that action preconditions are reachable from the 
        initial state. \\
    
        Static/state separation &
        Ensures that a predicate is not used both as static \kb knowledge and as a fluent in the initial or goal state. \\

        Static/dynamic separation &
        Ensures that action effects do not modify predicates declared as static \kb facts. \\

        Abstraction-layer consistency &
        Ensures that \HL actions do not use \LL predicates and that \LL action effects modify only \LL predicates. \\

        Mapping consistency &
        Ensures that mappings refer to existing \HL and \LL actions, and that \HL duration bounds cover the summed duration bounds of the mapped \LL sequence. \\
        \bottomrule
    \end{tabular}
    \caption{Examples of consistency checks applied to the generated Prolog \kb.}
    \label{tab:kms-consistency-checks}
\end{table}

%% file: sections/8-appendix/algorithms.tex
\subsection{Total-Order Plan Algorithm}\label{ssec:app_toPlanAlgo}

\begin{algorithm}
\footnotesize
\caption{Algorithm generating a TO plan with applied mappings}
\label{alg:toplanning}
\KwData{$MTTP=(F, DA, I, G, K, M)$}
\KwResult{\LL TO plan $\pi_L$}

\DontPrintSemicolon

\SetKwProg{plan}{PLAN}{}{}
\SetKwProg{expand}{APPLY\_MAPPINGS}{}{}
\SetKwProg{map}{APPLY\_ACTION\_MAP}{}{}
\SetKwInOut{Input}{In}
\SetKwInOut{Output}{Out}

\plan{}{
  \Input{}
  \Output{Low-level total-order plan $\pi_L$}

  $\pi_H \gets$ BFS\_PLANNER()\;

  \If{$\pi_H = [~]$}{
    report that no plan was found\;
    \KwRet{fail}\;
  }

  $S \gets I$\;
  $\pi_L \gets$ APPLY\_MAPPINGS($S,~\pi_H$)\;
  \KwRet{$(\pi_L)$}\;
}

\expand{($S, \pi_H$)}{
  \Input{Current state $S$ and high-level plan $\pi_H$}
  \Output{Low-level total-order plan $\pi_L$}

  $\pi_L \gets [~]$\;

  \ForEach{$\tau(a_H) \in \pi_H$}{
    $\pi_L$.append($\tau(a_H)$)\;
    retrieve $a_H=(Pre_S, Pre_E, Inv, Eff_S, Eff_E, Dur)$ from $DA_H$\;

    \uIf{$\tau = \mathit{start}$}{
      $S \gets$ apply $Eff_S$ to $S$\;
      retrieve mapping $M(a_H)=[a_1,\ldots,a_n]$\;
      $(S, \pi_L) \gets$ APPLY\_ACTION\_MAP($S, \pi_L, [a_1,\ldots,a_n]$)\;
    }
    \ElseIf{$\tau = \mathit{end}$}{
      $S \gets$ apply $Eff_E$ to $S$\;
    }
  }

  \KwRet{$\pi_L$}\;
}

\map{($S, TO_{LL}, [a_1,\ldots,a_n]$)}{
  \Input{Current state $S$, current mapped plan $\pi_L$, and mapped \LL actions}
  \Output{Updated state and mapped plan}

  \ForEach{$a_i \in [a_1,\ldots,a_n]$}{
    retrieve $a_i=(Pre_S, Pre_E, Inv, Eff_S, Eff_E, Dur)$ from $DA_L$\;

    \If{!\textnormal{check\_preconditions(}$Pre_S$, $S$\textnormal{)}}{
      fail\;
    }
    $\pi_L$.append($\mathit{start}(a_i)$)\;
    $S \gets$ apply $Eff_S$ to $S$\;

    \If{!\textnormal{check\_preconditions(}$Pre_E$, $S$\textnormal{)}}{
      fail\;
    }
    $\pi_L$.append($\mathit{end}(a_i)$)\;
    $S \gets$ apply $Eff_E$ to $S$\;
  }

  \KwRet{$(S, \pi_L)$}\;
}
\end{algorithm}

This enables the extraction of total-order plans that are consistent with the \kb provided. The \verb|BFS_PLANNER| function performs a breadth-first search in Prolog starting from the initial state $I$ and applying actions in the \kbase until one state satisfies the requirements of the goal state $G$. Since the search is breadth-first, the first plan found minimizes the number of high-level actions. Whenever a dead-end is reached, Prolog's backtracking explores alternative choices.

\subsection{Enablers Extraction Algorithm}\label{ssec:app_enablerExtractionAlgo}

\begin{algorithm}[htp]
\footnotesize
\caption{Algorithm extracting the actions enablers and the resources}
\label{alg:poplanning}
\KwData{$TP=(F, DA, I, G, K)$}
\KwResult{Enablers and resources $R$}

\DontPrintSemicolon

\SetKwProg{findenablers}{FIND\_ENABLERS}{}{}
\SetKwProg{isenabler}{IS\_ENABLER}{}{}

\SetKwInOut{Input} {In}
\SetKwInOut{Output}{Out}

\findenablers{$(\pi_L, a_i)$}{
  \Input{The total-order plan $\pi_L$, the $i$th action}
  \Output{The enablers $E$ for all the actions in the plan}

  \For{$a_j \in \pi_L, a_j\neq a_i$}{
    \uIf{$\tn{IS\_ENABLER}(a_j, a_i)$}{
      $E[a_i].add(a_j)$;
    }
  }

  \If{$a_i\neq \pi_L\tn{.back()}$}{
    $E \gets \tn{FIND\_ENABLERS}(\pi_L, a_{i+1})$\;
  }
  \KwRet{E}\;
}

\isenabler{$(a_j, a_i)$}{
  \Input{The action $a_j$ to test if it's enabler of $a_i$}
  \Output{True if $a_j$ is enabler of $a_i$}

  \ForEach{$e \in \eff{a_j}$}{
    \uIf{$\left(e=\tn{add}(l) \wedge l\in\pc{a_i})\right)$ OR \\
         $~~~\left(e=\tn{del}(l) \wedge \lnot l\in\pc{a_i}\right)$ OR \\
         ~~\tcp*[l]{Hierarchical precedence constraints}
         $~~~\left(\tn{isStart}(a_j) \wedge a_i \in m(a_j)\right)$ OR \\
         $~~~\left(\tn{isEnd}(a_i) \wedge a_j \in m(a_j)\right)$}
    {
        \KwRet{True};
    }
  }
  \KwRet{False};
}
\end{algorithm}

The function \verb|FIND_ENABLERS| takes the total-order plan and, starting with the first action in the plan, it extracts all the causal relationships between the actions. The auxiliary function \verb|IS_ENABLER| tests whether an action $a_j$ is an enabler of an action $a_i$ by checking the properties of~\autoref{eq:enablers} plus the precedence constraints  described in Equations~\ref{eq:constraint5}, \ref{eq:constraint4}. For each action $a_i$, Algorithm~\ref{alg:poplanning} will finally return a list $\ach{a_i}$ containing the IDs of the actions that are enablers of $a_i$ or have a hierarchical precedence.

% \subsection{Behavior-Tree Generation Algorithm}

%% file: sections/8-appendix/scenarios_descriptions.tex
This appendix provides the detailed scenario descriptions used in the experimental evaluation. The tables summarize the initial configuration, the goal configuration, and the main feature tested by each instance. These descriptions complement Section~\ref{ssec:scenariosDescription}, where the two benchmark-inspired domains are introduced at a higher level.

\subsection{Scenarios used for the Blocks World Use-Case}\label{ssec:app_BWScenarios}

\begin{table*}[h]
\centering
\small
\caption{Summary of the Blocks World benchmark scenarios.}
\label{tab:blocks-world-scenarios}
\begin{tabular}{@{}p{0.1\textwidth}p{0.28\textwidth}p{0.28\textwidth}p{0.23\textwidth}@{}}
\toprule
\textbf{Instance} & \textbf{Initial configuration} & \textbf{Goal configuration} & \textbf{Main feature} \\
\midrule
1 &
Three blocks are on the table at positions $(1,1)$, $(2,2)$, and $(3,3)$; two agents are available. &
Blocks $b1$ and $b2$ remain on the table, while $b3$ is placed on top of $b1$ at position $(1,1)$. &
Simple stacking task with two agents. \\

2 &
Six blocks are arranged as three stacks: $b4$ on $b1$, $b5$ on $b2$, and $b6$ on $b3$. &
Only $b4$ and $b5$ must be moved: $b4$ is placed on the table at $(10,10)$ and $b5$ is placed on top of $b4$. &
Partial-goal rearrangement where not all blocks are relevant to the goal. \\

3 &
Four blocks form a single tower at $(1,1)$, with $A$ at the bottom, followed by $B$, $C$, and $D$ on top. &
$A$ remains at $(1,1)$, $B$ and $C$ are moved to table positions $(2,1)$ and $(3,1)$, and $D$ is placed on top of $A$. &
Tower unstacking and partial reconstruction. \\

4 &
Three blocks are initially on the table at positions $(1,1)$, $(2,3)$, and $(3,4)$; one agent is available. &
The first block is moved to $(5,5)$, the second block is stacked on it, and the third block is stacked on the second. &
Single-agent stacking problem. \\

5 &
Three blocks are on the table; two blocks are support blocks and one longer block is an architrave. &
The support blocks are moved to target positions and the architrave is placed across them. &
Arch-construction scenario requiring a special placement action over two supports. \\

6 &
Two blocks are initially inside two bowls, which remain fixed on the table. &
The blocks are removed from the bowls and stacked at $(6,6)$, with $b1$ on the table and $b2$ on top of $b1$. &
Container-manipulation scenario with bowl-removal actions. \\
\bottomrule
\end{tabular}
\end{table*}

\subsection{Scenarios used for the Grippers Use-Case}\label{ssec:app_GScenarios}

\begin{table*}[h]
\centering
\small
\caption{Summary of the Grippers benchmark scenarios.}
\label{tab:grippers-scenarios}
\begin{tabular}{@{}p{0.1\textwidth}p{0.31\textwidth}p{0.25\textwidth}p{0.23\textwidth}@{}}
\toprule
\textbf{Instance} & \textbf{Initial configuration} & \textbf{Goal configuration} & \textbf{Main feature} \\
\midrule
1 &
One robot with two grippers starts in Room~A with three balls; all grippers are empty. &
All three balls must be transported to Room~B. &
Classical single-robot Grippers problem with two-ball carrying capacity. \\

2 &
Two robots, each with one gripper, start in Room~A with four balls. &
All four balls must be transported to Room~B. &
Two-robot transport with a wide door allowing simultaneous passage. \\

3 &
Two robots start in Room~A; three balls start in Room~A and one ball starts in Room~B. &
All balls must end in Room~B, but the ball initially in Room~B must first be returned to Room~A and then delivered again. &
Mandatory intermediate delivery represented by an additional bookkeeping condition. \\

4 &
Two robots and four balls start in Room~A; the rooms are connected by a narrow door. &
All four balls and both robots must end in Room~B. &
Narrow-door resource constraint: only one robot may pass through the door at a time. \\

5 &
Two robots and three balls start in Room~A; one ball is heavy. &
All three balls must be transported to Room~B. &
Cooperative heavy-ball transport requiring both robots and both grippers. \\

6 &
Four balls and robot~1 start in Room~A, while robot~2 starts in Room~B. &
All four balls must be transported to Room~B. &
Doorway handoff scenario, where robots can pass balls across the narrow doorway. \\
\bottomrule
\end{tabular}
\end{table*}

%% file: sections/8-appendix/kb_generation_experiments.tex
This appendix reports additional details on the knowledge-bases generated in the experiments of Section~\ref{ssec:KBGenValidation} and on the KMS process. \ref{ssec:app_examplesFewShotPrompting} shows three examples passed to the LLMs to guide them in the generation of the \kb. \ref{ssec:app_domainValidationInsights},~\ref{ssec:app_hlKBGenerationInsights} and~\ref{ssec:app_llKBGenerationInsights} report more details on the experiments evaluation for the high-level and low-level parts of the \kb generation. \ref{ssec:app_tabPredicatesKB} summarizes the size and structure of the generated Prolog representations, while~\ref{ssec:app_repairLoopResults} reports the number of automatic repair iterations required by the consistency-check loop. Together, these results provide a more detailed view of the modeling effort handled by the LLM and of the residual human supervision required after automatic validation.

\subsection{Examples for Few-Shot Prompting}\label{ssec:app_examplesFewShotPrompting}

Figure~\ref{fig:llmFewShotExamples} shows a set of examples passed to the LLM to guide it in the generation of the \kb. Figure~\ref{fig:genExample} shows a general message specifying the components of the high-level \kb and giving general pieces of information on it. The second example (Figure~\ref{fig:actionExample}) specifies how high-level durative actions should be modeled. Finally, Figure~\ref{fig:blocksExample} shows an examples based on Blocks World on how the initial and final states should be generated. 

\begin{figure*}[htp]
    \centering
    \begin{subfigure}{0.99\linewidth}
        \centering
        \begin{textbox}{}
\scriptsize
Q:
    role: ``user''
    content: $\vert$
      How is your code divided?

A:
    role: ``assistant''
    content: $\vert$
      The high-level knowledge-base is composed of 4 parts:
      - \`{}init\`{} section, which contains the initial state of the environment;
      - \`{}goal\`{} section, which contains the final state of the environment;
      - \`{}\kb\`{} section, which contains all information that is not modified by actions;
      - \`{}actions\`{} section, which contains the durative actions available to the agents.

      For example, \`{}available(A)\`{} is modified by actions and should therefore be placed
      in the \`{}init\`{} section. On the other hand, \`{}pos(X,Y)\`{} defines a valid position
      and is not modified by actions, so it belongs to the \`{}\kb\`{} section.
        \end{textbox}
        \vspace{-0.275cm}
        \caption{General example passed to the LLM to explain how the \kb should be structured.}
        \label{fig:genExample}
    \end{subfigure}
    \vspace{0.3cm}

    \begin{subfigure}{0.99\linewidth}
        \centering
        \begin{textbox}{}
\scriptsize
Q:
    role: ``user''
    content: $\vert$
      How do you describe the structure of an action?

A:
    role: ``assistant''
    content: $\vert$ In the knowledge base, actions are modeled as durative actions:
\begin{minted}[fontsize=\scriptsize]{prolog}
hl_d_action(
  Name(Argument1, Argument2, ...),
  PreconditionsAtStart,
  PreconditionsAtEnd,
  PreconditionsOverall,
  EffectsAtStart,
  EffectsAtEnd,
  Duration
).
\end{minted}
      Effects are written using \`{}add(P(...))\`{} and \`{}del(P(...))\`{}.
      Negative preconditions are expressed using \`{}neg(P(...))\`{}.
      The duration is represented as an interval \`{}[Min, Max]\`{}.
        \end{textbox}
        \vspace{-0.275cm}
        \caption{General example passed to the LLM to explain the durative action format.}
        \label{fig:actionExample}
    \end{subfigure}
    \vspace{0.3cm}

    \begin{subfigure}{0.99\linewidth}
        \centering
        \begin{textbox}{}
\footnotesize
Q:
    role: ``user''
    content: $\vert$
      Can you generate a Prolog snippet of code that describes the initial and final states for
      the following scenario?
      Given an initial state in which there are two blocks b1, b2 in position (1,1) and (2,2)
      respectively, move the block b1 to position (3,3) and place b2 on top of b1 using an
      agent, which is initially available and it will also be available at the end.

A:
    role : ``assistant''
    content: $\vert$
      Here is the initial state:
\begin{minted}[fontsize=\scriptsize]{prolog}
init_state([
  ontable(b1), ontable(b2), 
  at(b1,1,1), at(b2,2,2), 
  clear(b1), clear(b2), 
  available(a1)
]).
\end{minted}
      Here is the goal state:
\begin{minted}[fontsize=\scriptsize]{prolog}
goal_state([
  ontable(b1), on(b2,b1), 
  at(b1,3,3), at(b2,3,3), 
  clear(b2), 
  available(a1)
]).
\end{minted}
        \end{textbox}
        \vspace{-0.275cm}
        \caption{Scenario-specific example passed to the LLM for state generation.}
        \label{fig:blocksExample}
    \end{subfigure}
    \caption{Examples fed to the LLM through few-shot prompting.}
    \label{fig:llmFewShotExamples}
\end{figure*}

\subsection{Domain Validation Insights}\label{ssec:app_domainValidationInsights}

In the domain validation experiments (Section~\ref{sssec:expDomainValidation}), GPT 5.2 and Claude Opus 4.6 committed no mistakes. 

Sonnet 4.6 makes two mistake, on Grippers instance 2 and 5. In both cases, it incorrectly reports a mismatch between the \HL and \LL action durations. This is not a real inconsistency: the low-level durations refine the high-level actions without contradicting their semantics. After this manual check, the \kb generation can proceed, and the \LL result is generated correctly for instance 2, as shown in Table~\ref{tab:llRes}, while it contains some different logical errors for instance 5.

GPT 5.4 Mini makes three mistakes. In Blocks World instance 2.a, it incorrectly reports that the \LL prompt does not specify how to pick, carry, and place blocks, missing that the provided primitives are meant to refine the \HL move actions. In Blocks World instance 5.b, it fails to detect that a unit-sized block cannot serve as the architrave between the two support blocks. This suggests that the model reasons primarily at the symbolic-planning level and gives insufficient weight to the physical dimensions in the prompt. Finally, in Grippers instance 5, it incorrectly treats the low-level home-position requirement as an additional inconsistency, although this constraint is part of the intended implementation of the robot actions.

Among the open-weight models, Llama 3.3 and Qwen 3.6 perform best, with one and two mistakes, respectively. Llama 3.3 incorrectly rejects the valid Blocks World instance 5.a, arguing that the support blocks are insufficient to support the architrave and that its dimensions are underspecified. Qwen 3.6 instead fails to detect the inconsistency in Blocks World instance 2.b, and, in Grippers instance 5, complains about a duration mismatch between the high-level to move both robots together and the low-level actions.

\subsection{High-Level \kb Generation Insights}\label{ssec:app_hlKBGenerationInsights}

Blocks World instance 3 is particularly informative: all models made the same kind of action-level mistake in \verb|move_onblock_to_block|, i.e., an unnecessary overall condition on the block position. This produced a \kbase that looked structurally plausible and led to a formally correct plan, but not the optimal one. Indeed, the optimal solution exploits the duration of the previously mentioned action to reduce the total number of needed actions from 4 to 3. Similarly, in Grippers 3 and 5, Llama 3.3 missed domain-specific action structure: in Grippers 3 the model did not create a special action to return ball 4 merging it with the normal delivery action, while in Grippers 5 it omitted the explicit movement action for a jointly held heavy ball.

\subsection{Low-Level \kb Generation Insights}\label{ssec:app_llKBGenerationInsights}

The Blocks World mappings reveal a recurring but simple error. The intended \LL decomposition for an action moving a block is: move the arm to the source position, close the gripper, move the arm to the target position, and open the gripper. GPT 5.4 Mini often generated mappings in which the first arm movement started from a fixed home coordinate, e.g., \verb|ll_move_arm(Agent,4,4,X1,Y1)|. The correction consists in replacing this with anonymous variables, e.g., \verb|ll_move_arm(Agent,_,_,X1,Y1)|. Llama 3.3 also makes occasional sequencing mistakes in the mapping, for example opening before closing or omitting the final return/home-related step. Pointing out these errors is important because they do not make the Prolog file syntactically invalid and also often satisfy the consistency checks, hence making them difficult to be spotted.

In the simplest Grippers variants, the correct mapping for a pick action must move the arm to the ball, close the corresponding gripper, and return the arm to home; the drop action must lower the arm, open the gripper, and raise the arm back. Several original files omit the return-to-home step, use an unbound gripper or arm variable, or fail to connect the selected \HL gripper with the corresponding \LL arm. In the one-robot/two-gripper case, for instance, the corrected files add constraints such as a relation between arms and grippers so that the mapping does not choose an arbitrary arm for a gripper. These errors explain why even when $DA_L$ is otherwise plausible, the generated \kb can still be wrong: the primitive actions exist, but the \HL action is refined into the wrong primitive sequence.

The harder Grippers instances highlight the same problem. Instance 4 adds a constraint for which only one agent at a time can move through the narrow door. Several models incorrectly encode mutual exclusion as a room-occupancy constraint, preventing two robots from being in the same room rather than only preventing simultaneous traversal of the door. Instance 6 adds handoff actions at the doorway, which is the most demanding characteristic: the \LL model must represent waiting poses, handoff arm poses, source and receiver gripper states, and the transfer of possession without necessarily moving both robots through the door. In general, the errors are due to incorrect transition semantics: the models often generate plausible low-level actions but fail to coordinate the details, such as gripper occupancy, within the same \LL actions.

\subsection{Table Predicates Knowledge-Bases}\label{ssec:app_tabPredicatesKB}

\begin{table*}[h]
    \centering
    \footnotesize
    \setlength{\tabcolsep}{2pt}
    \renewcommand{\arraystretch}{1.0}

    \caption{Summary of \kb generation outcomes and final corrected \kb sizes, grouped by domain, abstraction level, and task instance. Aggregate rows report totals over all runs, while numbered rows report the corresponding per-experiment values. The table includes the number of runs, completely correct outputs, required logical changes, and the average size of the corrected \kb in terms of non-commented Prolog lines, facts, and action or mapping schemas.}
    \label{tab:predNumber}
    
    \begin{tabular}{c c r r r r r r}
        \toprule
        \multicolumn{2}{c}{Instance} & Runs & Correct & Changes & Lines & Facts & Schemas \\
        \midrule
        \multirow{6}{*}{Blocks World}  &   HL  & 36 & 25 & 20 & 73.4  & 33.8 & 4.6 \\
                      &    1  &  6 &  6 &  0 & 58.8  & 29.8 & 4.0 \\
                      &    2  &  6 &  5 &  3 & 76.0  & 47.7 & 4.0 \\
                      &    3  &  6 &  0 & 10 & 63.5  & 34.0 & 4.0 \\
                      &    4  &  6 &  6 &  0 & 58.2  & 27.0 & 4.0 \\
                      &    5  &  6 &  3 &  6 & 74.5  & 33.7 & 5.0 \\
                      &    6  &  6 &  5 &  1 & 109.2 & 30.5 & 6.5 \\
        \midrule
        \multirow{6}{*}{Blocks World}  &   LL  & 36 & 14 & 29 & 157.5 & 47.9 & 12.0 \\
                      &    1  &  6 &  3 &  3 & 143.7 & 45.0 & 11.0 \\
                      &    2  &  6 &  3 &  5 & 161.8 & 63.2 & 10.7 \\
                      &    3  &  6 &  2 &  6 & 135.0 & 48.7 & 10.3 \\
                      &    4  &  6 &  2 &  4 & 131.0 & 34.7 & 11.0 \\
                      &    5  &  6 &  3 &  4 & 165.5 & 49.7 & 13.0 \\
                      &    6  &  6 &  4 &  7 & 208.5 & 46.0 & 16.0 \\
        \midrule
        \multirow{6}{*}{Grippers}      &   HL  & 36 & 18 & 58 & 81.5  & 31.7 & 4.0 \\
                      &    1  &  6 &  5 &  1 & 71.3  & 25.7 & 3.0 \\
                      &    2  &  6 &  4 &  8 & 55.3  & 31.7 & 3.0 \\
                      &    3  &  6 &  3 &  7 & 77.5  & 32.0 & 4.3 \\
                      &    4  &  6 &  2 &  9 & 68.8  & 34.0 & 3.0 \\
                      &    5  &  6 &  1 & 17 & 117.0 & 32.7 & 6.0 \\
                      &    6  &  6 &  3 & 16 & 98.8  & 34.0 & 4.8 \\
        \midrule
        \multirow{6}{*}{Grippers}      &   LL  & 36 & 13 & 129 & 257.1 & 60.2 & 16.9 \\
                      &    1  &  6 &  1 &  17 & 214.7 & 48.3 & 13.3 \\
                      &    2  &  6 &  3 &  16 & 203.2 & 59.0 & 13.0 \\
                      &    3  &  6 &  2 &  36 & 229.7 & 59.7 & 16.3 \\
                      &    4  &  6 &  3 &  23 & 230.2 & 64.0 & 13.3 \\
                      &    5  &  6 &  3 &  25 & 342.5 & 56.8 & 23.0 \\
                      &    6  &  6 &  3 &  12 & 322.5 & 73.2 & 22.2 \\
        \midrule
        All           &   HL  & 72 & 43 & 78  & 77.4  & 32.7 & 4.3 \\
        All           &   LL  & 72 & 27 & 158 & 207.3 & 54.0 & 14.4 \\
        \bottomrule
    \end{tabular}
\end{table*}

\subsection{Repair Loop Results}\label{ssec:app_repairLoopResults}
\begin{table*}[h]
    \centering
    \caption{Repair-loop effort during \kb generation. Each entry reports the number of automatic consistency-repair iterations performed for the high-level and low-level \kbs, respectively, as (\HL, \LL), out of 3 maximum iterations. Underlined values mark cases where the corresponding generated \kb still required manual correction after the repair loop.}
    \label{tab:repairIterations}
    \setlength{\tabcolsep}{3pt}
    \renewcommand{\arraystretch}{1.05}
    \footnotesize
    \begin{tabular}{cccccccc}
        \toprule
        \multicolumn{2}{c}{Experiment} & Opus 4.6 & Sonnet 4.6 & GPT 5.2 & GPT 5.4 Mini & Qwen 3.6 & Llama 3.3 \\
        \midrule
        \multirow{6}{*}{\rotatebox[origin=c]{90}{Blocks World}}
        & 1 & (0,1) & (0,0) & (0,1) & (0,\underline{0}) & (0,\underline{0}) & (0,\underline{1}) \\
        & 2 & (0,1) & (0,2) & (0,1) & (0,\underline{1}) & (0,\underline{1}) & (\underline{0},\underline{1}) \\
        & 3 & (\underline{0},\underline{1}) & (\underline{0},\underline{1}) & (\underline{1},1) & (\underline{0},1) & (\underline{0},\underline{2}) & (\underline{0},\underline{1}) \\
        & 4 & (0,1) & (0,2) & (0,\underline{1}) & (0,\underline{2}) & (0,\underline{2}) & (0,\underline{1}) \\
        & 5 & (0,0) & (\underline{0},\underline{2}) & (1,1) & (0,0) & (\underline{0},\underline{1}) & (\underline{3},\underline{3}) \\
        & 6 & (1,1) & (1,1) & (1,1) & (3,0) & (\underline{1},\underline{0}) & (3,\underline{1}) \\
        \midrule
        \multirow{6}{*}{\rotatebox[origin=c]{90}{Grippers}}
        & 1 & (0,\underline{1}) & (0,\underline{2}) & (0,2) & (\underline{3},\underline{3}) & (1,\underline{3}) & (3,\underline{3}) \\
        & 2 & (0,1) & (\underline{0},1) & (0,1) & (\underline{3},\underline{3}) & (1,\underline{3}) & (3,\underline{3}) \\
        & 3 & (0,\underline{3}) & (1,3) & (0,3) & (\underline{1},\underline{3}) & (0,\underline{3}) & (\underline{3},\underline{3}) \\
        & 4 & (1,3) & (0,3) & (\underline{1},1) & (\underline{3},\underline{3}) & (\underline{1},\underline{3}) & (\underline{3},\underline{3}) \\
        & 5 & (0,3) & (\underline{1},3) & (\underline{1},3) & (\underline{0},\underline{3}) & (\underline{1},\underline{3}) & (\underline{3},\underline{3}) \\
        & 6 & (1,3) & (\underline{1},3) & (1,1) & (\underline{3},\underline{3}) & (\underline{1},\underline{3}) & (3,\underline{3}) \\
        \bottomrule
    \end{tabular}
\end{table*}

%% file: sections/8-appendix/planner_experiments.tex
\subsection{High-Level and Low-Level Total-Order Planner Performance}\label{ssec:app_toTimes}

\begin{table*}[h]
    \caption{Total-order planning and low-level ordering times for generated low-level \kbases, reported as mean and standard deviation in seconds over successful runs. An \texttt{X} is used when the planner could not finish within the set timeout.}
    \label{tab:toPoTimesTable}
    \tiny
    \centering
\input{figures/experiments/timings_to_po_table}
\end{table*}
Table~\ref{tab:toPoTimesTable} reports the detailed timing results for total-order plan generation. For each generated \kb, the table includes the average runtime and standard deviation over repeated executions. Entries marked with \texttt{X} indicate cases that were not solved within the timeout criterion described in Section~\ref{ssec:planRes}.

\subsection{Enablers Extraction Performance}\label{ssec:app_enalersTimes}

\begin{figure*}[h]
    \centering
    \begin{subfigure}{\textwidth}
        \centering
        \input{figures/experiments/timings_enablers_bw}
        \caption{Times for enablers extraction for the Blocks World use-case.}
        \label{fig:timesEnablersBW}
    \end{subfigure}
    \vspace{0.1cm}\\
    \begin{subfigure}{\textwidth}
        \centering
        \input{figures/experiments/timings_enablers_g}
        \caption{Times for enablers extraction for the Grippers use-case.}
        \label{fig:timesEnablersG}
    \end{subfigure}
    \caption{Times for enablers extraction.}
    \label{fig:timesEnablers}
\end{figure*}

Figure~\ref{fig:timesEnablers} reports the time required to extract the enablers from the low-level total-order plan, and it shows that this step is negligible when compared with total-order high-level plan search. The enabler extraction operates on the actions already present in the low-level plan and checks causal dependencies between their preconditions and effects. Therefore, its cost depends mainly on the length of the generated plan and on the number of predicates appearing in each action, rather than on the full symbolic search space.

\subsection{STN Generation and Optimization Performance}\label{ssec:app_STNTimes}

\begin{figure*}[h]
    \centering
    \begin{subfigure}{\textwidth}
        \centering
        \input{figures/experiments/timings_stn_bw}
        \caption{Times for STN generation and optimization for the Blocks World use-case.}
        \label{fig:timesSTNGenOptBW}
    \end{subfigure}
    \vspace{0.1cm}\\
    \begin{subfigure}{\textwidth}
        \centering
        \input{figures/experiments/timings_stn_g}
        \caption{Times for STN generation and optimization for the Grippers use-case.}
        \label{fig:timesSTNGenOptG}
    \end{subfigure}
    \caption{Average times for STN generation and optimization.}
    \label{fig:timesSTNGenOpt}
\end{figure*}

Figure~\ref{fig:timesSTNGenOpt} reports the time required to construct and optimize the STN and, also in this case, the computational cost is small compared with total-order high-level planning. STN construction consists of adding time-point nodes for the snap actions and temporal constraints for the causal dependencies and duration bounds. The optimization step then assigns timestamps while minimizing the ending time. The results indicate that, for the considered domains, temporal optimization is not the limiting factor of the framework.

%% file: figures/experiments/timings_to_po_table.tex
\renewcommand{\arraystretch}{0.85}
\begin{tabular}{ccccccrr}
\toprule
\multicolumn{2}{c}{Instance} & Model & Runs & OK & Timeout & Total order & Low-level order \\
\midrule
\multirow{30}{*}{\rotatebox[origin=c]{90}{Blocks World}}
& Opus 4.6 & 10 & 10 & 0 & $0.2996 \pm 0.0528$ & $0.0001 \pm 0.0000$ \\
& & Sonnet 4.6 & 10 & 10 & 0 & $0.1956 \pm 0.0021$ & $0.0001 \pm 0.0000$ \\
& & GPT 5.2 & 10 & 10 & 0 & $0.0504 \pm 0.0031$ & $0.0001 \pm 0.0000$ \\
& & GPT 5.4 Mini & 10 & 10 & 0 & $0.1983 \pm 0.0046$ & $0.0001 \pm 0.0000$ \\
& & Qwen 3.6 & 10 & 10 & 0 & $0.2813 \pm 0.0049$ & $0.0001 \pm 0.0000$ \\
& & Llama 3.3 & 10 & 10 & 0 & $0.1979 \pm 0.0069$ & $0.0001 \pm 0.0000$ \\
\cmidrule[0.01pt](r){2-8}
& \multirow{6}{*}{2}
& Opus 4.6 & 10 & 10 & 0 & $533.5196 \pm 15.0665$ & $0.0002 \pm 0.0000$ \\
& & Sonnet 4.6 & 10 & 10 & 0 & $342.0620 \pm 7.4417$ & $0.0001 \pm 0.0000$ \\
& & GPT 5.2 & 10 & 10 & 0 & $545.5565 \pm 14.8390$ & $0.0002 \pm 0.0000$ \\
& & GPT 5.4 Mini & 2 & 0 & 2 & \texttt{X} & \texttt{X} \\
& & Qwen 3.6 & 10 & 10 & 0 & $25.2076 \pm 0.7908$ & $0.0002 \pm 0.0000$ \\
& & Llama 3.3 & 10 & 10 & 0 & $50.5961 \pm 1.1608$ & $0.0001 \pm 0.0000$ \\
\cmidrule[0.01pt](r){2-8}
& \multirow{6}{*}{3}
& Opus 4.6 & 10 & 10 & 0 & $11.4265 \pm 0.2887$ & $0.0002 \pm 0.0000$ \\
& & Sonnet 4.6 & 10 & 10 & 0 & $11.4505 \pm 0.2985$ & $0.0002 \pm 0.0000$ \\
& & GPT 5.2 & 10 & 10 & 0 & $1.8707 \pm 0.0102$ & $0.0001 \pm 0.0000$ \\
& & GPT 5.4 Mini & 10 & 10 & 0 & $306.6168 \pm 5.0927$ & $0.0001 \pm 0.0000$ \\
& & Qwen 3.6 & 10 & 10 & 0 & $404.3395 \pm 5.8558$ & $0.0002 \pm 0.0000$ \\
& & Llama 3.3 & 10 & 10 & 0 & $192.8558 \pm 3.4856$ & $0.0002 \pm 0.0000$ \\
\cmidrule[0.01pt](r){2-8}
& \multirow{6}{*}{4} 
& Opus 4.6 & 10 & 10 & 0 & $1.0553 \pm 0.0360$ & $0.0002 \pm 0.0000$ \\
& & Sonnet 4.6 & 10 & 10 & 0 & $1.0440 \pm 0.0246$ & $0.0001 \pm 0.0000$ \\
& & GPT 5.2 & 10 & 10 & 0 & $1.0438 \pm 0.0251$ & $0.0001 \pm 0.0000$ \\
& & GPT 5.4 Mini & 10 & 10 & 0 & $0.8913 \pm 0.0885$ & $0.0001 \pm 0.0000$ \\
& & Qwen 3.6 & 10 & 10 & 0 & $1.0459 \pm 0.0269$ & $0.0002 \pm 0.0000$ \\
& & Llama 3.3 & 10 & 10 & 0 & $0.8672 \pm 0.0181$ & $0.0002 \pm 0.0000$ \\
\cmidrule[0.01pt](r){2-8}
& \multirow{6}{*}{5}
& Opus 4.6 & 2 & 0 & 2 & \texttt{X} & \texttt{X} \\
& & Sonnet 4.6 & 2 & 0 & 2 & \texttt{X} & \texttt{X} \\
& & GPT 5.2 & 2 & 0 & 2 & \texttt{X} & \texttt{X} \\
& & GPT 5.4 Mini & 2 & 0 & 2 & \texttt{X} & \texttt{X} \\
& & Qwen 3.6 & 2 & 0 & 2 & \texttt{X} & \texttt{X} \\
& & Llama 3.3 & 2 & 0 & 2 & \texttt{X} & \texttt{X} \\
\cmidrule[0.01pt](r){2-8}
& \multirow{6}{*}{6}
& Opus 4.6 & 10 & 10 & 0 & $0.4944 \pm 0.0069$ & $0.0001 \pm 0.0000$ \\
& & Sonnet 4.6 & 10 & 10 & 0 & $0.2209 \pm 0.0252$ & $0.0001 \pm 0.0000$ \\
& & GPT 5.2 & 10 & 10 & 0 & $0.3981 \pm 0.0175$ & $0.0001 \pm 0.0000$ \\
& & GPT 5.4 Mini & 10 & 10 & 0 & $0.0428 \pm 0.0004$ & $0.0001 \pm 0.0000$ \\
& & Qwen 3.6 & 10 & 10 & 0 & $0.4165 \pm 0.0133$ & $0.0001 \pm 0.0000$ \\
& & Llama 3.3 & 10 & 10 & 0 & $0.3129 \pm 0.0098$ & $0.0001 \pm 0.0000$ \\
\midrule
\multirow{36}{*}{\rotatebox[origin=c]{90}{Grippers}}
& \multirow{6}{*}{1} 
& Opus 4.6 & 10 & 10 & 0 & $0.1561 \pm 0.0026$ & $0.0003 \pm 0.0000$ \\
& & Sonnet 4.6 & 10 & 10 & 0 & $0.1663 \pm 0.0013$ & $0.0002 \pm 0.0000$ \\
& & GPT 5.2 & 10 & 10 & 0 & $0.1838 \pm 0.0012$ & $0.0003 \pm 0.0000$ \\
& & GPT 5.4 Mini & 10 & 10 & 0 & $2.4536 \pm 0.0897$ & $0.0002 \pm 0.0000$ \\
& & Qwen 3.6 & 10 & 10 & 0 & $0.1434 \pm 0.0010$ & $0.0002 \pm 0.0000$ \\
& & Llama 3.3 & 10 & 10 & 0 & $0.2467 \pm 0.0122$ & $0.0002 \pm 0.0000$ \\
\cmidrule[0.01pt](r){2-8}
& \multirow{6}{*}{2} 
& Opus 4.6 & 10 & 10 & 0 & $35.2801 \pm 1.6154$ & $0.0005 \pm 0.0000$ \\
& & Sonnet 4.6 & 10 & 10 & 0 & $37.9080 \pm 0.4970$ & $0.0004 \pm 0.0000$ \\
& & GPT 5.2 & 10 & 10 & 0 & $35.2407 \pm 1.6942$ & $0.0004 \pm 0.0000$ \\
& & GPT 5.4 Mini & 10 & 10 & 0 & $37.3384 \pm 1.6221$ & $0.0004 \pm 0.0000$ \\
& & Qwen 3.6 & 10 & 10 & 0 & $26.4839 \pm 0.9046$ & $0.0004 \pm 0.0000$ \\
& & Llama 3.3 & 10 & 10 & 0 & $31.8165 \pm 1.3131$ & $0.0003 \pm 0.0000$ \\
\cmidrule[0.01pt](r){2-8}
& \multirow{6}{*}{3} 
& Opus 4.6 & 10 & 10 & 0 & $84.6565 \pm 2.9336$ & $0.0005 \pm 0.0000$ \\
& & Sonnet 4.6 & 10 & 10 & 0 & $148.5577 \pm 15.0090$ & $0.0006 \pm 0.0000$ \\
& & GPT 5.2 & 10 & 10 & 0 & $157.6051 \pm 6.8779$ & $0.0005 \pm 0.0000$ \\
& & GPT 5.4 Mini & 10 & 10 & 0 & $77.6321 \pm 3.3945$ & $0.0006 \pm 0.0000$ \\
& & Qwen 3.6 & 10 & 10 & 0 & $134.2760 \pm 5.4431$ & $0.0004 \pm 0.0001$ \\
& & Llama 3.3 & 10 & 10 & 0 & $118.6948 \pm 3.1681$ & $0.0004 \pm 0.0000$ \\
\cmidrule[0.01pt](r){2-8}
& \multirow{6}{*}{4} 
& Opus 4.6 & 10 & 10 & 0 & $23.7966 \pm 0.7531$ & $0.0005 \pm 0.0000$ \\
& & Sonnet 4.6 & 10 & 10 & 0 & $25.4595 \pm 0.8862$ & $0.0006 \pm 0.0000$ \\
& & GPT 5.2 & 10 & 10 & 0 & $28.6600 \pm 1.3789$ & $0.0005 \pm 0.0000$ \\
& & GPT 5.4 Mini & 10 & 10 & 0 & $21.6864 \pm 2.3714$ & $0.0005 \pm 0.0000$ \\
& & Qwen 3.6 & 10 & 10 & 0 & $25.2426 \pm 0.7093$ & $0.0005 \pm 0.0000$ \\
& & Llama 3.3 & 10 & 10 & 0 & $26.2028 \pm 0.8225$ & $0.0005 \pm 0.0001$ \\
\cmidrule[0.01pt](r){2-8}
& \multirow{6}{*}{5} 
& Opus 4.6 & 10 & 10 & 0 & $1.1347 \pm 0.0386$ & $0.0004 \pm 0.0000$ \\
& & Sonnet 4.6 & 10 & 10 & 0 & $1.0397 \pm 0.0537$ & $0.0003 \pm 0.0000$ \\
& & GPT 5.2 & 10 & 10 & 0 & $11.3286 \pm 0.5409$ & $0.0003 \pm 0.0000$ \\
& & GPT 5.4 Mini & 10 & 10 & 0 & $0.9284 \pm 0.0244$ & $0.0004 \pm 0.0000$ \\
& & Qwen 3.6 & 10 & 10 & 0 & $0.8895 \pm 0.0068$ & $0.0003 \pm 0.0000$ \\
& & Llama 3.3 & 10 & 10 & 0 & $1.1244 \pm 0.0456$ & $0.0003 \pm 0.0000$ \\
\cmidrule[0.01pt](r){2-8}
& \multirow{6}{*}{6} 
& Opus 4.6 & 10 & 10 & 0 & $32.5850 \pm 1.1896$ & $0.0008 \pm 0.0000$ \\
& & Sonnet 4.6 & 10 & 10 & 0 & $27.1489 \pm 0.9987$ & $0.0007 \pm 0.0000$ \\
& & GPT 5.2 & 10 & 10 & 0 & $72.4745 \pm 8.2026$ & $0.0007 \pm 0.0000$ \\
& & GPT 5.4 Mini & 10 & 10 & 0 & $3.3065 \pm 0.1284$ & $0.0008 \pm 0.0000$ \\
& & Qwen 3.6 & 10 & 10 & 0 & $33.7885 \pm 1.4014$ & $0.0005 \pm 0.0000$ \\
& & Llama 3.3 & 10 & 10 & 0 & $31.1359 \pm 1.4736$ & $0.0005 \pm 0.0000$ \\
\bottomrule
\end{tabular}

%% file: figures/experiments/timings_enablers_bw.tex
\definecolor{mplBlue}{HTML}{1F77B4}
\definecolor{mplOrange}{HTML}{FF7F0E}
\definecolor{mplGreen}{HTML}{2CA02C}
\definecolor{mplRed}{HTML}{D62728}
\definecolor{mplPurple}{HTML}{9467BD}
\begin{tikzpicture}
\begin{axis}[
  width=0.96\textwidth,
  height=0.42\textwidth,
  xlabel={\footnotesize Test instance and model},
  ylabel={\footnotesize Mean time (s)},
  xmin=-0.5, xmax=28.5,
  ymin=0, ymax=0.0036564,
  xtick={0,1,2,3,4,5,6,7,8,9,10,11,12,13,14,15,16,17,18,19,20,21,22,23,24,25,26,27,28},
  xticklabels={{1 -- Opus 4.6},{1 -- Sonnet 4.6},{1 -- GPT5.2},{1 -- GPT5.4 Mini},{1 -- Qwen},{1 -- Llama},{2 -- Opus 4.6},{2 -- Sonnet 4.6},{2 -- GPT5.2},{2 -- Qwen},{2 -- Llama},{3 -- Opus 4.6},{3 -- Sonnet 4.6},{3 -- GPT5.2},{3 -- GPT5.4 Mini},{3 -- Qwen},{3 -- Llama},{4 -- Opus 4.6},{4 -- Sonnet 4.6},{4 -- GPT5.2},{4 -- GPT5.4 Mini},{4 -- Qwen},{4 -- Llama},{6 -- Opus 4.6},{6 -- Sonnet 4.6},{6 -- GPT5.2},{6 -- GPT5.4 Mini},{6 -- Qwen},{6 -- Llama}},
  tick align=outside,
  tick pos=left,
  x tick label style={rotate=90, anchor=east, font=\scriptsize},
  y tick label style={font=\scriptsize},
  label style={font=\small},
  ymajorgrids=true,
  grid style={dotted, gray!45},
  minor grid style={dotted, gray!25},
  legend style={at={(0.02,0.98)}, anchor=north west, draw=black!25, fill=white, fill opacity=0.88, text opacity=1, font=\scriptsize},
  legend cell align={left},
  axis line style={black!70},
  clip=false
]
\addplot[ybar, area legend, bar width=0.6512, fill=mplBlue, draw=black, line width=0.4pt] coordinates {(0,0.0022619) (1,0.00222697) (2,0.002209) (3,0.00220587) (4,0.00223897) (5,0.00220513) (6,0.00253668) (7,0.00255487) (8,0.0025964) (9,0.00249333) (10,0.00250957) (11,0.00287981) (12,0.00291147) (13,0.00289831) (14,0.00287631) (15,0.00286396) (16,0.00286679) (17,0.00306993) (18,0.00307193) (19,0.00294824) (20,0.00308509) (21,0.00307546) (22,0.00302691) (23,0.00233324) (24,0.00228097) (25,0.00232561) (26,0.00313661) (27,0.00240753) (28,0.00230877)};
\addlegendentry{Enablers}
\draw[black, line width=0.8pt] (axis cs:0,0.00221069) -- (axis cs:0,0.00231311);
\draw[black, line width=0.8pt] (axis cs:-0.045,0.00221069) -- (axis cs:0.045,0.00221069);
\draw[black, line width=0.8pt] (axis cs:-0.045,0.00231311) -- (axis cs:0.045,0.00231311);
\draw[black, line width=0.8pt] (axis cs:1,0.00216641) -- (axis cs:1,0.00228754);
\draw[black, line width=0.8pt] (axis cs:0.955,0.00216641) -- (axis cs:1.045,0.00216641);
\draw[black, line width=0.8pt] (axis cs:0.955,0.00228754) -- (axis cs:1.045,0.00228754);
\draw[black, line width=0.8pt] (axis cs:2,0.00185535) -- (axis cs:2,0.00256264);
\draw[black, line width=0.8pt] (axis cs:1.955,0.00185535) -- (axis cs:2.045,0.00185535);
\draw[black, line width=0.8pt] (axis cs:1.955,0.00256264) -- (axis cs:2.045,0.00256264);
\draw[black, line width=0.8pt] (axis cs:3,0.002133) -- (axis cs:3,0.00227874);
\draw[black, line width=0.8pt] (axis cs:2.955,0.002133) -- (axis cs:3.045,0.002133);
\draw[black, line width=0.8pt] (axis cs:2.955,0.00227874) -- (axis cs:3.045,0.00227874);
\draw[black, line width=0.8pt] (axis cs:4,0.00217225) -- (axis cs:4,0.00230568);
\draw[black, line width=0.8pt] (axis cs:3.955,0.00217225) -- (axis cs:4.045,0.00217225);
\draw[black, line width=0.8pt] (axis cs:3.955,0.00230568) -- (axis cs:4.045,0.00230568);
\draw[black, line width=0.8pt] (axis cs:5,0.00212601) -- (axis cs:5,0.00228426);
\draw[black, line width=0.8pt] (axis cs:4.955,0.00212601) -- (axis cs:5.045,0.00212601);
\draw[black, line width=0.8pt] (axis cs:4.955,0.00228426) -- (axis cs:5.045,0.00228426);
\draw[black, line width=0.8pt] (axis cs:6,0.00247965) -- (axis cs:6,0.00259371);
\draw[black, line width=0.8pt] (axis cs:5.955,0.00247965) -- (axis cs:6.045,0.00247965);
\draw[black, line width=0.8pt] (axis cs:5.955,0.00259371) -- (axis cs:6.045,0.00259371);
\draw[black, line width=0.8pt] (axis cs:7,0.00253358) -- (axis cs:7,0.00257615);
\draw[black, line width=0.8pt] (axis cs:6.955,0.00253358) -- (axis cs:7.045,0.00253358);
\draw[black, line width=0.8pt] (axis cs:6.955,0.00257615) -- (axis cs:7.045,0.00257615);
\draw[black, line width=0.8pt] (axis cs:8,0.00257417) -- (axis cs:8,0.00261863);
\draw[black, line width=0.8pt] (axis cs:7.955,0.00257417) -- (axis cs:8.045,0.00257417);
\draw[black, line width=0.8pt] (axis cs:7.955,0.00261863) -- (axis cs:8.045,0.00261863);
\draw[black, line width=0.8pt] (axis cs:9,0.0024755) -- (axis cs:9,0.00251117);
\draw[black, line width=0.8pt] (axis cs:8.955,0.0024755) -- (axis cs:9.045,0.0024755);
\draw[black, line width=0.8pt] (axis cs:8.955,0.00251117) -- (axis cs:9.045,0.00251117);
\draw[black, line width=0.8pt] (axis cs:10,0.00249368) -- (axis cs:10,0.00252546);
\draw[black, line width=0.8pt] (axis cs:9.955,0.00249368) -- (axis cs:10.045,0.00249368);
\draw[black, line width=0.8pt] (axis cs:9.955,0.00252546) -- (axis cs:10.045,0.00252546);
\draw[black, line width=0.8pt] (axis cs:11,0.00285174) -- (axis cs:11,0.00290788);
\draw[black, line width=0.8pt] (axis cs:10.955,0.00285174) -- (axis cs:11.045,0.00285174);
\draw[black, line width=0.8pt] (axis cs:10.955,0.00290788) -- (axis cs:11.045,0.00290788);
\draw[black, line width=0.8pt] (axis cs:12,0.00282814) -- (axis cs:12,0.0029948);
\draw[black, line width=0.8pt] (axis cs:11.955,0.00282814) -- (axis cs:12.045,0.00282814);
\draw[black, line width=0.8pt] (axis cs:11.955,0.0029948) -- (axis cs:12.045,0.0029948);
\draw[black, line width=0.8pt] (axis cs:13,0.0028718) -- (axis cs:13,0.00292482);
\draw[black, line width=0.8pt] (axis cs:12.955,0.0028718) -- (axis cs:13.045,0.0028718);
\draw[black, line width=0.8pt] (axis cs:12.955,0.00292482) -- (axis cs:13.045,0.00292482);
\draw[black, line width=0.8pt] (axis cs:14,0.00285191) -- (axis cs:14,0.00290071);
\draw[black, line width=0.8pt] (axis cs:13.955,0.00285191) -- (axis cs:14.045,0.00285191);
\draw[black, line width=0.8pt] (axis cs:13.955,0.00290071) -- (axis cs:14.045,0.00290071);
\draw[black, line width=0.8pt] (axis cs:15,0.00282521) -- (axis cs:15,0.0029027);
\draw[black, line width=0.8pt] (axis cs:14.955,0.00282521) -- (axis cs:15.045,0.00282521);
\draw[black, line width=0.8pt] (axis cs:14.955,0.0029027) -- (axis cs:15.045,0.0029027);
\draw[black, line width=0.8pt] (axis cs:16,0.00283952) -- (axis cs:16,0.00289407);
\draw[black, line width=0.8pt] (axis cs:15.955,0.00283952) -- (axis cs:16.045,0.00283952);
\draw[black, line width=0.8pt] (axis cs:15.955,0.00289407) -- (axis cs:16.045,0.00289407);
\draw[black, line width=0.8pt] (axis cs:17,0.00303521) -- (axis cs:17,0.00310464);
\draw[black, line width=0.8pt] (axis cs:16.955,0.00303521) -- (axis cs:17.045,0.00303521);
\draw[black, line width=0.8pt] (axis cs:16.955,0.00310464) -- (axis cs:17.045,0.00310464);
\draw[black, line width=0.8pt] (axis cs:18,0.00304323) -- (axis cs:18,0.00310063);
\draw[black, line width=0.8pt] (axis cs:17.955,0.00304323) -- (axis cs:18.045,0.00304323);
\draw[black, line width=0.8pt] (axis cs:17.955,0.00310063) -- (axis cs:18.045,0.00310063);
\draw[black, line width=0.8pt] (axis cs:19,0.00288491) -- (axis cs:19,0.00301156);
\draw[black, line width=0.8pt] (axis cs:18.955,0.00288491) -- (axis cs:19.045,0.00288491);
\draw[black, line width=0.8pt] (axis cs:18.955,0.00301156) -- (axis cs:19.045,0.00301156);
\draw[black, line width=0.8pt] (axis cs:20,0.0030613) -- (axis cs:20,0.00310887);
\draw[black, line width=0.8pt] (axis cs:19.955,0.0030613) -- (axis cs:20.045,0.0030613);
\draw[black, line width=0.8pt] (axis cs:19.955,0.00310887) -- (axis cs:20.045,0.00310887);
\draw[black, line width=0.8pt] (axis cs:21,0.00305547) -- (axis cs:21,0.00309544);
\draw[black, line width=0.8pt] (axis cs:20.955,0.00305547) -- (axis cs:21.045,0.00305547);
\draw[black, line width=0.8pt] (axis cs:20.955,0.00309544) -- (axis cs:21.045,0.00309544);
\draw[black, line width=0.8pt] (axis cs:22,0.00300049) -- (axis cs:22,0.00305334);
\draw[black, line width=0.8pt] (axis cs:21.955,0.00300049) -- (axis cs:22.045,0.00300049);
\draw[black, line width=0.8pt] (axis cs:21.955,0.00305334) -- (axis cs:22.045,0.00305334);
\draw[black, line width=0.8pt] (axis cs:23,0.00228933) -- (axis cs:23,0.00237714);
\draw[black, line width=0.8pt] (axis cs:22.955,0.00228933) -- (axis cs:23.045,0.00228933);
\draw[black, line width=0.8pt] (axis cs:22.955,0.00237714) -- (axis cs:23.045,0.00237714);
\draw[black, line width=0.8pt] (axis cs:24,0.00222665) -- (axis cs:24,0.0023353);
\draw[black, line width=0.8pt] (axis cs:23.955,0.00222665) -- (axis cs:24.045,0.00222665);
\draw[black, line width=0.8pt] (axis cs:23.955,0.0023353) -- (axis cs:24.045,0.0023353);
\draw[black, line width=0.8pt] (axis cs:25,0.00226233) -- (axis cs:25,0.00238889);
\draw[black, line width=0.8pt] (axis cs:24.955,0.00226233) -- (axis cs:25.045,0.00226233);
\draw[black, line width=0.8pt] (axis cs:24.955,0.00238889) -- (axis cs:25.045,0.00238889);
\draw[black, line width=0.8pt] (axis cs:26,0.00309374) -- (axis cs:26,0.00317948);
\draw[black, line width=0.8pt] (axis cs:25.955,0.00309374) -- (axis cs:26.045,0.00309374);
\draw[black, line width=0.8pt] (axis cs:25.955,0.00317948) -- (axis cs:26.045,0.00317948);
\draw[black, line width=0.8pt] (axis cs:27,0.00235496) -- (axis cs:27,0.0024601);
\draw[black, line width=0.8pt] (axis cs:26.955,0.00235496) -- (axis cs:27.045,0.00235496);
\draw[black, line width=0.8pt] (axis cs:26.955,0.0024601) -- (axis cs:27.045,0.0024601);
\draw[black, line width=0.8pt] (axis cs:28,0.00223641) -- (axis cs:28,0.00238114);
\draw[black, line width=0.8pt] (axis cs:27.955,0.00223641) -- (axis cs:28.045,0.00223641);
\draw[black, line width=0.8pt] (axis cs:27.955,0.00238114) -- (axis cs:28.045,0.00238114);
\end{axis}
\end{tikzpicture}

%% file: figures/experiments/timings_enablers_g.tex
% Requires: \usepackage{pgfplots}
% Recommended: \pgfplotsset{compat=1.18}
\definecolor{mplBlue}{HTML}{1F77B4}
\definecolor{mplOrange}{HTML}{FF7F0E}
\definecolor{mplGreen}{HTML}{2CA02C}
\definecolor{mplRed}{HTML}{D62728}
\definecolor{mplPurple}{HTML}{9467BD}
\begin{tikzpicture}
\begin{axis}[
  width=0.96\textwidth,
  height=0.42\textwidth,
  xlabel={\footnotesize Test instance and model},
  ylabel={\footnotesize Mean time (s)},
  xmin=-0.5, xmax=35.5,
  ymin=0, ymax=0.0200321,
  xtick={0,1,2,3,4,5,6,7,8,9,10,11,12,13,14,15,16,17,18,19,20,21,22,23,24,25,26,27,28,29,30,31,32,33,34,35},
  xticklabels={{1 -- Opus 4.6},{1 -- Sonnet 4.6},{1 -- GPT5.2},{1 -- GPT5.4 Mini},{1 -- Qwen},{1 -- Llama},{2 -- Opus 4.6},{2 -- Sonnet 4.6},{2 -- GPT5.2},{2 -- GPT5.4 Mini},{2 -- Qwen},{2 -- Llama},{3 -- Opus 4.6},{3 -- Sonnet 4.6},{3 -- GPT5.2},{3 -- GPT5.4 Mini},{3 -- Qwen},{3 -- Llama},{4 -- Opus 4.6},{4 -- Sonnet 4.6},{4 -- GPT5.2},{4 -- GPT5.4 Mini},{4 -- Qwen},{4 -- Llama},{5 -- Opus 4.6},{5 -- Sonnet 4.6},{5 -- GPT5.2},{5 -- GPT5.4 Mini},{5 -- Qwen},{5 -- Llama},{6 -- Opus 4.6},{6 -- Sonnet 4.6},{6 -- GPT5.2},{6 -- GPT5.4 Mini},{6 -- Qwen},{6 -- Llama}},
  tick align=outside,
  tick pos=left,
  x tick label style={rotate=90, anchor=east, font=\scriptsize},
  y tick label style={font=\scriptsize},
  label style={font=\small},
  ymajorgrids=true,
  grid style={dotted, gray!45},
  minor grid style={dotted, gray!25},
  legend style={at={(0.02,0.98)}, anchor=north west, draw=black!25, fill=white, fill opacity=0.88, text opacity=1, font=\scriptsize},
  legend cell align={left},
  axis line style={black!70},
  clip=false
]
\addplot[ybar, area legend, bar width=0.6512, fill=mplBlue, draw=black, line width=0.4pt] coordinates {(0,0.00478966) (1,0.00466869) (2,0.00434208) (3,0.00422101) (4,0.00378602) (5,0.00432615) (6,0.00817077) (7,0.00869865) (8,0.00686851) (9,0.00676293) (10,0.00806322) (11,0.00826268) (12,0.010448) (13,0.0116675) (14,0.00939229) (15,0.00956829) (16,0.00824475) (17,0.00917287) (18,0.00820627) (19,0.0106264) (20,0.00831165) (21,0.00994203) (22,0.00877414) (23,0.00716126) (24,0.00752795) (25,0.00636594) (26,0.00655863) (27,0.00700338) (28,0.00668697) (29,0.00585892) (30,0.0147551) (31,0.0137424) (32,0.0117665) (33,0.0143883) (34,0.0109456) (35,0.00991249)};
\addlegendentry{Enablers}
\draw[black, line width=0.8pt] (axis cs:0,0.00473013) -- (axis cs:0,0.0048492);
\draw[black, line width=0.8pt] (axis cs:-0.045,0.00473013) -- (axis cs:0.045,0.00473013);
\draw[black, line width=0.8pt] (axis cs:-0.045,0.0048492) -- (axis cs:0.045,0.0048492);
\draw[black, line width=0.8pt] (axis cs:1,0.00462727) -- (axis cs:1,0.0047101);
\draw[black, line width=0.8pt] (axis cs:0.955,0.00462727) -- (axis cs:1.045,0.00462727);
\draw[black, line width=0.8pt] (axis cs:0.955,0.0047101) -- (axis cs:1.045,0.0047101);
\draw[black, line width=0.8pt] (axis cs:2,0.00428948) -- (axis cs:2,0.00439468);
\draw[black, line width=0.8pt] (axis cs:1.955,0.00428948) -- (axis cs:2.045,0.00428948);
\draw[black, line width=0.8pt] (axis cs:1.955,0.00439468) -- (axis cs:2.045,0.00439468);
\draw[black, line width=0.8pt] (axis cs:3,0.00416294) -- (axis cs:3,0.00427908);
\draw[black, line width=0.8pt] (axis cs:2.955,0.00416294) -- (axis cs:3.045,0.00416294);
\draw[black, line width=0.8pt] (axis cs:2.955,0.00427908) -- (axis cs:3.045,0.00427908);
\draw[black, line width=0.8pt] (axis cs:4,0.0034919) -- (axis cs:4,0.00408013);
\draw[black, line width=0.8pt] (axis cs:3.955,0.0034919) -- (axis cs:4.045,0.0034919);
\draw[black, line width=0.8pt] (axis cs:3.955,0.00408013) -- (axis cs:4.045,0.00408013);
\draw[black, line width=0.8pt] (axis cs:5,0.00426564) -- (axis cs:5,0.00438667);
\draw[black, line width=0.8pt] (axis cs:4.955,0.00426564) -- (axis cs:5.045,0.00426564);
\draw[black, line width=0.8pt] (axis cs:4.955,0.00438667) -- (axis cs:5.045,0.00438667);
\draw[black, line width=0.8pt] (axis cs:6,0.00806378) -- (axis cs:6,0.00827777);
\draw[black, line width=0.8pt] (axis cs:5.955,0.00806378) -- (axis cs:6.045,0.00806378);
\draw[black, line width=0.8pt] (axis cs:5.955,0.00827777) -- (axis cs:6.045,0.00827777);
\draw[black, line width=0.8pt] (axis cs:7,0.00863718) -- (axis cs:7,0.00876013);
\draw[black, line width=0.8pt] (axis cs:6.955,0.00863718) -- (axis cs:7.045,0.00863718);
\draw[black, line width=0.8pt] (axis cs:6.955,0.00876013) -- (axis cs:7.045,0.00876013);
\draw[black, line width=0.8pt] (axis cs:8,0.0068018) -- (axis cs:8,0.00693521);
\draw[black, line width=0.8pt] (axis cs:7.955,0.0068018) -- (axis cs:8.045,0.0068018);
\draw[black, line width=0.8pt] (axis cs:7.955,0.00693521) -- (axis cs:8.045,0.00693521);
\draw[black, line width=0.8pt] (axis cs:9,0.00662471) -- (axis cs:9,0.00690115);
\draw[black, line width=0.8pt] (axis cs:8.955,0.00662471) -- (axis cs:9.045,0.00662471);
\draw[black, line width=0.8pt] (axis cs:8.955,0.00690115) -- (axis cs:9.045,0.00690115);
\draw[black, line width=0.8pt] (axis cs:10,0.00798196) -- (axis cs:10,0.00814449);
\draw[black, line width=0.8pt] (axis cs:9.955,0.00798196) -- (axis cs:10.045,0.00798196);
\draw[black, line width=0.8pt] (axis cs:9.955,0.00814449) -- (axis cs:10.045,0.00814449);
\draw[black, line width=0.8pt] (axis cs:11,0.00817095) -- (axis cs:11,0.00835442);
\draw[black, line width=0.8pt] (axis cs:10.955,0.00817095) -- (axis cs:11.045,0.00817095);
\draw[black, line width=0.8pt] (axis cs:10.955,0.00835442) -- (axis cs:11.045,0.00835442);
\draw[black, line width=0.8pt] (axis cs:12,0.00970169) -- (axis cs:12,0.0111943);
\draw[black, line width=0.8pt] (axis cs:11.955,0.00970169) -- (axis cs:12.045,0.00970169);
\draw[black, line width=0.8pt] (axis cs:11.955,0.0111943) -- (axis cs:12.045,0.0111943);
\draw[black, line width=0.8pt] (axis cs:13,0.0113384) -- (axis cs:13,0.0119966);
\draw[black, line width=0.8pt] (axis cs:12.955,0.0113384) -- (axis cs:13.045,0.0113384);
\draw[black, line width=0.8pt] (axis cs:12.955,0.0119966) -- (axis cs:13.045,0.0119966);
\draw[black, line width=0.8pt] (axis cs:14,0.00931001) -- (axis cs:14,0.00947456);
\draw[black, line width=0.8pt] (axis cs:13.955,0.00931001) -- (axis cs:14.045,0.00931001);
\draw[black, line width=0.8pt] (axis cs:13.955,0.00947456) -- (axis cs:14.045,0.00947456);
\draw[black, line width=0.8pt] (axis cs:15,0.00901172) -- (axis cs:15,0.0101249);
\draw[black, line width=0.8pt] (axis cs:14.955,0.00901172) -- (axis cs:15.045,0.00901172);
\draw[black, line width=0.8pt] (axis cs:14.955,0.0101249) -- (axis cs:15.045,0.0101249);
\draw[black, line width=0.8pt] (axis cs:16,0.0081128) -- (axis cs:16,0.0083767);
\draw[black, line width=0.8pt] (axis cs:15.955,0.0081128) -- (axis cs:16.045,0.0081128);
\draw[black, line width=0.8pt] (axis cs:15.955,0.0083767) -- (axis cs:16.045,0.0083767);
\draw[black, line width=0.8pt] (axis cs:17,0.0090877) -- (axis cs:17,0.00925804);
\draw[black, line width=0.8pt] (axis cs:16.955,0.0090877) -- (axis cs:17.045,0.0090877);
\draw[black, line width=0.8pt] (axis cs:16.955,0.00925804) -- (axis cs:17.045,0.00925804);
\draw[black, line width=0.8pt] (axis cs:18,0.00810015) -- (axis cs:18,0.00831239);
\draw[black, line width=0.8pt] (axis cs:17.955,0.00810015) -- (axis cs:18.045,0.00810015);
\draw[black, line width=0.8pt] (axis cs:17.955,0.00831239) -- (axis cs:18.045,0.00831239);
\draw[black, line width=0.8pt] (axis cs:19,0.0105374) -- (axis cs:19,0.0107155);
\draw[black, line width=0.8pt] (axis cs:18.955,0.0105374) -- (axis cs:19.045,0.0105374);
\draw[black, line width=0.8pt] (axis cs:18.955,0.0107155) -- (axis cs:19.045,0.0107155);
\draw[black, line width=0.8pt] (axis cs:20,0.00821747) -- (axis cs:20,0.00840583);
\draw[black, line width=0.8pt] (axis cs:19.955,0.00821747) -- (axis cs:20.045,0.00821747);
\draw[black, line width=0.8pt] (axis cs:19.955,0.00840583) -- (axis cs:20.045,0.00840583);
\draw[black, line width=0.8pt] (axis cs:21,0.00961246) -- (axis cs:21,0.0102716);
\draw[black, line width=0.8pt] (axis cs:20.955,0.00961246) -- (axis cs:21.045,0.00961246);
\draw[black, line width=0.8pt] (axis cs:20.955,0.0102716) -- (axis cs:21.045,0.0102716);
\draw[black, line width=0.8pt] (axis cs:22,0.00868355) -- (axis cs:22,0.00886472);
\draw[black, line width=0.8pt] (axis cs:21.955,0.00868355) -- (axis cs:22.045,0.00868355);
\draw[black, line width=0.8pt] (axis cs:21.955,0.00886472) -- (axis cs:22.045,0.00886472);
\draw[black, line width=0.8pt] (axis cs:23,0.00697035) -- (axis cs:23,0.00735217);
\draw[black, line width=0.8pt] (axis cs:22.955,0.00697035) -- (axis cs:23.045,0.00697035);
\draw[black, line width=0.8pt] (axis cs:22.955,0.00735217) -- (axis cs:23.045,0.00735217);
\draw[black, line width=0.8pt] (axis cs:24,0.00739004) -- (axis cs:24,0.00766586);
\draw[black, line width=0.8pt] (axis cs:23.955,0.00739004) -- (axis cs:24.045,0.00739004);
\draw[black, line width=0.8pt] (axis cs:23.955,0.00766586) -- (axis cs:24.045,0.00766586);
\draw[black, line width=0.8pt] (axis cs:25,0.00629328) -- (axis cs:25,0.00643861);
\draw[black, line width=0.8pt] (axis cs:24.955,0.00629328) -- (axis cs:25.045,0.00629328);
\draw[black, line width=0.8pt] (axis cs:24.955,0.00643861) -- (axis cs:25.045,0.00643861);
\draw[black, line width=0.8pt] (axis cs:26,0.00651486) -- (axis cs:26,0.00660241);
\draw[black, line width=0.8pt] (axis cs:25.955,0.00651486) -- (axis cs:26.045,0.00651486);
\draw[black, line width=0.8pt] (axis cs:25.955,0.00660241) -- (axis cs:26.045,0.00660241);
\draw[black, line width=0.8pt] (axis cs:27,0.00693471) -- (axis cs:27,0.00707205);
\draw[black, line width=0.8pt] (axis cs:26.955,0.00693471) -- (axis cs:27.045,0.00693471);
\draw[black, line width=0.8pt] (axis cs:26.955,0.00707205) -- (axis cs:27.045,0.00707205);
\draw[black, line width=0.8pt] (axis cs:28,0.00664906) -- (axis cs:28,0.00672488);
\draw[black, line width=0.8pt] (axis cs:27.955,0.00664906) -- (axis cs:28.045,0.00664906);
\draw[black, line width=0.8pt] (axis cs:27.955,0.00672488) -- (axis cs:28.045,0.00672488);
\draw[black, line width=0.8pt] (axis cs:29,0.00582225) -- (axis cs:29,0.00589559);
\draw[black, line width=0.8pt] (axis cs:28.955,0.00582225) -- (axis cs:29.045,0.00582225);
\draw[black, line width=0.8pt] (axis cs:28.955,0.00589559) -- (axis cs:29.045,0.00589559);
\draw[black, line width=0.8pt] (axis cs:30,0.0146681) -- (axis cs:30,0.014842);
\draw[black, line width=0.8pt] (axis cs:29.955,0.0146681) -- (axis cs:30.045,0.0146681);
\draw[black, line width=0.8pt] (axis cs:29.955,0.014842) -- (axis cs:30.045,0.014842);
\draw[black, line width=0.8pt] (axis cs:31,0.0136402) -- (axis cs:31,0.0138446);
\draw[black, line width=0.8pt] (axis cs:30.955,0.0136402) -- (axis cs:31.045,0.0136402);
\draw[black, line width=0.8pt] (axis cs:30.955,0.0138446) -- (axis cs:31.045,0.0138446);
\draw[black, line width=0.8pt] (axis cs:32,0.0113143) -- (axis cs:32,0.0122187);
\draw[black, line width=0.8pt] (axis cs:31.955,0.0113143) -- (axis cs:32.045,0.0113143);
\draw[black, line width=0.8pt] (axis cs:31.955,0.0122187) -- (axis cs:32.045,0.0122187);
\draw[black, line width=0.8pt] (axis cs:33,0.0142588) -- (axis cs:33,0.0145177);
\draw[black, line width=0.8pt] (axis cs:32.955,0.0142588) -- (axis cs:33.045,0.0142588);
\draw[black, line width=0.8pt] (axis cs:32.955,0.0145177) -- (axis cs:33.045,0.0145177);
\draw[black, line width=0.8pt] (axis cs:34,0.0108635) -- (axis cs:34,0.0110277);
\draw[black, line width=0.8pt] (axis cs:33.955,0.0108635) -- (axis cs:34.045,0.0108635);
\draw[black, line width=0.8pt] (axis cs:33.955,0.0110277) -- (axis cs:34.045,0.0110277);
\draw[black, line width=0.8pt] (axis cs:35,0.00973263) -- (axis cs:35,0.0100924);
\draw[black, line width=0.8pt] (axis cs:34.955,0.00973263) -- (axis cs:35.045,0.00973263);
\draw[black, line width=0.8pt] (axis cs:34.955,0.0100924) -- (axis cs:35.045,0.0100924);
\end{axis}
\end{tikzpicture}

%% file: figures/experiments/timings_stn_bw.tex
% Requires: \usepackage{pgfplots}
% Recommended: \pgfplotsset{compat=1.18}
\definecolor{mplBlue}{HTML}{1F77B4}
\definecolor{mplOrange}{HTML}{FF7F0E}
\definecolor{mplGreen}{HTML}{2CA02C}
\definecolor{mplRed}{HTML}{D62728}
\definecolor{mplPurple}{HTML}{9467BD}
\begin{tikzpicture}
\begin{axis}[
  width=0.96\textwidth,
  height=0.42\textwidth,
  xlabel={\footnotesize Test instance and model},
  ylabel={\footnotesize Mean time (s)},
  xmin=-0.5, xmax=28.5,
  ymin=0, ymax=0.05,
  ytick={0,0.01,0.02,0.03,0.04},
  xtick={0,1,2,3,4,5,6,7,8,9,10,11,12,13,14,15,16,17,18,19,20,21,22,23,24,25,26,27,28},
  xticklabels={{1 -- Opus 4.6},{1 -- Sonnet 4.6},{1 -- GPT5.2},{1 -- GPT5.4 Mini},{1 -- Qwen},{1 -- Llama},{2 -- Opus 4.6},{2 -- Sonnet 4.6},{2 -- GPT5.2},{2 -- Qwen},{2 -- Llama},{3 -- Opus 4.6},{3 -- Sonnet 4.6},{3 -- GPT5.2},{3 -- GPT5.4 Mini},{3 -- Qwen},{3 -- Llama},{4 -- Opus 4.6},{4 -- Sonnet 4.6},{4 -- GPT5.2},{4 -- GPT5.4 Mini},{4 -- Qwen},{4 -- Llama},{6 -- Opus 4.6},{6 -- Sonnet 4.6},{6 -- GPT5.2},{6 -- GPT5.4 Mini},{6 -- Qwen},{6 -- Llama}},
  tick align=outside,
  tick pos=left,
  x tick label style={rotate=90, anchor=east, font=\scriptsize},
  y tick label style={font=\scriptsize},
  label style={font=\small},
  ymajorgrids=true,
  grid style={dotted, gray!45},
  minor grid style={dotted, gray!25},
  legend style={at={(0.02,0.98)}, anchor=north west, draw=black!25, fill=white, fill opacity=0.88, text opacity=1, font=\scriptsize},
  legend cell align={left},
  axis line style={black!70},
  clip=false
]
\addplot[ybar, area legend, bar width=0.3256, fill=mplBlue, draw=black, line width=0.4pt] coordinates {(-0.185,0.0026218) (0.815,0.0025776) (1.815,0.002757) (2.815,0.0026058) (3.815,0.0024919) (4.815,0.0024781) (5.815,0.0046555) (6.815,0.0045268) (7.815,0.0045458) (8.815,0.0045907) (9.815,0.0045604) (10.815,0.0067089) (11.815,0.0064807) (12.815,0.0066003) (13.815,0.0065786) (14.815,0.0063654) (15.815,0.0063554) (16.815,0.006398) (17.815,0.0066556) (18.815,0.0064007) (19.815,0.0064735) (20.815,0.0066604) (21.815,0.006446) (22.815,0.0045184) (23.815,0.00458) (24.815,0.0045388) (25.815,0.0045573) (26.815,0.0044818) (27.815,0.0044994)};
\addlegendentry{STN build}
\addplot[ybar, area legend, bar width=0.3256, fill=mplOrange, draw=black, line width=0.4pt] coordinates {(0.185,0.0292018) (1.185,0.0292966) (2.185,0.0300634) (3.185,0.0286596) (4.185,0.0286071) (5.185,0.0299999) (6.185,0.0289472) (7.185,0.0291573) (8.185,0.0293286) (9.185,0.028992) (10.185,0.0292183) (11.185,0.0299428) (12.185,0.0297446) (13.185,0.0302368) (14.185,0.0299702) (15.185,0.0298399) (16.185,0.0297835) (17.185,0.0301079) (18.185,0.0297258) (19.185,0.0302255) (20.185,0.0298539) (21.185,0.0300362) (22.185,0.0299314) (23.185,0.0290974) (24.185,0.0291046) (25.185,0.0289261) (26.185,0.0290268) (27.185,0.0290031) (28.185,0.0290279)};
\addlegendentry{Optimization}
\draw[black, line width=0.8pt] (axis cs:-0.185,0.00248828) -- (axis cs:-0.185,0.00275532);
\draw[black, line width=0.8pt] (axis cs:-0.23,0.00248828) -- (axis cs:-0.14,0.00248828);
\draw[black, line width=0.8pt] (axis cs:-0.23,0.00275532) -- (axis cs:-0.14,0.00275532);
\draw[black, line width=0.8pt] (axis cs:0.815,0.00250284) -- (axis cs:0.815,0.00265236);
\draw[black, line width=0.8pt] (axis cs:0.77,0.00250284) -- (axis cs:0.86,0.00250284);
\draw[black, line width=0.8pt] (axis cs:0.77,0.00265236) -- (axis cs:0.86,0.00265236);
\draw[black, line width=0.8pt] (axis cs:1.815,0.00247606) -- (axis cs:1.815,0.00303794);
\draw[black, line width=0.8pt] (axis cs:1.77,0.00247606) -- (axis cs:1.86,0.00247606);
\draw[black, line width=0.8pt] (axis cs:1.77,0.00303794) -- (axis cs:1.86,0.00303794);
\draw[black, line width=0.8pt] (axis cs:2.815,0.00236376) -- (axis cs:2.815,0.00284784);
\draw[black, line width=0.8pt] (axis cs:2.77,0.00236376) -- (axis cs:2.86,0.00236376);
\draw[black, line width=0.8pt] (axis cs:2.77,0.00284784) -- (axis cs:2.86,0.00284784);
\draw[black, line width=0.8pt] (axis cs:3.815,0.00235311) -- (axis cs:3.815,0.00263069);
\draw[black, line width=0.8pt] (axis cs:3.77,0.00235311) -- (axis cs:3.86,0.00235311);
\draw[black, line width=0.8pt] (axis cs:3.77,0.00263069) -- (axis cs:3.86,0.00263069);
\draw[black, line width=0.8pt] (axis cs:4.815,0.00242942) -- (axis cs:4.815,0.00252678);
\draw[black, line width=0.8pt] (axis cs:4.77,0.00242942) -- (axis cs:4.86,0.00242942);
\draw[black, line width=0.8pt] (axis cs:4.77,0.00252678) -- (axis cs:4.86,0.00252678);
\draw[black, line width=0.8pt] (axis cs:5.815,0.00439378) -- (axis cs:5.815,0.00491722);
\draw[black, line width=0.8pt] (axis cs:5.77,0.00439378) -- (axis cs:5.86,0.00439378);
\draw[black, line width=0.8pt] (axis cs:5.77,0.00491722) -- (axis cs:5.86,0.00491722);
\draw[black, line width=0.8pt] (axis cs:6.815,0.00449874) -- (axis cs:6.815,0.00455486);
\draw[black, line width=0.8pt] (axis cs:6.77,0.00449874) -- (axis cs:6.86,0.00449874);
\draw[black, line width=0.8pt] (axis cs:6.77,0.00455486) -- (axis cs:6.86,0.00455486);
\draw[black, line width=0.8pt] (axis cs:7.815,0.00447284) -- (axis cs:7.815,0.00461876);
\draw[black, line width=0.8pt] (axis cs:7.77,0.00447284) -- (axis cs:7.86,0.00447284);
\draw[black, line width=0.8pt] (axis cs:7.77,0.00461876) -- (axis cs:7.86,0.00461876);
\draw[black, line width=0.8pt] (axis cs:8.815,0.00439275) -- (axis cs:8.815,0.00478865);
\draw[black, line width=0.8pt] (axis cs:8.77,0.00439275) -- (axis cs:8.86,0.00439275);
\draw[black, line width=0.8pt] (axis cs:8.77,0.00478865) -- (axis cs:8.86,0.00478865);
\draw[black, line width=0.8pt] (axis cs:9.815,0.0045044) -- (axis cs:9.815,0.0046164);
\draw[black, line width=0.8pt] (axis cs:9.77,0.0045044) -- (axis cs:9.86,0.0045044);
\draw[black, line width=0.8pt] (axis cs:9.77,0.0046164) -- (axis cs:9.86,0.0046164);
\draw[black, line width=0.8pt] (axis cs:10.815,0.0062741) -- (axis cs:10.815,0.0071437);
\draw[black, line width=0.8pt] (axis cs:10.77,0.0062741) -- (axis cs:10.86,0.0062741);
\draw[black, line width=0.8pt] (axis cs:10.77,0.0071437) -- (axis cs:10.86,0.0071437);
\draw[black, line width=0.8pt] (axis cs:11.815,0.00617987) -- (axis cs:11.815,0.00678153);
\draw[black, line width=0.8pt] (axis cs:11.77,0.00617987) -- (axis cs:11.86,0.00617987);
\draw[black, line width=0.8pt] (axis cs:11.77,0.00678153) -- (axis cs:11.86,0.00678153);
\draw[black, line width=0.8pt] (axis cs:12.815,0.00622872) -- (axis cs:12.815,0.00697188);
\draw[black, line width=0.8pt] (axis cs:12.77,0.00622872) -- (axis cs:12.86,0.00622872);
\draw[black, line width=0.8pt] (axis cs:12.77,0.00697188) -- (axis cs:12.86,0.00697188);
\draw[black, line width=0.8pt] (axis cs:13.815,0.0062556) -- (axis cs:13.815,0.0069016);
\draw[black, line width=0.8pt] (axis cs:13.77,0.0062556) -- (axis cs:13.86,0.0062556);
\draw[black, line width=0.8pt] (axis cs:13.77,0.0069016) -- (axis cs:13.86,0.0069016);
\draw[black, line width=0.8pt] (axis cs:14.815,0.00632465) -- (axis cs:14.815,0.00640615);
\draw[black, line width=0.8pt] (axis cs:14.77,0.00632465) -- (axis cs:14.86,0.00632465);
\draw[black, line width=0.8pt] (axis cs:14.77,0.00640615) -- (axis cs:14.86,0.00640615);
\draw[black, line width=0.8pt] (axis cs:15.815,0.00628501) -- (axis cs:15.815,0.00642579);
\draw[black, line width=0.8pt] (axis cs:15.77,0.00628501) -- (axis cs:15.86,0.00628501);
\draw[black, line width=0.8pt] (axis cs:15.77,0.00642579) -- (axis cs:15.86,0.00642579);
\draw[black, line width=0.8pt] (axis cs:16.815,0.00631558) -- (axis cs:16.815,0.00648042);
\draw[black, line width=0.8pt] (axis cs:16.77,0.00631558) -- (axis cs:16.86,0.00631558);
\draw[black, line width=0.8pt] (axis cs:16.77,0.00648042) -- (axis cs:16.86,0.00648042);
\draw[black, line width=0.8pt] (axis cs:17.815,0.00623872) -- (axis cs:17.815,0.00707248);
\draw[black, line width=0.8pt] (axis cs:17.77,0.00623872) -- (axis cs:17.86,0.00623872);
\draw[black, line width=0.8pt] (axis cs:17.77,0.00707248) -- (axis cs:17.86,0.00707248);
\draw[black, line width=0.8pt] (axis cs:18.815,0.00634831) -- (axis cs:18.815,0.00645309);
\draw[black, line width=0.8pt] (axis cs:18.77,0.00634831) -- (axis cs:18.86,0.00634831);
\draw[black, line width=0.8pt] (axis cs:18.77,0.00645309) -- (axis cs:18.86,0.00645309);
\draw[black, line width=0.8pt] (axis cs:19.815,0.0062002) -- (axis cs:19.815,0.0067468);
\draw[black, line width=0.8pt] (axis cs:19.77,0.0062002) -- (axis cs:19.86,0.0062002);
\draw[black, line width=0.8pt] (axis cs:19.77,0.0067468) -- (axis cs:19.86,0.0067468);
\draw[black, line width=0.8pt] (axis cs:20.815,0.00621723) -- (axis cs:20.815,0.00710357);
\draw[black, line width=0.8pt] (axis cs:20.77,0.00621723) -- (axis cs:20.86,0.00621723);
\draw[black, line width=0.8pt] (axis cs:20.77,0.00710357) -- (axis cs:20.86,0.00710357);
\draw[black, line width=0.8pt] (axis cs:21.815,0.00640687) -- (axis cs:21.815,0.00648513);
\draw[black, line width=0.8pt] (axis cs:21.77,0.00640687) -- (axis cs:21.86,0.00640687);
\draw[black, line width=0.8pt] (axis cs:21.77,0.00648513) -- (axis cs:21.86,0.00648513);
\draw[black, line width=0.8pt] (axis cs:22.815,0.00447291) -- (axis cs:22.815,0.00456389);
\draw[black, line width=0.8pt] (axis cs:22.77,0.00447291) -- (axis cs:22.86,0.00447291);
\draw[black, line width=0.8pt] (axis cs:22.77,0.00456389) -- (axis cs:22.86,0.00456389);
\draw[black, line width=0.8pt] (axis cs:23.815,0.00435582) -- (axis cs:23.815,0.00480418);
\draw[black, line width=0.8pt] (axis cs:23.77,0.00435582) -- (axis cs:23.86,0.00435582);
\draw[black, line width=0.8pt] (axis cs:23.77,0.00480418) -- (axis cs:23.86,0.00480418);
\draw[black, line width=0.8pt] (axis cs:24.815,0.00439004) -- (axis cs:24.815,0.00468756);
\draw[black, line width=0.8pt] (axis cs:24.77,0.00439004) -- (axis cs:24.86,0.00439004);
\draw[black, line width=0.8pt] (axis cs:24.77,0.00468756) -- (axis cs:24.86,0.00468756);
\draw[black, line width=0.8pt] (axis cs:25.815,0.0043586) -- (axis cs:25.815,0.004756);
\draw[black, line width=0.8pt] (axis cs:25.77,0.0043586) -- (axis cs:25.86,0.0043586);
\draw[black, line width=0.8pt] (axis cs:25.77,0.004756) -- (axis cs:25.86,0.004756);
\draw[black, line width=0.8pt] (axis cs:26.815,0.00444816) -- (axis cs:26.815,0.00451544);
\draw[black, line width=0.8pt] (axis cs:26.77,0.00444816) -- (axis cs:26.86,0.00444816);
\draw[black, line width=0.8pt] (axis cs:26.77,0.00451544) -- (axis cs:26.86,0.00451544);
\draw[black, line width=0.8pt] (axis cs:27.815,0.00446073) -- (axis cs:27.815,0.00453807);
\draw[black, line width=0.8pt] (axis cs:27.77,0.00446073) -- (axis cs:27.86,0.00446073);
\draw[black, line width=0.8pt] (axis cs:27.77,0.00453807) -- (axis cs:27.86,0.00453807);
\draw[black, line width=0.8pt] (axis cs:0.185,0.0283425) -- (axis cs:0.185,0.0300611);
\draw[black, line width=0.8pt] (axis cs:0.14,0.0283425) -- (axis cs:0.23,0.0283425);
\draw[black, line width=0.8pt] (axis cs:0.14,0.0300611) -- (axis cs:0.23,0.0300611);
\draw[black, line width=0.8pt] (axis cs:1.185,0.0284449) -- (axis cs:1.185,0.0301483);
\draw[black, line width=0.8pt] (axis cs:1.14,0.0284449) -- (axis cs:1.23,0.0284449);
\draw[black, line width=0.8pt] (axis cs:1.14,0.0301483) -- (axis cs:1.23,0.0301483);
\draw[black, line width=0.8pt] (axis cs:2.185,0.0274043) -- (axis cs:2.185,0.0327225);
\draw[black, line width=0.8pt] (axis cs:2.14,0.0274043) -- (axis cs:2.23,0.0274043);
\draw[black, line width=0.8pt] (axis cs:2.14,0.0327225) -- (axis cs:2.23,0.0327225);
\draw[black, line width=0.8pt] (axis cs:3.185,0.028074) -- (axis cs:3.185,0.0292452);
\draw[black, line width=0.8pt] (axis cs:3.14,0.028074) -- (axis cs:3.23,0.028074);
\draw[black, line width=0.8pt] (axis cs:3.14,0.0292452) -- (axis cs:3.23,0.0292452);
\draw[black, line width=0.8pt] (axis cs:4.185,0.0280845) -- (axis cs:4.185,0.0291297);
\draw[black, line width=0.8pt] (axis cs:4.14,0.0280845) -- (axis cs:4.23,0.0280845);
\draw[black, line width=0.8pt] (axis cs:4.14,0.0291297) -- (axis cs:4.23,0.0291297);
\draw[black, line width=0.8pt] (axis cs:5.185,0.026437) -- (axis cs:5.185,0.0335628);
\draw[black, line width=0.8pt] (axis cs:5.14,0.026437) -- (axis cs:5.23,0.026437);
\draw[black, line width=0.8pt] (axis cs:5.14,0.0335628) -- (axis cs:5.23,0.0335628);
\draw[black, line width=0.8pt] (axis cs:6.185,0.028695) -- (axis cs:6.185,0.0291994);
\draw[black, line width=0.8pt] (axis cs:6.14,0.028695) -- (axis cs:6.23,0.028695);
\draw[black, line width=0.8pt] (axis cs:6.14,0.0291994) -- (axis cs:6.23,0.0291994);
\draw[black, line width=0.8pt] (axis cs:7.185,0.0286423) -- (axis cs:7.185,0.0296723);
\draw[black, line width=0.8pt] (axis cs:7.14,0.0286423) -- (axis cs:7.23,0.0286423);
\draw[black, line width=0.8pt] (axis cs:7.14,0.0296723) -- (axis cs:7.23,0.0296723);
\draw[black, line width=0.8pt] (axis cs:8.185,0.0286784) -- (axis cs:8.185,0.0299788);
\draw[black, line width=0.8pt] (axis cs:8.14,0.0286784) -- (axis cs:8.23,0.0286784);
\draw[black, line width=0.8pt] (axis cs:8.14,0.0299788) -- (axis cs:8.23,0.0299788);
\draw[black, line width=0.8pt] (axis cs:9.185,0.0283702) -- (axis cs:9.185,0.0296138);
\draw[black, line width=0.8pt] (axis cs:9.14,0.0283702) -- (axis cs:9.23,0.0283702);
\draw[black, line width=0.8pt] (axis cs:9.14,0.0296138) -- (axis cs:9.23,0.0296138);
\draw[black, line width=0.8pt] (axis cs:10.185,0.0285928) -- (axis cs:10.185,0.0298438);
\draw[black, line width=0.8pt] (axis cs:10.14,0.0285928) -- (axis cs:10.23,0.0285928);
\draw[black, line width=0.8pt] (axis cs:10.14,0.0298438) -- (axis cs:10.23,0.0298438);
\draw[black, line width=0.8pt] (axis cs:11.185,0.0294026) -- (axis cs:11.185,0.030483);
\draw[black, line width=0.8pt] (axis cs:11.14,0.0294026) -- (axis cs:11.23,0.0294026);
\draw[black, line width=0.8pt] (axis cs:11.14,0.030483) -- (axis cs:11.23,0.030483);
\draw[black, line width=0.8pt] (axis cs:12.185,0.0289372) -- (axis cs:12.185,0.030552);
\draw[black, line width=0.8pt] (axis cs:12.14,0.0289372) -- (axis cs:12.23,0.0289372);
\draw[black, line width=0.8pt] (axis cs:12.14,0.030552) -- (axis cs:12.23,0.030552);
\draw[black, line width=0.8pt] (axis cs:13.185,0.0296264) -- (axis cs:13.185,0.0308472);
\draw[black, line width=0.8pt] (axis cs:13.14,0.0296264) -- (axis cs:13.23,0.0296264);
\draw[black, line width=0.8pt] (axis cs:13.14,0.0308472) -- (axis cs:13.23,0.0308472);
\draw[black, line width=0.8pt] (axis cs:14.185,0.0294162) -- (axis cs:14.185,0.0305242);
\draw[black, line width=0.8pt] (axis cs:14.14,0.0294162) -- (axis cs:14.23,0.0294162);
\draw[black, line width=0.8pt] (axis cs:14.14,0.0305242) -- (axis cs:14.23,0.0305242);
\draw[black, line width=0.8pt] (axis cs:15.185,0.0291162) -- (axis cs:15.185,0.0305636);
\draw[black, line width=0.8pt] (axis cs:15.14,0.0291162) -- (axis cs:15.23,0.0291162);
\draw[black, line width=0.8pt] (axis cs:15.14,0.0305636) -- (axis cs:15.23,0.0305636);
\draw[black, line width=0.8pt] (axis cs:16.185,0.0293445) -- (axis cs:16.185,0.0302225);
\draw[black, line width=0.8pt] (axis cs:16.14,0.0293445) -- (axis cs:16.23,0.0293445);
\draw[black, line width=0.8pt] (axis cs:16.14,0.0302225) -- (axis cs:16.23,0.0302225);
\draw[black, line width=0.8pt] (axis cs:17.185,0.0294264) -- (axis cs:17.185,0.0307894);
\draw[black, line width=0.8pt] (axis cs:17.14,0.0294264) -- (axis cs:17.23,0.0294264);
\draw[black, line width=0.8pt] (axis cs:17.14,0.0307894) -- (axis cs:17.23,0.0307894);
\draw[black, line width=0.8pt] (axis cs:18.185,0.0292341) -- (axis cs:18.185,0.0302175);
\draw[black, line width=0.8pt] (axis cs:18.14,0.0292341) -- (axis cs:18.23,0.0292341);
\draw[black, line width=0.8pt] (axis cs:18.14,0.0302175) -- (axis cs:18.23,0.0302175);
\draw[black, line width=0.8pt] (axis cs:19.185,0.0296258) -- (axis cs:19.185,0.0308252);
\draw[black, line width=0.8pt] (axis cs:19.14,0.0296258) -- (axis cs:19.23,0.0296258);
\draw[black, line width=0.8pt] (axis cs:19.14,0.0308252) -- (axis cs:19.23,0.0308252);
\draw[black, line width=0.8pt] (axis cs:20.185,0.0295699) -- (axis cs:20.185,0.0301379);
\draw[black, line width=0.8pt] (axis cs:20.14,0.0295699) -- (axis cs:20.23,0.0295699);
\draw[black, line width=0.8pt] (axis cs:20.14,0.0301379) -- (axis cs:20.23,0.0301379);
\draw[black, line width=0.8pt] (axis cs:21.185,0.0295091) -- (axis cs:21.185,0.0305633);
\draw[black, line width=0.8pt] (axis cs:21.14,0.0295091) -- (axis cs:21.23,0.0295091);
\draw[black, line width=0.8pt] (axis cs:21.14,0.0305633) -- (axis cs:21.23,0.0305633);
\draw[black, line width=0.8pt] (axis cs:22.185,0.0293402) -- (axis cs:22.185,0.0305226);
\draw[black, line width=0.8pt] (axis cs:22.14,0.0293402) -- (axis cs:22.23,0.0293402);
\draw[black, line width=0.8pt] (axis cs:22.14,0.0305226) -- (axis cs:22.23,0.0305226);
\draw[black, line width=0.8pt] (axis cs:23.185,0.0285147) -- (axis cs:23.185,0.0296801);
\draw[black, line width=0.8pt] (axis cs:23.14,0.0285147) -- (axis cs:23.23,0.0285147);
\draw[black, line width=0.8pt] (axis cs:23.14,0.0296801) -- (axis cs:23.23,0.0296801);
\draw[black, line width=0.8pt] (axis cs:24.185,0.0284999) -- (axis cs:24.185,0.0297093);
\draw[black, line width=0.8pt] (axis cs:24.14,0.0284999) -- (axis cs:24.23,0.0284999);
\draw[black, line width=0.8pt] (axis cs:24.14,0.0297093) -- (axis cs:24.23,0.0297093);
\draw[black, line width=0.8pt] (axis cs:25.185,0.0285316) -- (axis cs:25.185,0.0293206);
\draw[black, line width=0.8pt] (axis cs:25.14,0.0285316) -- (axis cs:25.23,0.0285316);
\draw[black, line width=0.8pt] (axis cs:25.14,0.0293206) -- (axis cs:25.23,0.0293206);
\draw[black, line width=0.8pt] (axis cs:26.185,0.0284119) -- (axis cs:26.185,0.0296417);
\draw[black, line width=0.8pt] (axis cs:26.14,0.0284119) -- (axis cs:26.23,0.0284119);
\draw[black, line width=0.8pt] (axis cs:26.14,0.0296417) -- (axis cs:26.23,0.0296417);
\draw[black, line width=0.8pt] (axis cs:27.185,0.0282958) -- (axis cs:27.185,0.0297104);
\draw[black, line width=0.8pt] (axis cs:27.14,0.0282958) -- (axis cs:27.23,0.0282958);
\draw[black, line width=0.8pt] (axis cs:27.14,0.0297104) -- (axis cs:27.23,0.0297104);
\draw[black, line width=0.8pt] (axis cs:28.185,0.0284809) -- (axis cs:28.185,0.0295749);
\draw[black, line width=0.8pt] (axis cs:28.14,0.0284809) -- (axis cs:28.23,0.0284809);
\draw[black, line width=0.8pt] (axis cs:28.14,0.0295749) -- (axis cs:28.23,0.0295749);
\end{axis}
\end{tikzpicture}

%% file: figures/experiments/timings_stn_g.tex
% Requires: \usepackage{pgfplots}
% Recommended: \pgfplotsset{compat=1.18}
\definecolor{mplBlue}{HTML}{1F77B4}
\definecolor{mplOrange}{HTML}{FF7F0E}
\definecolor{mplGreen}{HTML}{2CA02C}
\definecolor{mplRed}{HTML}{D62728}
\definecolor{mplPurple}{HTML}{9467BD}
\begin{tikzpicture}
\begin{axis}[
  width=0.96\textwidth,
  height=0.42\textwidth,
  xlabel={\footnotesize Test instance and model},
  ylabel={\footnotesize Mean time (s)},
  xmin=-0.5, xmax=35.5,
  ymin=0, ymax=0.05,
  ytick={0,0.01,0.02,0.03,0.04},
  xtick={0,1,2,3,4,5,6,7,8,9,10,11,12,13,14,15,16,17,18,19,20,21,22,23,24,25,26,27,28,29,30,31,32,33,34,35},
  xticklabels={{1 -- Opus 4.6},{1 -- Sonnet 4.6},{1 -- GPT5.2},{1 -- GPT5.4 Mini},{1 -- Qwen},{1 -- Llama},{2 -- Opus 4.6},{2 -- Sonnet 4.6},{2 -- GPT5.2},{2 -- GPT5.4 Mini},{2 -- Qwen},{2 -- Llama},{3 -- Opus 4.6},{3 -- Sonnet 4.6},{3 -- GPT5.2},{3 -- GPT5.4 Mini},{3 -- Qwen},{3 -- Llama},{4 -- Opus 4.6},{4 -- Sonnet 4.6},{4 -- GPT5.2},{4 -- GPT5.4 Mini},{4 -- Qwen},{4 -- Llama},{5 -- Opus 4.6},{5 -- Sonnet 4.6},{5 -- GPT5.2},{5 -- GPT5.4 Mini},{5 -- Qwen},{5 -- Llama},{6 -- Opus 4.6},{6 -- Sonnet 4.6},{6 -- GPT5.2},{6 -- GPT5.4 Mini},{6 -- Qwen},{6 -- Llama}},
  tick align=outside,
  tick pos=left,
  x tick label style={rotate=90, anchor=east, font=\scriptsize},
  y tick label style={font=\scriptsize},
  label style={font=\small},
  ymajorgrids=true,
  grid style={dotted, gray!45},
  minor grid style={dotted, gray!25},
  legend style={at={(0.02,0.98)}, anchor=north west, draw=black!25, fill=white, fill opacity=0.88, text opacity=1, font=\scriptsize},
  legend cell align={left},
  axis line style={black!70},
  clip=false
]
\addplot[ybar, area legend, bar width=0.3256, fill=mplBlue, draw=black, line width=0.4pt] coordinates {(-0.185,0.0130964) (0.815,0.0131005) (1.815,0.0128007) (2.815,0.0130951) (3.815,0.0104775) (4.815,0.0127274) (5.815,0.0184626) (6.815,0.0186336) (7.815,0.0179424) (8.815,0.0177725) (9.815,0.018419) (10.815,0.0188036) (11.815,0.0222055) (12.815,0.0231478) (13.815,0.0244197) (14.815,0.0218028) (15.815,0.0212243) (16.815,0.0216568) (17.815,0.019117) (18.815,0.0235598) (19.815,0.0182097) (20.815,0.0194416) (21.815,0.0186639) (22.815,0.0183457) (23.815,0.0171943) (24.815,0.0160447) (25.815,0.0198624) (26.815,0.0189584) (27.815,0.0153474) (28.815,0.0170362) (29.815,0.028859) (30.815,0.0282619) (31.815,0.02915) (32.815,0.0316742) (33.815,0.0292131) (34.815,0.0260606)};
\addlegendentry{STN build}
\addplot[ybar, area legend, bar width=0.3256, fill=mplOrange, draw=black, line width=0.4pt] coordinates {(0.185,0.0314124) (1.185,0.0313772) (2.185,0.0317071) (3.185,0.0319595) (4.185,0.0303762) (5.185,0.0312095) (6.185,0.0333414) (7.185,0.0340693) (8.185,0.0332116) (9.185,0.0334346) (10.185,0.0338399) (11.185,0.0337583) (12.185,0.0345383) (13.185,0.0355825) (14.185,0.036168) (15.185,0.0347827) (16.185,0.0345167) (17.185,0.034501) (18.185,0.0337397) (19.185,0.0363167) (20.185,0.03363) (21.185,0.0342576) (22.185,0.0339166) (23.185,0.0335019) (24.185,0.0339338) (25.185,0.0330728) (26.185,0.0351862) (27.185,0.0352431) (28.185,0.0326589) (29.185,0.0334964) (30.185,0.0414709) (31.185,0.0409058) (32.185,0.0414278) (33.185,0.044678) (34.185,0.0425304) (35.185,0.0406997)};
\addlegendentry{Optimization}
\draw[black, line width=0.8pt] (axis cs:-0.185,0.0128504) -- (axis cs:-0.185,0.0133424);
\draw[black, line width=0.8pt] (axis cs:-0.23,0.0128504) -- (axis cs:-0.14,0.0128504);
\draw[black, line width=0.8pt] (axis cs:-0.23,0.0133424) -- (axis cs:-0.14,0.0133424);
\draw[black, line width=0.8pt] (axis cs:0.815,0.0130105) -- (axis cs:0.815,0.0131905);
\draw[black, line width=0.8pt] (axis cs:0.77,0.0130105) -- (axis cs:0.86,0.0130105);
\draw[black, line width=0.8pt] (axis cs:0.77,0.0131905) -- (axis cs:0.86,0.0131905);
\draw[black, line width=0.8pt] (axis cs:1.815,0.0125444) -- (axis cs:1.815,0.013057);
\draw[black, line width=0.8pt] (axis cs:1.77,0.0125444) -- (axis cs:1.86,0.0125444);
\draw[black, line width=0.8pt] (axis cs:1.77,0.013057) -- (axis cs:1.86,0.013057);
\draw[black, line width=0.8pt] (axis cs:2.815,0.0124398) -- (axis cs:2.815,0.0137504);
\draw[black, line width=0.8pt] (axis cs:2.77,0.0124398) -- (axis cs:2.86,0.0124398);
\draw[black, line width=0.8pt] (axis cs:2.77,0.0137504) -- (axis cs:2.86,0.0137504);
\draw[black, line width=0.8pt] (axis cs:3.815,0.0103661) -- (axis cs:3.815,0.0105889);
\draw[black, line width=0.8pt] (axis cs:3.77,0.0103661) -- (axis cs:3.86,0.0103661);
\draw[black, line width=0.8pt] (axis cs:3.77,0.0105889) -- (axis cs:3.86,0.0105889);
\draw[black, line width=0.8pt] (axis cs:4.815,0.0125606) -- (axis cs:4.815,0.0128942);
\draw[black, line width=0.8pt] (axis cs:4.77,0.0125606) -- (axis cs:4.86,0.0125606);
\draw[black, line width=0.8pt] (axis cs:4.77,0.0128942) -- (axis cs:4.86,0.0128942);
\draw[black, line width=0.8pt] (axis cs:5.815,0.017505) -- (axis cs:5.815,0.0194202);
\draw[black, line width=0.8pt] (axis cs:5.77,0.017505) -- (axis cs:5.86,0.017505);
\draw[black, line width=0.8pt] (axis cs:5.77,0.0194202) -- (axis cs:5.86,0.0194202);
\draw[black, line width=0.8pt] (axis cs:6.815,0.0177255) -- (axis cs:6.815,0.0195417);
\draw[black, line width=0.8pt] (axis cs:6.77,0.0177255) -- (axis cs:6.86,0.0177255);
\draw[black, line width=0.8pt] (axis cs:6.77,0.0195417) -- (axis cs:6.86,0.0195417);
\draw[black, line width=0.8pt] (axis cs:7.815,0.0178214) -- (axis cs:7.815,0.0180634);
\draw[black, line width=0.8pt] (axis cs:7.77,0.0178214) -- (axis cs:7.86,0.0178214);
\draw[black, line width=0.8pt] (axis cs:7.77,0.0180634) -- (axis cs:7.86,0.0180634);
\draw[black, line width=0.8pt] (axis cs:8.815,0.0176602) -- (axis cs:8.815,0.0178848);
\draw[black, line width=0.8pt] (axis cs:8.77,0.0176602) -- (axis cs:8.86,0.0176602);
\draw[black, line width=0.8pt] (axis cs:8.77,0.0178848) -- (axis cs:8.86,0.0178848);
\draw[black, line width=0.8pt] (axis cs:9.815,0.0174563) -- (axis cs:9.815,0.0193817);
\draw[black, line width=0.8pt] (axis cs:9.77,0.0174563) -- (axis cs:9.86,0.0174563);
\draw[black, line width=0.8pt] (axis cs:9.77,0.0193817) -- (axis cs:9.86,0.0193817);
\draw[black, line width=0.8pt] (axis cs:10.815,0.0185076) -- (axis cs:10.815,0.0190996);
\draw[black, line width=0.8pt] (axis cs:10.77,0.0185076) -- (axis cs:10.86,0.0185076);
\draw[black, line width=0.8pt] (axis cs:10.77,0.0190996) -- (axis cs:10.86,0.0190996);
\draw[black, line width=0.8pt] (axis cs:11.815,0.0219655) -- (axis cs:11.815,0.0224455);
\draw[black, line width=0.8pt] (axis cs:11.77,0.0219655) -- (axis cs:11.86,0.0219655);
\draw[black, line width=0.8pt] (axis cs:11.77,0.0224455) -- (axis cs:11.86,0.0224455);
\draw[black, line width=0.8pt] (axis cs:12.815,0.0218058) -- (axis cs:12.815,0.0244898);
\draw[black, line width=0.8pt] (axis cs:12.77,0.0218058) -- (axis cs:12.86,0.0218058);
\draw[black, line width=0.8pt] (axis cs:12.77,0.0244898) -- (axis cs:12.86,0.0244898);
\draw[black, line width=0.8pt] (axis cs:13.815,0.0236122) -- (axis cs:13.815,0.0252272);
\draw[black, line width=0.8pt] (axis cs:13.77,0.0236122) -- (axis cs:13.86,0.0236122);
\draw[black, line width=0.8pt] (axis cs:13.77,0.0252272) -- (axis cs:13.86,0.0252272);
\draw[black, line width=0.8pt] (axis cs:14.815,0.0215658) -- (axis cs:14.815,0.0220398);
\draw[black, line width=0.8pt] (axis cs:14.77,0.0215658) -- (axis cs:14.86,0.0215658);
\draw[black, line width=0.8pt] (axis cs:14.77,0.0220398) -- (axis cs:14.86,0.0220398);
\draw[black, line width=0.8pt] (axis cs:15.815,0.0209554) -- (axis cs:15.815,0.0214932);
\draw[black, line width=0.8pt] (axis cs:15.77,0.0209554) -- (axis cs:15.86,0.0209554);
\draw[black, line width=0.8pt] (axis cs:15.77,0.0214932) -- (axis cs:15.86,0.0214932);
\draw[black, line width=0.8pt] (axis cs:16.815,0.0215211) -- (axis cs:16.815,0.0217925);
\draw[black, line width=0.8pt] (axis cs:16.77,0.0215211) -- (axis cs:16.86,0.0215211);
\draw[black, line width=0.8pt] (axis cs:16.77,0.0217925) -- (axis cs:16.86,0.0217925);
\draw[black, line width=0.8pt] (axis cs:17.815,0.018108) -- (axis cs:17.815,0.020126);
\draw[black, line width=0.8pt] (axis cs:17.77,0.018108) -- (axis cs:17.86,0.018108);
\draw[black, line width=0.8pt] (axis cs:17.77,0.020126) -- (axis cs:17.86,0.020126);
\draw[black, line width=0.8pt] (axis cs:18.815,0.0234066) -- (axis cs:18.815,0.023713);
\draw[black, line width=0.8pt] (axis cs:18.77,0.0234066) -- (axis cs:18.86,0.0234066);
\draw[black, line width=0.8pt] (axis cs:18.77,0.023713) -- (axis cs:18.86,0.023713);
\draw[black, line width=0.8pt] (axis cs:19.815,0.0173593) -- (axis cs:19.815,0.0190601);
\draw[black, line width=0.8pt] (axis cs:19.77,0.0173593) -- (axis cs:19.86,0.0173593);
\draw[black, line width=0.8pt] (axis cs:19.77,0.0190601) -- (axis cs:19.86,0.0190601);
\draw[black, line width=0.8pt] (axis cs:20.815,0.0193385) -- (axis cs:20.815,0.0195447);
\draw[black, line width=0.8pt] (axis cs:20.77,0.0193385) -- (axis cs:20.86,0.0193385);
\draw[black, line width=0.8pt] (axis cs:20.77,0.0195447) -- (axis cs:20.86,0.0195447);
\draw[black, line width=0.8pt] (axis cs:21.815,0.0184996) -- (axis cs:21.815,0.0188282);
\draw[black, line width=0.8pt] (axis cs:21.77,0.0184996) -- (axis cs:21.86,0.0184996);
\draw[black, line width=0.8pt] (axis cs:21.77,0.0188282) -- (axis cs:21.86,0.0188282);
\draw[black, line width=0.8pt] (axis cs:22.815,0.0182634) -- (axis cs:22.815,0.018428);
\draw[black, line width=0.8pt] (axis cs:22.77,0.0182634) -- (axis cs:22.86,0.0182634);
\draw[black, line width=0.8pt] (axis cs:22.77,0.018428) -- (axis cs:22.86,0.018428);
\draw[black, line width=0.8pt] (axis cs:23.815,0.01634) -- (axis cs:23.815,0.0180486);
\draw[black, line width=0.8pt] (axis cs:23.77,0.01634) -- (axis cs:23.86,0.01634);
\draw[black, line width=0.8pt] (axis cs:23.77,0.0180486) -- (axis cs:23.86,0.0180486);
\draw[black, line width=0.8pt] (axis cs:24.815,0.0159311) -- (axis cs:24.815,0.0161583);
\draw[black, line width=0.8pt] (axis cs:24.77,0.0159311) -- (axis cs:24.86,0.0159311);
\draw[black, line width=0.8pt] (axis cs:24.77,0.0161583) -- (axis cs:24.86,0.0161583);
\draw[black, line width=0.8pt] (axis cs:25.815,0.0188978) -- (axis cs:25.815,0.020827);
\draw[black, line width=0.8pt] (axis cs:25.77,0.0188978) -- (axis cs:25.86,0.0188978);
\draw[black, line width=0.8pt] (axis cs:25.77,0.020827) -- (axis cs:25.86,0.020827);
\draw[black, line width=0.8pt] (axis cs:26.815,0.018834) -- (axis cs:26.815,0.0190828);
\draw[black, line width=0.8pt] (axis cs:26.77,0.018834) -- (axis cs:26.86,0.018834);
\draw[black, line width=0.8pt] (axis cs:26.77,0.0190828) -- (axis cs:26.86,0.0190828);
\draw[black, line width=0.8pt] (axis cs:27.815,0.0152249) -- (axis cs:27.815,0.0154699);
\draw[black, line width=0.8pt] (axis cs:27.77,0.0152249) -- (axis cs:27.86,0.0152249);
\draw[black, line width=0.8pt] (axis cs:27.77,0.0154699) -- (axis cs:27.86,0.0154699);
\draw[black, line width=0.8pt] (axis cs:28.815,0.0168066) -- (axis cs:28.815,0.0172658);
\draw[black, line width=0.8pt] (axis cs:28.77,0.0168066) -- (axis cs:28.86,0.0168066);
\draw[black, line width=0.8pt] (axis cs:28.77,0.0172658) -- (axis cs:28.86,0.0172658);
\draw[black, line width=0.8pt] (axis cs:29.815,0.028639) -- (axis cs:29.815,0.029079);
\draw[black, line width=0.8pt] (axis cs:29.77,0.028639) -- (axis cs:29.86,0.028639);
\draw[black, line width=0.8pt] (axis cs:29.77,0.029079) -- (axis cs:29.86,0.029079);
\draw[black, line width=0.8pt] (axis cs:30.815,0.0280868) -- (axis cs:30.815,0.028437);
\draw[black, line width=0.8pt] (axis cs:30.77,0.0280868) -- (axis cs:30.86,0.0280868);
\draw[black, line width=0.8pt] (axis cs:30.77,0.028437) -- (axis cs:30.86,0.028437);
\draw[black, line width=0.8pt] (axis cs:31.815,0.0273832) -- (axis cs:31.815,0.0309168);
\draw[black, line width=0.8pt] (axis cs:31.77,0.0273832) -- (axis cs:31.86,0.0273832);
\draw[black, line width=0.8pt] (axis cs:31.77,0.0309168) -- (axis cs:31.86,0.0309168);
\draw[black, line width=0.8pt] (axis cs:32.815,0.031357) -- (axis cs:32.815,0.0319914);
\draw[black, line width=0.8pt] (axis cs:32.77,0.031357) -- (axis cs:32.86,0.031357);
\draw[black, line width=0.8pt] (axis cs:32.77,0.0319914) -- (axis cs:32.86,0.0319914);
\draw[black, line width=0.8pt] (axis cs:33.815,0.0289421) -- (axis cs:33.815,0.0294841);
\draw[black, line width=0.8pt] (axis cs:33.77,0.0289421) -- (axis cs:33.86,0.0289421);
\draw[black, line width=0.8pt] (axis cs:33.77,0.0294841) -- (axis cs:33.86,0.0294841);
\draw[black, line width=0.8pt] (axis cs:34.815,0.024917) -- (axis cs:34.815,0.0272042);
\draw[black, line width=0.8pt] (axis cs:34.77,0.024917) -- (axis cs:34.86,0.024917);
\draw[black, line width=0.8pt] (axis cs:34.77,0.0272042) -- (axis cs:34.86,0.0272042);
\draw[black, line width=0.8pt] (axis cs:0.185,0.0310327) -- (axis cs:0.185,0.0317921);
\draw[black, line width=0.8pt] (axis cs:0.14,0.0310327) -- (axis cs:0.23,0.0310327);
\draw[black, line width=0.8pt] (axis cs:0.14,0.0317921) -- (axis cs:0.23,0.0317921);
\draw[black, line width=0.8pt] (axis cs:1.185,0.0310715) -- (axis cs:1.185,0.0316829);
\draw[black, line width=0.8pt] (axis cs:1.14,0.0310715) -- (axis cs:1.23,0.0310715);
\draw[black, line width=0.8pt] (axis cs:1.14,0.0316829) -- (axis cs:1.23,0.0316829);
\draw[black, line width=0.8pt] (axis cs:2.185,0.0310622) -- (axis cs:2.185,0.032352);
\draw[black, line width=0.8pt] (axis cs:2.14,0.0310622) -- (axis cs:2.23,0.0310622);
\draw[black, line width=0.8pt] (axis cs:2.14,0.032352) -- (axis cs:2.23,0.032352);
\draw[black, line width=0.8pt] (axis cs:3.185,0.0313372) -- (axis cs:3.185,0.0325818);
\draw[black, line width=0.8pt] (axis cs:3.14,0.0313372) -- (axis cs:3.23,0.0313372);
\draw[black, line width=0.8pt] (axis cs:3.14,0.0325818) -- (axis cs:3.23,0.0325818);
\draw[black, line width=0.8pt] (axis cs:4.185,0.0297572) -- (axis cs:4.185,0.0309952);
\draw[black, line width=0.8pt] (axis cs:4.14,0.0297572) -- (axis cs:4.23,0.0297572);
\draw[black, line width=0.8pt] (axis cs:4.14,0.0309952) -- (axis cs:4.23,0.0309952);
\draw[black, line width=0.8pt] (axis cs:5.185,0.0307074) -- (axis cs:5.185,0.0317116);
\draw[black, line width=0.8pt] (axis cs:5.14,0.0307074) -- (axis cs:5.23,0.0307074);
\draw[black, line width=0.8pt] (axis cs:5.14,0.0317116) -- (axis cs:5.23,0.0317116);
\draw[black, line width=0.8pt] (axis cs:6.185,0.0327215) -- (axis cs:6.185,0.0339613);
\draw[black, line width=0.8pt] (axis cs:6.14,0.0327215) -- (axis cs:6.23,0.0327215);
\draw[black, line width=0.8pt] (axis cs:6.14,0.0339613) -- (axis cs:6.23,0.0339613);
\draw[black, line width=0.8pt] (axis cs:7.185,0.0334816) -- (axis cs:7.185,0.034657);
\draw[black, line width=0.8pt] (axis cs:7.14,0.0334816) -- (axis cs:7.23,0.0334816);
\draw[black, line width=0.8pt] (axis cs:7.14,0.034657) -- (axis cs:7.23,0.034657);
\draw[black, line width=0.8pt] (axis cs:8.185,0.0326235) -- (axis cs:8.185,0.0337997);
\draw[black, line width=0.8pt] (axis cs:8.14,0.0326235) -- (axis cs:8.23,0.0326235);
\draw[black, line width=0.8pt] (axis cs:8.14,0.0337997) -- (axis cs:8.23,0.0337997);
\draw[black, line width=0.8pt] (axis cs:9.185,0.0327261) -- (axis cs:9.185,0.0341431);
\draw[black, line width=0.8pt] (axis cs:9.14,0.0327261) -- (axis cs:9.23,0.0327261);
\draw[black, line width=0.8pt] (axis cs:9.14,0.0341431) -- (axis cs:9.23,0.0341431);
\draw[black, line width=0.8pt] (axis cs:10.185,0.0333507) -- (axis cs:10.185,0.0343291);
\draw[black, line width=0.8pt] (axis cs:10.14,0.0333507) -- (axis cs:10.23,0.0333507);
\draw[black, line width=0.8pt] (axis cs:10.14,0.0343291) -- (axis cs:10.23,0.0343291);
\draw[black, line width=0.8pt] (axis cs:11.185,0.0332817) -- (axis cs:11.185,0.0342349);
\draw[black, line width=0.8pt] (axis cs:11.14,0.0332817) -- (axis cs:11.23,0.0332817);
\draw[black, line width=0.8pt] (axis cs:11.14,0.0342349) -- (axis cs:11.23,0.0342349);
\draw[black, line width=0.8pt] (axis cs:12.185,0.0337406) -- (axis cs:12.185,0.035336);
\draw[black, line width=0.8pt] (axis cs:12.14,0.0337406) -- (axis cs:12.23,0.0337406);
\draw[black, line width=0.8pt] (axis cs:12.14,0.035336) -- (axis cs:12.23,0.035336);
\draw[black, line width=0.8pt] (axis cs:13.185,0.0348958) -- (axis cs:13.185,0.0362692);
\draw[black, line width=0.8pt] (axis cs:13.14,0.0348958) -- (axis cs:13.23,0.0348958);
\draw[black, line width=0.8pt] (axis cs:13.14,0.0362692) -- (axis cs:13.23,0.0362692);
\draw[black, line width=0.8pt] (axis cs:14.185,0.0355116) -- (axis cs:14.185,0.0368244);
\draw[black, line width=0.8pt] (axis cs:14.14,0.0355116) -- (axis cs:14.23,0.0355116);
\draw[black, line width=0.8pt] (axis cs:14.14,0.0368244) -- (axis cs:14.23,0.0368244);
\draw[black, line width=0.8pt] (axis cs:15.185,0.0343011) -- (axis cs:15.185,0.0352643);
\draw[black, line width=0.8pt] (axis cs:15.14,0.0343011) -- (axis cs:15.23,0.0343011);
\draw[black, line width=0.8pt] (axis cs:15.14,0.0352643) -- (axis cs:15.23,0.0352643);
\draw[black, line width=0.8pt] (axis cs:16.185,0.0337908) -- (axis cs:16.185,0.0352426);
\draw[black, line width=0.8pt] (axis cs:16.14,0.0337908) -- (axis cs:16.23,0.0337908);
\draw[black, line width=0.8pt] (axis cs:16.14,0.0352426) -- (axis cs:16.23,0.0352426);
\draw[black, line width=0.8pt] (axis cs:17.185,0.033987) -- (axis cs:17.185,0.035015);
\draw[black, line width=0.8pt] (axis cs:17.14,0.033987) -- (axis cs:17.23,0.033987);
\draw[black, line width=0.8pt] (axis cs:17.14,0.035015) -- (axis cs:17.23,0.035015);
\draw[black, line width=0.8pt] (axis cs:18.185,0.0330909) -- (axis cs:18.185,0.0343885);
\draw[black, line width=0.8pt] (axis cs:18.14,0.0330909) -- (axis cs:18.23,0.0330909);
\draw[black, line width=0.8pt] (axis cs:18.14,0.0343885) -- (axis cs:18.23,0.0343885);
\draw[black, line width=0.8pt] (axis cs:19.185,0.0358238) -- (axis cs:19.185,0.0368096);
\draw[black, line width=0.8pt] (axis cs:19.14,0.0358238) -- (axis cs:19.23,0.0358238);
\draw[black, line width=0.8pt] (axis cs:19.14,0.0368096) -- (axis cs:19.23,0.0368096);
\draw[black, line width=0.8pt] (axis cs:20.185,0.0330079) -- (axis cs:20.185,0.0342521);
\draw[black, line width=0.8pt] (axis cs:20.14,0.0330079) -- (axis cs:20.23,0.0330079);
\draw[black, line width=0.8pt] (axis cs:20.14,0.0342521) -- (axis cs:20.23,0.0342521);
\draw[black, line width=0.8pt] (axis cs:21.185,0.0337039) -- (axis cs:21.185,0.0348113);
\draw[black, line width=0.8pt] (axis cs:21.14,0.0337039) -- (axis cs:21.23,0.0337039);
\draw[black, line width=0.8pt] (axis cs:21.14,0.0348113) -- (axis cs:21.23,0.0348113);
\draw[black, line width=0.8pt] (axis cs:22.185,0.0332963) -- (axis cs:22.185,0.0345369);
\draw[black, line width=0.8pt] (axis cs:22.14,0.0332963) -- (axis cs:22.23,0.0332963);
\draw[black, line width=0.8pt] (axis cs:22.14,0.0345369) -- (axis cs:22.23,0.0345369);
\draw[black, line width=0.8pt] (axis cs:23.185,0.0327468) -- (axis cs:23.185,0.034257);
\draw[black, line width=0.8pt] (axis cs:23.14,0.0327468) -- (axis cs:23.23,0.0327468);
\draw[black, line width=0.8pt] (axis cs:23.14,0.034257) -- (axis cs:23.23,0.034257);
\draw[black, line width=0.8pt] (axis cs:24.185,0.0331597) -- (axis cs:24.185,0.0347079);
\draw[black, line width=0.8pt] (axis cs:24.14,0.0331597) -- (axis cs:24.23,0.0331597);
\draw[black, line width=0.8pt] (axis cs:24.14,0.0347079) -- (axis cs:24.23,0.0347079);
\draw[black, line width=0.8pt] (axis cs:25.185,0.0326314) -- (axis cs:25.185,0.0335142);
\draw[black, line width=0.8pt] (axis cs:25.14,0.0326314) -- (axis cs:25.23,0.0326314);
\draw[black, line width=0.8pt] (axis cs:25.14,0.0335142) -- (axis cs:25.23,0.0335142);
\draw[black, line width=0.8pt] (axis cs:26.185,0.0342769) -- (axis cs:26.185,0.0360955);
\draw[black, line width=0.8pt] (axis cs:26.14,0.0342769) -- (axis cs:26.23,0.0342769);
\draw[black, line width=0.8pt] (axis cs:26.14,0.0360955) -- (axis cs:26.23,0.0360955);
\draw[black, line width=0.8pt] (axis cs:27.185,0.0347519) -- (axis cs:27.185,0.0357343);
\draw[black, line width=0.8pt] (axis cs:27.14,0.0347519) -- (axis cs:27.23,0.0347519);
\draw[black, line width=0.8pt] (axis cs:27.14,0.0357343) -- (axis cs:27.23,0.0357343);
\draw[black, line width=0.8pt] (axis cs:28.185,0.0321511) -- (axis cs:28.185,0.0331667);
\draw[black, line width=0.8pt] (axis cs:28.14,0.0321511) -- (axis cs:28.23,0.0321511);
\draw[black, line width=0.8pt] (axis cs:28.14,0.0331667) -- (axis cs:28.23,0.0331667);
\draw[black, line width=0.8pt] (axis cs:29.185,0.0329328) -- (axis cs:29.185,0.03406);
\draw[black, line width=0.8pt] (axis cs:29.14,0.0329328) -- (axis cs:29.23,0.0329328);
\draw[black, line width=0.8pt] (axis cs:29.14,0.03406) -- (axis cs:29.23,0.03406);
\draw[black, line width=0.8pt] (axis cs:30.185,0.0408835) -- (axis cs:30.185,0.0420583);
\draw[black, line width=0.8pt] (axis cs:30.14,0.0408835) -- (axis cs:30.23,0.0408835);
\draw[black, line width=0.8pt] (axis cs:30.14,0.0420583) -- (axis cs:30.23,0.0420583);
\draw[black, line width=0.8pt] (axis cs:31.185,0.0404333) -- (axis cs:31.185,0.0413783);
\draw[black, line width=0.8pt] (axis cs:31.14,0.0404333) -- (axis cs:31.23,0.0404333);
\draw[black, line width=0.8pt] (axis cs:31.14,0.0413783) -- (axis cs:31.23,0.0413783);
\draw[black, line width=0.8pt] (axis cs:32.185,0.0406714) -- (axis cs:32.185,0.0421842);
\draw[black, line width=0.8pt] (axis cs:32.14,0.0406714) -- (axis cs:32.23,0.0406714);
\draw[black, line width=0.8pt] (axis cs:32.14,0.0421842) -- (axis cs:32.23,0.0421842);
\draw[black, line width=0.8pt] (axis cs:33.185,0.0441648) -- (axis cs:33.185,0.0451912);
\draw[black, line width=0.8pt] (axis cs:33.14,0.0441648) -- (axis cs:33.23,0.0441648);
\draw[black, line width=0.8pt] (axis cs:33.14,0.0451912) -- (axis cs:33.23,0.0451912);
\draw[black, line width=0.8pt] (axis cs:34.185,0.0419047) -- (axis cs:34.185,0.0431561);
\draw[black, line width=0.8pt] (axis cs:34.14,0.0419047) -- (axis cs:34.23,0.0419047);
\draw[black, line width=0.8pt] (axis cs:34.14,0.0431561) -- (axis cs:34.23,0.0431561);
\draw[black, line width=0.8pt] (axis cs:35.185,0.0400599) -- (axis cs:35.185,0.0413395);
\draw[black, line width=0.8pt] (axis cs:35.14,0.0400599) -- (axis cs:35.23,0.0400599);
\draw[black, line width=0.8pt] (axis cs:35.14,0.0413395) -- (axis cs:35.23,0.0413395);
\end{axis}
\end{tikzpicture}